\documentclass{article}
\usepackage{graphicx} 
\usepackage[T1]{fontenc}         
\usepackage{amsfonts}      
\usepackage{nicefrac}      
\usepackage{microtype}   
\usepackage{CJKutf8}
\usepackage{array}

\newcommand{\zh}[1]{%
    \begin{CJK*}{UTF8}{gbsn}#1\end{CJK*}%
}
\usepackage[colorlinks,
            linkcolor=red,
            anchorcolor=blue,
            citecolor=blue, 
            pagebackref=true
           ]{hyperref}

\usepackage[a4paper,top=0.8in,bottom=1in,left=1in,right=1in]{geometry}
\usepackage{tabularx}
\usepackage{float}
\usepackage{enumitem}
\usepackage{subcaption}
\usepackage{wrapfig}
\usepackage{booktabs}

\usepackage{mmll}

\usepackage[capitalize,noabbrev]{cleveref}

\title{Extremely Sparse Supervision Incentivizes Reasoning Ability}
\author{
\textbf{Zhishuai Liu}$^{1,2}$\thanks{Work done during internship at Amazon.} \quad
\textbf{Xingzi Xu}$^{1}$ \quad
\textbf{Mehmet Saygin Seyfioglu}$^{1}$ \\[0.2em]
\textbf{Pan Xu}$^{2}$ \quad
\textbf{Karim Bouyarmane}$^{1}$ \\[0.8em]
$^{1}$Amazon, Seattle, WA 98109, USA \\
\texttt{\{zhishuai,xingzixu,mseyfiog,bouykari\}@amazon.com} \\[0.3em]
$^{2}$Duke University, Durham, NC 27708, USA \\
\texttt{\{zhishuai.liu,pan.xu\}@duke.edu}
}
\date{}

\begin{document}

\maketitle

\begin{abstract}
Large language models demonstrate increasingly strong reasoning capabilities through effective post-training. Yet, prevailing post-training methods optimize over massive numbers of tokens, implicitly assuming that effective learning must be token-intensive. We revisit this assumption in the on-policy distillation (OPD) setting, which naturally admits dense teacher supervision at every generated token. Using the Qwen3 family, \textbf{we discover a counter-intuitive phenomenon: reasoning can be effectively incentivized by an extremely small fraction of generated tokens---as few as one or two tokens per reasoning trajectory, corresponding to only 0.05\% of all tokens}. Surprisingly, this sparse supervision in most cases matches or surpasses full-token training in improving reasoning ability, despite excluding the vast majority of generated tokens from the training objective. This phenomenon is consistently observed across nine teacher--student configurations spanning different model scales on mathematical reasoning tasks, and is further validated on coding reasoning, Llama models and Proximal Policy Optimization (PPO)-based reinforcement learning with verifiable reward (RLVR).
Interestingly, such extremely sparse supervision may be closer to the natural learning process: rather than correcting every step word by word, one reflects on a few critical reasoning steps, updates one’s prior understanding, and continues the trial-and-error, avoiding micro-level corrections while remaining remarkably effective. Overall, our results challenge the assumption that effective post-training must be token-intensive and point to a new direction for understanding and designing more efficient post-training algorithms.

\begin{figure}[H]
    \centering
    \includegraphics[width=0.45\linewidth]{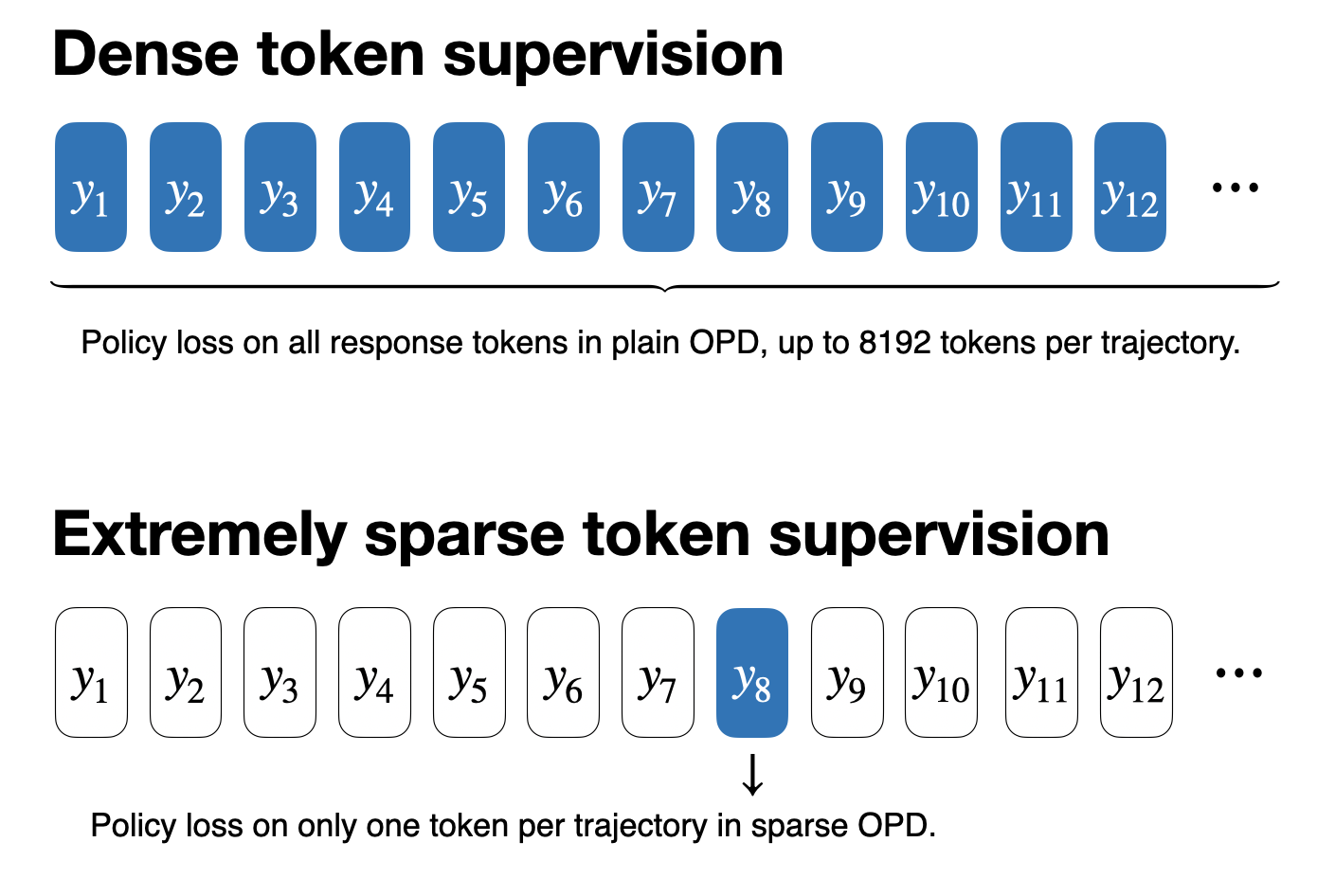}
    \includegraphics[width=0.48\linewidth]{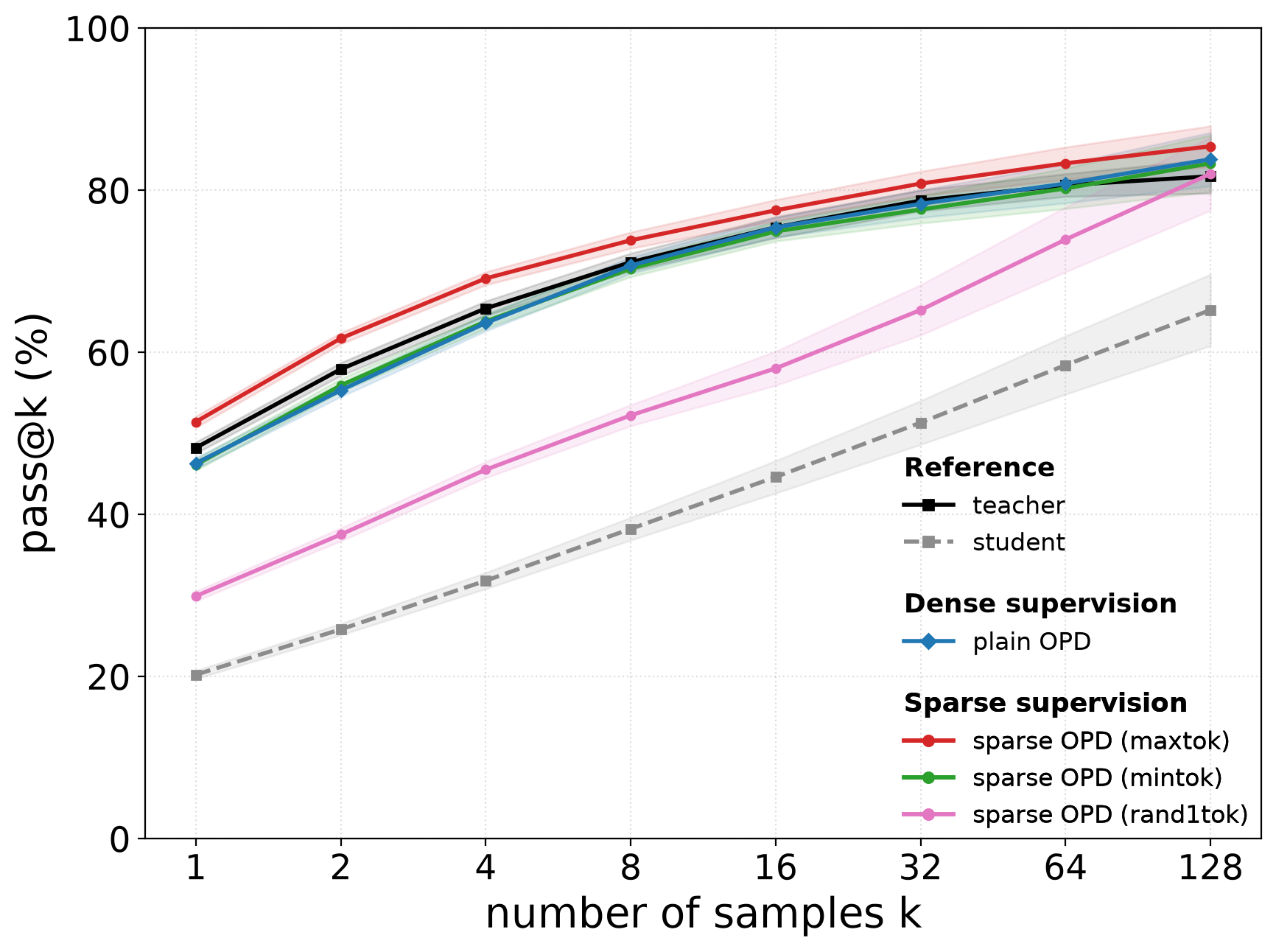}
    \caption{\textbf{Illustration of the mechanism and effectiveness of sparse OPD.} Left: sparse OPD supervises only one token per response trajectory. Right: AIME 2025 results for Qwen3-8B (student) $\leftarrow$ Qwen3-4B-Instruct-2507 (teacher), comparing the base student, teacher, plain OPD, and three sparse OPD variants: \textit{maxtok}, \textit{mintok}, and \textit{rand1tok}. \textit{maxtok} and \textit{mintok} select the tokens with the most positive and negative rewards, respectively, while \textit{rand1tok} randomly selects one token per trajectory. Shaded areas denote 95\% confidence intervals. Remarkably, \textit{maxtok} produces a student that outperforms both plain OPD and the teacher, \textit{mintok} matches plain OPD, and even \textit{rand1tok} substantially improves over the base student. \textbf{All three sparse OPD variants only use one token per trajectory.}}
    \label{fig:placeholder}
\end{figure}
\end{abstract}

\section{Introduction}

Post-training has become a critical stage for aligning large language models (LLMs) with human preferences \citep{ouyang2022training, bai2022training} and eliciting their reasoning capabilities \citep{lightman2024let, shao2024deepseekmath, guo2025deepseek, team2026kimi}. 
Several seminal post-training paradigms have emerged, including reinforcement learning from human feedback (RLHF), reinforcement learning from verifiable rewards (RLVR), and on-policy distillation (OPD), etc.  
Similar to pre-training, which scales model capability through increasing compute and dataset size, collecting hundreds of trillions of tokens in total pretraining data \citep{kaplan2020scaling, hoffmann2022training}, post-training has also been traditionally regarded as a token-intensive process. For example, in the DAPO training procedure \citep{yu2026dapo} on the Qwen2.5-32B Base model \citep{qwen2025qwen25technicalreport}, approximately 1,500 RL updates are performed, with each update collecting 8,192 rollouts of up to 16,384 tokens per trajectory. This corresponds to a total of hundreds of billions of generated tokens contributing gradient signals during RL optimization to improve the model's reasoning capability.
OPD improves the training efficiency of RL-based post-training by leveraging dense token-level supervision. From an information-theoretic perspective, \citet{lu2025onpolicydistillation} argues that reinforcement learning provides only $O(1)$ bits of learning signal per episode \citep{schulman2025lora}, while distillation provides $O(T)$ bits per episode, where $T$ denotes the sequence length. Consequently, OPD can substantially improve training efficiency over RLVR approaches by extracting richer learning signals from each generated trajectory. Nevertheless, OPD still operates in a token-intensive regime, retaining dense supervision over hundreds of millions of generated tokens during training. 

Recently, a computationally efficient variant of OPD, called sampled-token OPD, has attracted substantial attention due to its effectiveness in fine-tuning both small-scale models \citep{lu2025onpolicydistillation} and large-scale frontier models \citep{yang2025qwen3, team2026kimi}. It improves the student model by receiving token-level supervision from the teacher model, yielding a different learning paradigm from classic RLVR algorithms, such as Group Relative Policy Optimization (GRPO; \citet{shao2024deepseekmath}) and Decoupled Clip
and Dynamic sAmpling Policy Optimization (DAPO; \citet{yu2026dapo}).
In particular, GRPO and DAPO fine-tune models based on external reward signals for domains with automatically verifiable outcomes, including mathematics \citep{AIME24, yu2026dapo} and coding \citep{chen2021evaluating}. Their optimization objective has a clear interpretation: policy updates reinforce reasoning trajectories that lead to correct solutions while suppressing those that do not, thereby directly improving the model's reasoning capability. Moreover, the advantage assigned to each token admits a natural credit assignment interpretation, quantifying its contribution to the final outcome. 
In stark contrast, sampled-token OPD assigns each token a reward that reflects teacher preference, rather than quantifying its contribution to the correctness of the final answer. 
This token-level dense supervision has been considered one of the key advantages of OPD \citep{lu2025onpolicydistillation, li2026rethinking, xiao2026mimo}, but also makes it difficult to precisely understand how OPD improves the student's reasoning capability. To better understand the mechanism of OPD, we systematically sparsify its token-level supervision and examine how OPD performs under increasingly sparse supervision. Our empirical results discover a counter-intuitive phenomenon: supervising only a few thousand tokens can induce substantial parameter changes and significant improvements in reasoning capability, challenging the conventional assumption that dense token-level supervision is necessary for reasoning improvement.

We summarize our main findings and contributions as follows:

\begin{itemize}
    \item Using Qwen3 series models on math reasoning benchmarks as a testbed, across nine teacher-student combinations spanning large-scale teacher/small-scale student, same-scale teacher-student, and small-scale teacher/large-scale student settings, we consistently observe that supervising only one randomly selected token per trajectory consistently improves the reasoning capacity of the student model. Further, selectively supervising only one or two tokens with extreme teacher-student probability difference can match and in most cases outperform standard OPD with dense signal.
    \item The extremely sparse supervision also provides a useful lens for investigating and understanding OPD by isolating different components of the token-level learning signal and examine their individual contributions. 
    First, we reveal a non-monotonic relationship between the amount of supervision and reasoning performance: as supervision becomes increasingly sparse, performance initially degrades but can recover and even surpass dense OPD under extreme sparsity. Second, an expressive student, as in the same-scale teacher--student and large-scale student/small-scale teacher settings, benefits more from tokens with positive rewards, whereas a less expressive student in the large-scale teacher/small-scale student setting benefits more from tokens with negative rewards. 
    \item Although standard OPD is formulated as minimizing the teacher--student reverse KL divergence, the sparse OPD variants that achieve the best reasoning performance do not necessarily reduce this divergence. In fact, we observe that they can even increase the teacher--student reverse KL divergence, suggesting that the reasoning improvement induced by OPD cannot be explained solely by making the student distribution closer to the teacher distribution.
    \item We further validate the extremely sparse supervision phenomenon beyond the Qwen3--OPD--Math setting through cross-task, cross-family, and cross-algorithm experiments. Specifically, experiments on coding reasoning demonstrate that the phenomenon extends beyond mathematical reasoning; experiments with Llama models provide evidence that it is not specific to the Qwen3 family; and experiments with PPO demonstrate that extremely sparse supervision can induce meaningful reasoning improvements beyond OPD, even in settings where dense supervision fails.
\end{itemize}

More broadly, our results provide empirical evidence that post-training does not necessarily need to be token-intensive to substantially improve reasoning capability. A surprisingly small amount of token-level supervision can produce improvements comparable to, and sometimes larger than, dense supervision. This finding opens a new direction for understanding and designing post-training, motivating a shift toward more token- and memory-efficient training paradigms.

\paragraph{Notations:} Denote a large language model (LLM) with parameters $\theta$ by $\pi_\theta$. Given a prompt $\bx$, the model generates a response sequence $\by = (y_1, \ldots, y_T)$ according to $\by \sim \pi_\theta(\cdot | \bx)$. Specifically, the sequence is generated autoregressively, where at each step $t$, conditioned on the prompt $\bx$ and the previously generated prefix $\by_{1:t-1}$, the next token is sampled as $y_t \sim \pi_\theta(\cdot |\bx, \by_{1:t-1})$. Denote the vocabulary of the tokenizer as $\cV$.  For any two distributions $p$ and $q$ over $\cV$, define the Kullback–Leibler (KL) divergence between $p$ and $q$ as $\text{KL}(p||q) = \sum_{x\in\cV}p(x)\log(p(x)/q(x))$.

\section{Related Work}

Post-training has become a central stage for improving the capabilities and behavior of LLMs, playing an important role in both instruction following and reasoning. Supervised fine-tuning (SFT, \cite{wei2021finetuned}) adapts pretrained models using curated demonstrations, while reinforcement learning from human feedback (RLHF) further optimizes model behavior according to human preferences \citep{ouyang2022training,bai2022training}. Reinforcement learning from verifiable rewards (RLVR) has emerged as an effective approach for eliciting reasoning capabilities on tasks with automatically verifiable outcomes, such as mathematics and coding \citep{shao2024deepseekmath, guo2025deepseek, yu2026dapo}. In RLVR, models generate responses on-policy and receive rewards based on the correctness of the resulting solutions, with policy optimization reinforcing trajectories that lead to desirable outcomes while suppressing unsuccessful ones.
More recently, on-policy distillation (OPD) provides a complementary paradigm for post-training. Rather than relying on a scalar outcome reward, OPD generates trajectories from the student model and obtains token-level supervision from a stronger teacher model \citep{agarwal2024policy, lu2025onpolicydistillation}. 
This dense token-level supervision is one of the defining advantages of OPD and has contributed to its recent adoption for reasoning-oriented post-training \citep{yang2025qwen3, team2026kimi}.

Despite differences in their learning objectives, these post-training paradigms share an important characteristic: training involves large numbers of generated tokens. In particular, online RL requires generating long reasoning trajectories, and a large collection of generated tokens contributes to policy optimization. The substantial number of token-level learning signals involved in post-training has motivated recent work to investigate whether all generated tokens are equally useful for optimization. In RLVR, \cite{wang2026beyond} show that the learning signal is highly non-uniform across tokens: training with only a minority of high-entropy tokens can outperform training with all tokens. \cite{xu2026tip} study token importance in OPD and investigate how to identify tokens that carry particularly useful learning signals. These works suggest that selective use of token-level supervision can improve training effectiveness while discarding a substantial portion of the available tokens.  Recent works therefore increasingly ask which tokens should contribute to the learning objective.
Our work is motivated by this line of studies but is fundamentally orthogonal to them.
Rather than identifying which tokens are most informative, we discover a counter-intuitive phenomenon that has not been previously reported in literature: extremely sparse supervision, down to one token per trajectory, can effectively incentivize reasoning ability.

\section{Preliminary}

In this section, we introduce the formulation of on-policy distillation and provide the necessary background and motivation for its use. 

\subsection{Supervised Fine-Tuning as Sampled-Token Off-Policy Distillation}

Knowledge distillation (KD, \citet{hinton2015distilling}) transfers knowledge from a teacher model $\pi^{\text{teacher}}$ to a student model $\pi_\theta$ by minimizing the divergence between their output distributions. For autoregressive language models, KD minimizes the token-level full-vocabulary \textit{forward} KL divergence  
$$\mathcal{L}_{\mathrm{KD}}(\theta)
=
\mathbb{E}_{\bx\sim\mathcal{D},\by\sim\pi^{\text{teacher}}(\cdot|\bx)}\bigg[
\sum_{t=1}^{T}
D_{\mathrm{KL}}
\big(
\pi^{\text{teacher}}(\cdot|\bx,\by_{1:t-1})
\|
\pi_\theta(\cdot|\bx,\by_{1:t-1})
\big)\bigg],$$
where $\cD$ is a dataset collected by the teacher model $\pi^{\text{teacher}}$, thus KD is also known as the off-policy distillation. Replacing the full-vocabulary KL divergence in $\mathcal{L}_{\mathrm{KD}}(\theta)$ by an unbiased estimator using teacher-sampled tokens, $\log \pi^{\text{teacher}}(y_t|\bx, \by_{1:t-1}) - \log \pi_\theta(y_t|\bx, \by_{1:t-1})$, and removing the first term that is independent of $\theta$, we obtain the classic supervised fine-tuning (SFT, \cite{wei2021finetuned}) objective: 
$$
\mathcal{L}_{\mathrm{SFT}}(\theta)
=
-\mathbb{E}_{\bx\sim\mathcal{D},\by\sim\pi^{\text{teacher}}(\cdot|\bx)}
\bigg[
\sum_{t=1}^{T}
\log
\pi_\theta(y_t|\bx,\by_{1:t-1})
\bigg].
$$
SFT can therefore be viewed as off-policy distillation with sampled tokens, which bypasses the computation burden of obtaining the full-vocabulary teacher distribution at each token.

\subsection{Sampled-Token On-Policy Distillation}

In forward KL distillation, the student model is optimized on prefixes sampled from the teacher distribution, and not on those induced by the student model. During inference, the student may generate prefixes that are never observed during training and thus fail to reason reliably. This is the classic compounding error issue in sequence generation tasks \citep{ross2011reduction, bengio2015scheduled} caused by the distribution mismatch between training and inference. On-policy distillation (OPD, \citet{agarwal2024policy, gu2024minillm, yang2025qwen3}) addresses this issue by sampling trajectories from the 
student model $\pi_\theta$, allowing the teacher to provide supervision signals on prefixes that 
are actually visited by the student. Specifically, 
OPD minimizes the following \textit{reverse} KL objective function:
$$
\mathcal{L}_{\mathrm{OPD}}(\theta)
=
\mathbb{E}_{\bx\sim\mathcal{D},\by\sim\pi_\theta(\cdot|\bx)}
\bigg[
\sum_{t=1}^{T}
D_{\mathrm{KL}}
\big(
\pi_\theta(\cdot|\bx,\by_{1:t-1})
\|
\pi^{\text{teacher}}(\cdot|\bx,\by_{1:t-1})
\big)
\bigg],
$$
where the tokens are generated on-policy by the student model, i.e., $\by\sim\pi_\theta(\cdot|\bx)$.
An efficient and prevailing variant of OPD proposed by
\citet{lu2025onpolicydistillation} replaces the full-vocabulary inverse KL divergence at each step $t$ by an unbiased estimator, $
\log\pi_\theta(y_t|\bx,\by_{1:t-1})-\log\pi^{\text{teacher}}(y_t|\bx,\by_{1:t-1})$, using the student-sampled token $y_t$. The objective function of this sampled-token OPD variant is defined as 
\begin{align}
\label{eq:opd objective}
    \mathcal{L}(\theta)=-\mathbb{E}_{\bx\sim\mathcal{D},\by\sim\pi_\theta(\cdot|\bx)}\bigg[\sum_{t=1}^Tr_t(\bx, \by_{1:t-1})\bigg], ~ r_t(\bx, \by_{1:t-1}) = \log\pi^{\text{teacher}}(y_t|\bx,\by_{1:t-1})
-
\log\pi_\theta(y_t|\bx,\by_{1:t-1}).
\end{align}
Given a prompt $\bx$ and a prefix $\by_{1:t-1}$ at step $t$, a positive reward $r_t(\bx, \by_{1:t-1})$ indicates that the teacher assigns a higher probability to the sampled token $y_t$ than the student, thereby encouraging the student to increase the likelihood of generating that token. Conversely, a negative reward indicates that the teacher assigns a lower probability to $y_t$ than the student, discouraging the student from generating it.
We refer to the sampled-token variant of OPD simply as OPD unless otherwise specified. 
OPD has recently been adopted in the post-training of frontier large language models \citep{xiao2026mimo, team2026kimi}, where it proves remarkably effective. OPD naturally admits an online RL formulation, where trajectories are generated on-policy by the student, and token-level teacher supervision serves as a dense reward signal.
From a theoretical perspective, the optimal policy that maximizes the OPD objective \eqref{eq:opd objective} is exactly the teacher model $\pi^{\text{teacher}}$. Therefore, OPD improves the student model by directly optimizing it to imitate the teacher model's behavior.

\section{Extremely Sparse Supervision Incentivizes Reasoning Ability}
\label{sec:Extremely Sparse Supervision Incentivizes Reasoning Ability}

Token-level dense supervision has been regarded as one of the key advantages of OPD \citep{lu2025onpolicydistillation, li2026rethinking, xiao2026mimo}. A growing body of recent works \citep{yang2025qwen3, lu2025onpolicydistillation, li2026rethinking, yang2026learning, xu2026tip} has demonstrated the effectiveness of OPD and its variants on the Qwen3 family \citep{yang2025qwen3}, making it a natural testbed for studying what drives the effectiveness of OPD. This raises a fundamental question: is the effectiveness of OPD indeed driven by its dense token-level supervision? In this section, we systematically investigate this question by sparsifying the token-level supervision. Focusing on Qwen3 models and mathematical reasoning tasks, we uncover a surprising phenomenon: even extremely sparse supervision can effectively improve the reasoning capability of the student model.

\subsection{Randomly Supervising One Token per Trajectory}
As a warm-up, we consider an extreme case of sparse OPD, where only one randomly selected token from each generated trajectory is supervised and contributes to the gradient calculation and parameter update.

\paragraph{Sparse OPD.}
Define the token-level advantage $A_t = \log \pi^{\text{teacher}}(y_t|\bx, \by_{1:t-1}) - \log \pi_\theta (y_t|\bx, \by_{1:t-1})$, and
the PPO-style $\epsilon$-clipped per-token loss $\ell_t(\theta)=-\min(
       w_t(\theta) A_t,
       \mathrm{clip}(w_t(\theta),1-\epsilon,1+\epsilon) A_t
    )$, where $w_t(\theta) = \pi_\theta(y_t|\bx, \by_{1:t-1})/ \pi_{\theta_{\mathrm{old}}}(y_t|\bx, \by_{1:t-1})$ is the importance ratio.
The OPD objective is the token-mean loss over all response tokens,
\begin{align*}
    \mathcal{L}_{\mathrm{OPD}}(\theta) =
    \mathbb{E}_{\bx\sim\mathcal{D}, \by\sim\pi_{\theta_{\mathrm{old}}}(\cdot|\bx)}
    \bigg[
       {\frac{1}{|\by|}\sum_{t=1}^{|\by|} \ell_t(\theta)}
    \bigg].
\end{align*}
We insert a mask $m_t$ to exclude the majority of tokens from the OPD objective 
\begin{align}
\label{eq:sparse opd obj}
    \mathcal{L}_{\mathrm{SOPD}}(\theta) =
    \mathbb{E}_{\bx\sim\mathcal{D}, \by\sim\pi_{\theta_{\mathrm{old}}}(\cdot|\bx)}
    \bigg[
       {\frac{1}{|\by|}\sum_{t=1}^{|\by|} \ell_t(\theta)\cdot m_t}
    \bigg].
\end{align}
We denote any algorithm that minimizes  \eqref{eq:sparse opd obj} with sparse masks as the \textbf{sparse OPD} algorithm. We can define different types of masks and obtain various sparse OPD variants.
As a warm-up experiment, we uniformly at random select  exactly \textit{one} token $t$ from each response $\by$ and set $m_t=1, m_{t'}=0$ for any $t'\neq t$. We denote this sparse OPD variant with only one token supervision per trajectory as \textit{rand1tok}. 

\paragraph{Models and Datasets.} 
We conduct experiments using models from the Qwen3 family~\citep{yang2025qwen3}. For the teacher models, we first train the Qwen3-4B-Base model with GRPO for one and five epochs, obtaining \textbf{Qwen3-4B-GRPO-1ep} and \textbf{Qwen3-4B-GRPO-5ep}, respectively. We additionally include \textbf{Qwen3-4B-Instruct-2507} and \textbf{Qwen3-30B-A3B-Instruct-2507} as two off-the-shelf teacher models. For the student models, we consider \textbf{Qwen3-1.7B-Base}, \textbf{Qwen3-4B-Base}, \textbf{Qwen3-1.7B}, \textbf{Qwen3-4B}, and \textbf{Qwen3-8B}. Unless otherwise specified, both teacher and student models operate in \emph{no-think} mode.
Using these models, we construct nine teacher--student families (\Cref{tab:families-overview}) spanning three representative distillation regimes: \emph{large-scale teacher/small-scale student}, \emph{same-scale teacher and student}, and \emph{small-scale teacher/large-scale student}. These regimes cover varying degrees of teacher--student capability gaps and student representation capacity, enabling us to systematically evaluate the effectiveness of sparse supervision across diverse distillation settings. For both GRPO and OPD training, we use DAPO-Math-17K \citep{yu2026dapo} dataset. Overall, our setup follows a standard reasoning-oriented RLVR and OPD setting using mathematical reasoning data.  We provide the detailed training configuration in \Cref{sec:Experiment Configuration}, \Cref{tab:setup-hyperparams}. 

\paragraph{Evaluation Metrics and Benchmarks.}
We are interested in measuring the \textbf{reasoning ability boundary} of the student models fine-tuned by the OPD and sparse OPD. Following the evaluation protocol in the literature \citep{chen2026does, zhu2026surprising}, we adopt the unbiased low-variance estimator of $\mathrm{pass@}k$ proposed by \citet{chen2021evaluating}, which measures whether the model can produce a correct answer within $k$ independent attempts, thereby reflecting the model's reasoning ability boundary. Specifically, for each prompt $\bx_i$ in an evaluation dataset $\cD$, we generate $n$ responses $\{\by_j\}_{j=1}^n$ and count the number of correct responses as $c_i$. 
Then the unbiased estimator of $\mathrm{pass@}k$ on dataset $\cD$ is defined as
\begin{align*}
\mathrm{pass@}k
=
\frac{1}{|\cD|}\sum_{i=1}^{|\cD|}
\Bigg[
1
-
\frac{\binom{n-c_i}{k}}
{\binom{n}{k}}
\Bigg].
\end{align*}
We calculate a spectrum of $\mathrm{pass@}k$ with $k\in\{1,2,4,8,\cdots,128\}$ using $n=256$. For each $\mathrm{pass@}k$ estimator, we also provide the $95\%$ confidence interval (CI) to account for the randomness induced by finite (n=256) sampling. In particular, $ \mathrm{Var}(\mathrm{pass@}k) = \frac{1}{|\cD|^2}\sum_{i=1}^{|\cD|} \mathrm{Var}(g_k(c_i, n))$,
where $g_k(c_i,n) = 1- \binom{n-c_i}{k}/\binom{n}{k}$. Denote the  probability of generating a correct response for $\bx_i$ as $p_i$, we have $c_i \sim \text{Binomial}(n,p_i)$. Replacing the unknown $p_i$ by $\hat{p}_i=c_i/n$, we have 
\begin{align*}
    \widehat{\mathrm{Var}}(g_k(c_i,n)) = \sum_{j=0}^{n} g^2_k(j, n)\cdot \text{Binomial}(j; n, \hat{p}_i) - \Big(\sum_{j=0}^{n} g_k(j, n)\cdot \text{Binomial}(j; n, \hat{p}_i)\Big)^2,
\end{align*}
where
$\operatorname{Binomial}(j;n,\hat{p}_i)
=
\binom{n}{j}\hat{p}_i^j(1-\hat{p}_i)^{n-j}$.
Inserting the variance estimator to $\mathrm{Var}(\mathrm{pass@}k)$, we have 
$\hat{\sigma} = \sqrt{\frac{1}{|\cD|^2}\sum_{i=1}^{|\cD|}\widehat{\mathrm{Var}}(g_k(c_i,n))}$ and the CI is calculated as $\mathrm{pass@}k \pm 1.96\cdot\hat{\sigma}$. We report $\mathrm{pass@}k$ of the teacher model, base student model and student models fined-tuned by plain OPD and different variants of sparse OPD on AIME 24 \citep{AIME24} and AIME 25 \citep{AIME25}.

Another metric widely adopted in the literature is $\mathrm{avg@}k$, which measures the  \textbf{efficiency in sampling correct responses}. We adopt $\mathrm{avg}@$8 in this paper: for each prompt $\bx_i$, the model generates eight responses, among which $c_i$ are correct. The $\mathrm{avg}@$8 score is then computed as $\mathrm{avg}@8 = \frac{1}{|\cD|}\sum_{i=1}^{|\cD|}{c_i}/{8}$. We report $\mathrm{avg}@$8 on three math benchmarks, AIME 24, AIME 25 and HMMT-Feb 25 \citep{balunovic2026matharena}, as well as the mean $\mathrm{avg}@$8 over the three benchmarks.

\begin{table}[t]
\centering
\small
\setlength{\tabcolsep}{3pt}
\caption{The nine OPD families: student $\leftarrow$ teacher combinations.}
\label{tab:families-overview}
\begin{tabular}{c l l c c}
\toprule
Family & Student & Teacher & Regime & Results \\
\midrule
1 & Qwen3-1.7B-Base & Qwen3-4B-GRPO-1ep
  & large teacher/small student & Table~\ref{tab:family1-tokenmask}, Figure~\ref{fig:family1_passk_combined} \\
2 & Qwen3-1.7B-Base & Qwen3-4B-GRPO-5ep
  & large teacher/small student & Table~\ref{tab:family2-tokenmask}, Figure~\ref{fig:family2_passk_combined} \\
3 & Qwen3-4B-Base & Qwen3-4B-GRPO-1ep
  & same-scale teacher student & Table~\ref{tab:family3-tokenmask}, Figure~\ref{fig:family3_passk_combined} \\
4 & Qwen3-4B-Base & Qwen3-4B-GRPO-5ep
  & same-scale teacher  student & Table~\ref{tab:family4-tokenmask-rep}, Figure~\ref{fig:family4_passk_combined-rep} \\
5 & Qwen3-1.7B & Qwen3-4B-Instruct-2507
  & large teacher/small student & Table~\ref{tab:family5-tokenmask}, Figure~\ref{fig:family5_passk_combined} \\
6 & Qwen3-8B & Qwen3-30B-A3B-Instruct-2507
  & large teacher/small student & Table~\ref{tab:family6-tokenmask}, Figure~\ref{fig:family6_passk_combined} \\
7 & Qwen3-4B & Qwen3-30B-A3B-Instruct-2507
  & large teacher/small student & Table~\ref{tab:family7-tokenmask}, Figure~\ref{fig:family7_passk_combined} \\
8 & Qwen3-1.7B & Qwen3-30B-A3B-Instruct-2507
  & large teacher/small student & Table~\ref{tab:family8-tokenmask-rep}, Figure~\ref{fig:family8_passk_combined-rep} \\
9 & Qwen3-8B & Qwen3-4B-Instruct-2507
  & small teacher/large student & Table~\ref{tab:family9-tokenmask-rep}, Figure~\ref{fig:family9_passk_combined-rep} \\
\bottomrule
\end{tabular}
\end{table}

\begin{table}[t]
\centering
\small
\caption{A summary of all sparse OPD variants and the number of tokens supervised per trajectory.}
\label{tab:tokenmask-activation}
\begin{tabular}{l l}
\toprule
Variant & Number of tokens supervised per trajectory   \\
\midrule
plain OPD       & all tokens (up to 8192 tokens per trajectory)\\
randmask 0.1\%  & $0.1\%$ of tokens (uniformly at random) \\
pctltail 0.05\% & $0.1\%$ of tokens (bottom and top $0.05\%$ of tokens by OPD reward) \\
rand1tok        & one token (uniformly at random) \\
minmaxtok       & two tokens (tokens with the highest and lowest reward) \\
mintok          & one token (token with the highest reward) \\
maxtok          & one token (token with the lowest reward) \\
\bottomrule
\end{tabular}
\end{table}

\paragraph{Experiment Results: \textit{rand1tok} Effectively Enables Reasoning Improvement.}

To investigate whether the sparse OPD variant \textit{rand1tok} improves the base student, we compare the performance of student model fine-tuned by \textit{rand1tok} with (1) the base student model; (2) the student model fine-tuned with plain OPD, as well as (3) the teacher model.
We present the evaluation results of the teacher model, base student model, student model after \textit{rand1tok} training, and student model after the plain OPD training. We select Family 8 (\Cref{fig:family8_passk_combined} and \Cref{tab:family8-tokenmask}), Family 4 (\Cref{fig:family4_passk_combined} and \Cref{tab:family4-tokenmask}) and Family 9 (\Cref{fig:family9_passk_combined} and \Cref{tab:family9-tokenmask}) as representative cases of the large-scale teacher/small-scale student setting, same-scale teacher--student setting, and small-scale teacher/large-scale student setting, respectively. The complete results across all nine families are provided in \Cref{sec:full experiment results}.

Recall that \textit{rand1tok} discards almost all of the token-level supervision used by plain OPD: a trajectory containing thousands of tokens provides supervision through only a single randomly-selected token. One might expect OPD to completely fail to transfer the teacher's capabilities under such an extreme reduction in supervision.
However, the pass@$k$ results reveal the opposite:
\textbf{\textit{rand1tok} consistently improves the reasoning capability of base students across all nine teacher--student families}. Moreover, in Families 1--3, \textit{rand1tok} surpasses plain OPD in the large-$k$ regime of pass@$k=128$, suggesting that extreme sparsification can sometimes lead to broader reasoning exploration. 
The avg@$8$ results 
further demonstrate that \textbf{\textit{rand1tok} consistently improves sampling efficiency of base students across all families}. The consistent improvement across both pass@$k$ and avg@$8$ metrics, as well as nine diverse teacher--student configurations, indicates that this phenomenon is not an artifact of evaluation variance. Instead, these results reveal that dense token-level supervision, despite being a central design choice of OPD, is not necessary for transferring reasoning capability, as one randomly selected supervised token per trajectory is sufficient to induce substantial parameter changes and reasoning improvement.

\section{Extreme Sparse Supervision Can Outperform Dense OPD}
The surprising effectiveness of \textit{rand1tok} motivates us to further investigate the extremely sparse supervision in OPD. Random token selection provides an approximately unbiased estimator (up to scale) of the dense OPD objective, suggesting that \textit{rand1tok} still optimizes the same underlying teacher--student alignment objective, albeit with substantially higher variance. However, the success of random sparse supervision raises several deeper questions: \textit{Is dense token-level supervision necessary for improving reasoning ability?
Does the reasoning ability improvement of the student model arise purely from better mimicking the teacher model? How different token-level signals play distinct roles in shaping the student's reasoning capability?}
In this section, we investigate sparse OPD variants that selectively supervise tokens with extreme teacher--student probability differences, and provide empirical answers to the above questions.

\subsection{Tokens with Extreme OPD Rewards}
\paragraph{Supervision on Tokens with Extreme OPD Rewards.}
We focus on the tokens with the most positive and most negative OPD rewards, as well as their combination within each trajectory, and investigate whether selecting tokens with extreme rewards provides more effective sparse supervision than random token selection. Specifically, if the token mask in \eqref{eq:opd objective} is defined as 
\[
m_t=\ind\big\{t=\arg\max_i r_i\big\},
\]
then only the token with the maximum reward contributes to the gradient update. We refer to this sparse OPD variant as \textit{maxtok}. Such tokens typically receive large positive OPD rewards (e.g., $r_t > 3$), indicating that the teacher assigns substantially higher probability to the sampled token than the student. Empirically, these tokens are usually associated with high student entropy.

Conversely, if the token mask in \eqref{eq:opd objective} is defined as 
\[
m_t=\ind\big\{t=\arg\min_i r_i\big\},
\]
then only the token with the minimum reward contributes to the gradient update. We refer to this variant as \textit{mintok}. These tokens typically have an extremely negative reward (e.g., $r_t<-16$), indicating that the teacher assigns exponentially lower probability to the sampled token than the student. Unlike \textit{maxtok}, the student entropy on these tokens can be either high or low. 

Next we combine the two selection strategies by retaining both the maximum and minimum reward tokens in each trajectory:
\[
m_t=\ind\big\{t=\arg\max_i r_i \;\lor\; t=\arg\min_i r_i\big\}.
\]
We refer to this sparse OPD variant as \textit{minmaxtok}.

Finally, for ablations we slightly increase the supervision budget from one token per trajectory to 0.1\% of the generated tokens, corresponding to approximately 3--4 supervised tokens per trajectory. Specifically, the token mask $m_t$ in \eqref{eq:opd objective} is independently sampled from a Bernoulli distribution with success probability $p=0.001$, i.e., $P(m_t=1)=0.001$ and $P(m_t=0)=0.999$. We refer to this sparse OPD variant as \textit{randmask 0.1\%}. Moreover, let $q_{0.05\%}$ and $q_{99.95\%}$ denote the 0.05th and 99.95th percentiles of the token rewards within a trajectory. We define
\[
m_t=\ind\left\{r_t \le q_{0.05\%}\;\lor\; r_t \ge q_{99.95\%}\right\},
\]
and refer to this sparse OPD variant as \textit{pctltail 0.05\%}. 

\paragraph{Hypothesis.}
An intuitive motivation for supervising one or two tokens with extreme OPD rewards comes from the natural learning process: a learner first attempts to solve a problem, then reflects on a few consequential reasoning steps, incorporates the resulting feedback into its existing knowledge, and improves through subsequent trial and error. In contrast, standard dense-token supervision on every token may be highly redundant or even harmful. In an on-policy trajectory, later tokens are conditioned on the particular prefix produced by the current student. Some of these prefixes may result from an early mistake and may never occur again after the model parameters are updated. Consequently, applying dense supervision to every subsequent token may correct behaviors conditioned on states that are unlikely to be encountered after the parameter update, making such supervision largely ineffective or even redundant. In contrast, targeted interventions, such as correcting one particularly important error or reinforcing one useful behavior per response, may be sufficient to shift the student's future behavior towards generating correct answer, especially given student model's substantial pretrained knowledge and existing reasoning capabilities. Moreover, since the teacher model is not perfect, sparse supervision may also reduce the risk of unnecessarily transferring the teacher's mistakes or limitations.

\subsection{Experiment Results}
We introduce an additional evaluation metric in the following analysis. Standard OPD aims to minimize the reverse KL divergence between student and teacher. We are interested in how different variants of sparse OPD change the reverse KL. Given a model $\pi_\theta$, we let $\pi_\theta$ generate one response for each prompt $\bx_i$ in a dataset $\cD$ and calculate the full-vocabulary reverse KL as 
$$\mathrm{revKL}(\pi_\theta, \pi^{\text{teacher}}) = \frac{1}{|\cD|}\sum_{i=1}^{|\cD|}\sum_{t=1}^{|\by_i|}\mathrm{KL}(\pi_\theta(\cdot|\bx_i, \by_{i, 1:t-1})||\pi^{\text{teacher}}(\cdot|\bx_i, \by_{i, 1:t-1})).$$
We set $\cD = \{\text{AIME 24}, \text{AIME 25}, \text{HMMT-Feb 25}\}$.  In the following we provide a thorough analysis of the experiment results.
\begin{figure}
    \centering
    \includegraphics[width=\linewidth]{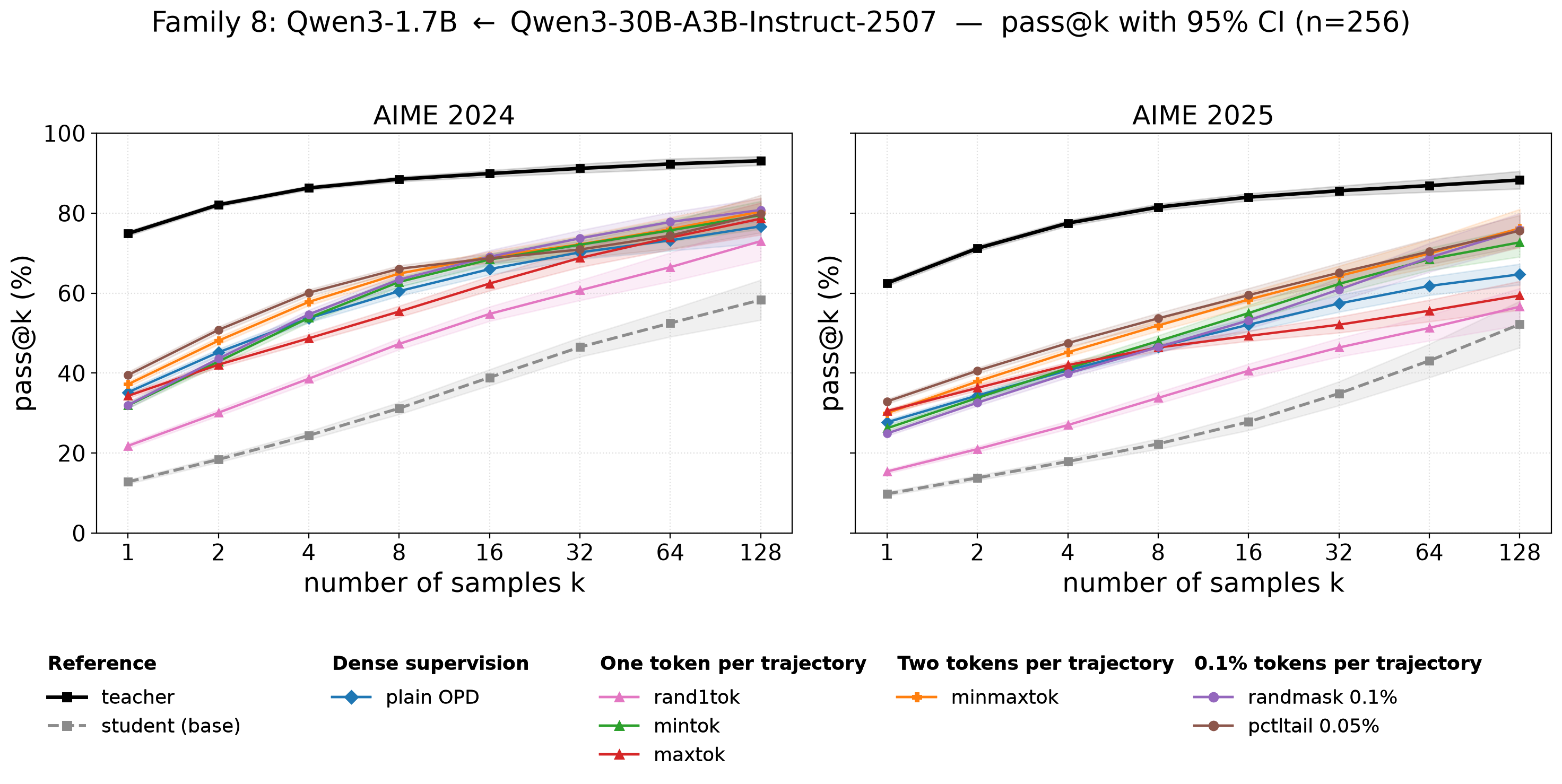}
    \caption{Family 8 pass@k curves of base student model, teacher model, student model trained with plain OPD and student models trained with sparse OPD variants across AIME 24 and AIME 25. Shaded area is the 95\% confidence interval.}
    \label{fig:family8_passk_combined}
\end{figure}

\begin{figure}
    \centering
    \includegraphics[width=\linewidth]{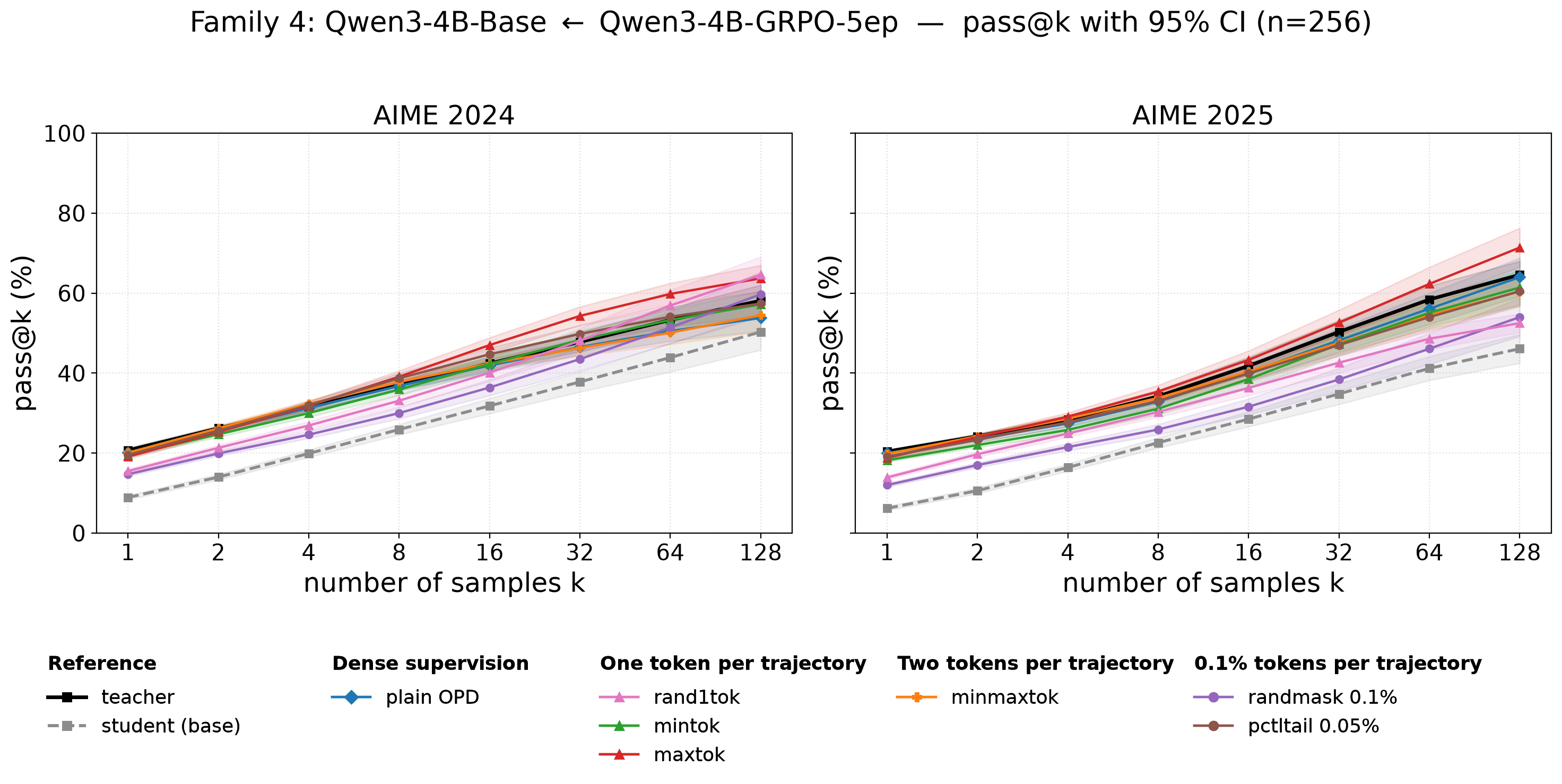}
    \caption{Family 4 pass@k curves of base student model, teacher model, student model trained with plain OPD and student models trained with sparse OPD variants across AIME 24 and AIME 25. Shaded area is the 95\% confidence interval.}
    \label{fig:family4_passk_combined}
\end{figure}

\begin{figure}
    \centering
    \includegraphics[width=\linewidth]{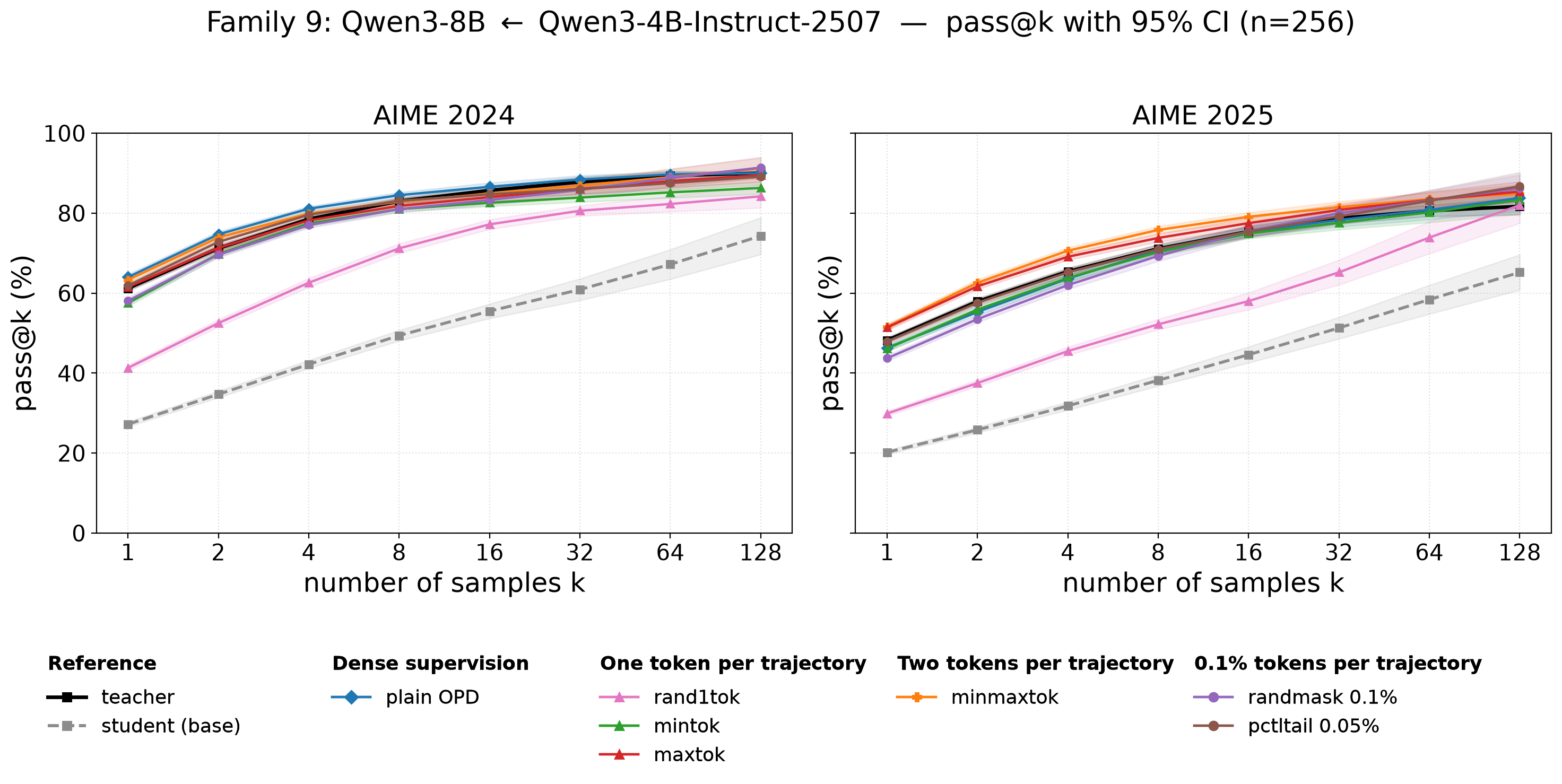}
    \caption{Family 9 pass@k curves of base student model, teacher model, student model trained with plain OPD and student models trained with sparse OPD variants across AIME 24 and AIME 25. Shaded area is the 95\% confidence interval.}
    \label{fig:family9_passk_combined}
\end{figure}

\begin{table}[t]
\centering
\small
\caption{Family 8: Qwen3-1.7B $\leftarrow$ Qwen3-30B-A3B-Instruct-2507. avg@8 scores of the base student model, teacher model, student model trained with plain OPD and student models trained with sparse OPD variants across AIME 2024, AIME 2025 and HMMT Feb 2025. Mean averages over AIME24/AIME25/HMMT.
\emph{keep\_frac (\%)}: fraction of response tokens supervised among all generated tokens; 
\emph{revKL}: reverse KL divergence between the student and teacher; \emph{freeze (\%)}: fraction of student network unchanged parameters after training.}
\label{tab:family8-tokenmask}
\begin{tabular}{l cccc c c c}
\toprule
 & \multicolumn{4}{c}{avg@8 (\%)} & & & \\
\cmidrule(lr){2-5}
Method & AIME24 & AIME25 & HMMT & Mean & keep\_frac (\%) & revKL & freeze (\%) \\
\midrule
Teacher & 72.5 & 62.1 & 42.5 & 59.0 & --    & 0.000 & --   \\
Student & 12.9 & 8.3  & 5.0  & 8.7  & --    & 0.368 & --   \\
plain OPD & 37.1 & 27.9 & 17.5 & 27.5 & 100     & 0.190 & 80.2 \\
\cmidrule(lr){1-8}
\multicolumn{8}{c}{\emph{one token supervision per trajectory}} \\[3pt]
rand1tok & 21.7 & 15.8 & 9.2  & 15.6 & 0.0379  & 0.299 & 95.0 \\
mintok & 32.9 & 30.0 & 17.5 & 26.8 & 0.0254  & 0.260 & 91.4 \\
maxtok & 35.8 & 32.1 & 19.2 & 29.0 & 0.0176  & 0.750 & 89.5 \\
\cmidrule(lr){1-8}
\multicolumn{8}{c}{\emph{two tokens supervision per trajectory}} \\[3pt]
minmaxtok & 40.4 & 29.6 & 16.7 & 28.9 & 0.0409  & 0.374 & 91.3 \\
\cmidrule(lr){1-8}
\multicolumn{8}{c}{\emph{0.1\% tokens supervision per trajectory}} \\[3pt]
randmask 0.1\% & 37.1 & 23.3 & 16.2 & 25.6 & 0.0988  & 0.267 & 92.6 \\
pctltail 0.05\% & 38.8 & 31.2 & 20.4 & \textbf{30.1} & 0.1380  & 0.404 & 89.6 \\
\bottomrule
\end{tabular}
\end{table}

\begin{table}[t]
\centering
\small
\caption{Family 4: Qwen3-4B-Base $\leftarrow$ Qwen3-4B-GRPO-5ep. avg@8 scores of the base student model, teacher model, student model trained with plain OPD and student models trained with sparse OPD variants across AIME 2024, AIME 2025 and HMMT Feb 2025. Mean averages over AIME24/AIME25/HMMT.
{keep\_frac (\%)}: fraction of response tokens supervised among all generated tokens; 
{revKL}: reverse KL divergence between the student and teacher; {freeze (\%)}: fraction of student network unchanged parameters after training.}
\label{tab:family4-tokenmask}
\begin{tabular}{l cccc c c c}
\toprule
 & \multicolumn{4}{c}{avg@8 (\%)} & & & \\
\cmidrule(lr){2-5}
Method & AIME24 & AIME25 & HMMT & Mean & keep\_frac (\%) & revKL & freeze (\%) \\
\midrule
Teacher & 20.8 & 17.9 & 8.8 & 15.8 & --    & 0.000 & --   \\
Student & 9.6  & 7.9  & 0.8 & 6.1  & --    & 0.441 & --   \\
plain OPD & 23.3 & 17.5 & 7.5 & 16.1 & 100     & 0.005 & 75.0 \\
\cmidrule(lr){1-8}
\multicolumn{8}{c}{\emph{one token supervision per trajectory}} \\[3pt]
rand1tok & 12.1 & 13.8 & 6.7 & 10.8 & 0.0601  & 0.078 & 90.2 \\
mintok & 19.2 & 17.5 & 6.2 & 14.3 & 0.0369  & 0.018 & 81.9 \\
maxtok & 22.9 & 18.8 & 7.5 & \textbf{16.4} & 0.0364  & 0.428 & 89.5 \\
\cmidrule(lr){1-8}
\multicolumn{8}{c}{\emph{two tokens supervision per trajectory}} \\[3pt]
minmaxtok & 21.7 & 19.6 & 5.0 & 15.4 & 0.0622  & 0.015 & 81.8 \\
\cmidrule(lr){1-8}
\multicolumn{8}{c}{\emph{0.1\% tokens supervision per trajectory}} \\[3pt]
randmask 0.1\% & 16.2 & 12.5 & 5.4 & 11.4 & 0.1000  & 0.119 & 88.6 \\
pctltail 0.05\% & 19.2 & 18.3 & 5.4 & 14.3 & 0.1770  & 0.012 & 80.9 \\
\bottomrule
\end{tabular}
\end{table}

\begin{table}[t]
\centering
\small
\caption{Family 9: Qwen3-8B $\leftarrow$ Qwen3-4B-Instruct-2507. avg@8 scores of the base student model, teacher model, student model trained with plain OPD and student models trained with sparse OPD variants across AIME 2024, AIME 2025 and HMMT Feb 2025. Mean averages over AIME24/AIME25/HMMT.
{keep\_frac (\%)}: fraction of response tokens supervised among all generated tokens; 
{revKL}: reverse KL divergence between the student and teacher; {freeze (\%)}: fraction of student network unchanged parameters after training.}
\label{tab:family9-tokenmask}
\begin{tabular}{l cccc c c c}
\toprule
 & \multicolumn{4}{c}{avg@8 (\%)} & & & \\
\cmidrule(lr){2-5}
Method & AIME24 & AIME25 & HMMT & Mean & keep\_frac (\%) & revKL & freeze (\%) \\
\midrule
Teacher & 62.1 & 48.3 & 30.0 & 46.8 & --      & 0.000 & --   \\
Student & 26.2 & 20.8 & 10.8 & 19.3 & --      & 0.268 & --   \\
plain OPD & 66.7 & 44.6 & 31.2 & 47.5 & 100     & 0.184 & 78.2 \\
\cmidrule(lr){1-8}
\multicolumn{8}{c}{\emph{one token supervision per trajectory}} \\[3pt]
rand1tok & 42.1 & 30.0 & 17.5 & 29.9 & 0.0392  & 0.253 & 95.5 \\
mintok & 55.4 & 47.1 & 25.8 & 42.8 & 0.0259  & 0.213 & 87.3 \\
maxtok & 63.8 & 50.8 & 29.6 & 48.1 & 0.0167  & 1.116 & 90.7 \\
\cmidrule(lr){1-8}
\multicolumn{8}{c}{\emph{two tokens supervision per trajectory}} \\[3pt]
minmaxtok & 62.9 & 55.0 & 30.0 & \textbf{49.3} & 0.0472  & 0.319 & 88.1 \\
\cmidrule(lr){1-8}
\multicolumn{8}{c}{\emph{0.1\% tokens supervision per trajectory}} \\[3pt]
randmask 0.1\% & 59.6 & 45.4 & 28.3 & 44.4 & 0.0999  & 0.279 & 94.1 \\
pctltail 0.05\% & 62.1 & 47.5 & 28.8 & 46.1 & 0.120   & 0.254 & 86.1 \\
\bottomrule
\end{tabular}
\end{table}

\paragraph{Supervising One or Two Tokens per Trajectory Is Sufficient to Match or Outperform Plain OPD.}
Selected experimental results are shown in \Cref{fig:family4_passk_combined,fig:family8_passk_combined,fig:family9_passk_combined} and \Cref{tab:family4-tokenmask,tab:family8-tokenmask,tab:family9-tokenmask}, and complete experiment results are postponed to \Cref{sec:full experiment results}. The key takeaway is that, across all nine teacher--student families, we can always identify a sparse OPD variant that matches (two out of nine families) or outperforms (seven out of nine families) plain OPD in terms of both reasoning capability boundary and sampling efficiency. In particular, \textit{minmaxtok} and \textit{pctltail 0.05\%} are consistently among the strongest variants, and the best-performing sparse OPD variant is typically one of the two.  
Moreover, \textit{randmask 0.1\%} is substantially more stable than \textit{rand1tok} and, in the large-pass@$k$ regime, consistently achieves performance comparable to plain OPD across most teacher--student families. However, its avg@$8$ performance remains below that of plain OPD, indicating that while randomly supervising only 0.1\% of generated tokens is sufficient to recover the reasoning capability improvement reflected by high pass@$k$, tokens with extreme rewards provide a noticeable advantage in sampling efficiency. Moreover, we observe that sparse OPD sometimes produce students that outperform their teachers, an effect that is particularly pronounced in Family 9, where the student has substantially greater representation capacity than the teacher.
Moreover, we note that \textit{maxtok} performs particularly well in the same-scale teacher--student setting (Families 3 and 4) and the small-scale teacher/large-scale student setting (Family 9), achieving strong improvements in both pass@$k$ and avg@$8$. 
In contrast, \textit{mintok} performs consistently well across all nine families. Although it falls slightly short of plain OPD in a few cases, it remains highly competitive despite supervising only one token per trajectory.
We postpone a detailed discussion on the effect of extreme-reward tokens in terms of student-teacher distribution gap and learning dynamics to the next section.

\paragraph{Sparse Supervision Does Not Simply Improve Teacher Imitation.}
According to the reverse KL divergence reported in \Cref{tab:family4-tokenmask,tab:family8-tokenmask,tab:family9-tokenmask}, sparse OPD variants do not necessarily reduce the teacher--student reverse KL divergence. On the contrary, many of the best-performing sparse OPD variants exhibit a larger reverse KL divergence than plain OPD. We highlight two particularly striking examples, Families 4 and 9, where \textit{maxtok} substantially increases the reverse KL divergence while producing students that significantly outperform plain OPD and even their teachers. More broadly, across the nine teacher--student families, the strongest reasoning performance is generally not achieved by the student with the smallest reverse KL divergence to the teacher. These observations suggest that the reasoning improvement induced by OPD cannot be fully explained by simply making the student distribution closer to the teacher distribution.

\begin{remark}
    We have empirically established an intriguing phenomenon: an extremely small amount of token-level supervision can match or even surpass dense supervision in plain OPD, despite discarding the vast majority of the available token-level training signal. The fact that this phenomenon emerges under multiple sparse supervision strategies, across diverse teacher--student configurations, calls into question whether the effectiveness of OPD is fundamentally driven by its dense token-level supervision. 
    We emphasize that our goal is neither to argue the learning signal in OPD necessarily originates from any particular type of token nor to advocate for a specific token-selection strategy. We also do not claim supervising one single token per trajectory represents the limit of achievable sparsity. For example, our preliminary experiments suggest that supervising only the minimum-reward token in \textit{incorrect} trajectories can achieve comparable performances. Moreover, the sparse OPD still requires the same number of on-policy rollouts as the plain OPD, which dominates the overall computational cost and training time. We leave a systematic investigation of even sparser or more targeted supervision schemes, as well as their potential for reducing overall training cost, to future work.
\end{remark} 

\paragraph{Sparse Subnetwork Update}
\cite{mukherjee2026reinforcement} discover that RLVR training updates only a small subnetwork of LLM comprising just 5\%-30\% of the parameters. We are interested in whether the standard OPD admits the same phenomenon and whether sparse OPD variants admit sparser subnetwork updates. As a quantification, we report two sparsity metrics: the fraction of activated tokens during the training (keep\_frac (\%)) and the fraction of parameters updated in the neural network (freeze (\%)). Specifically, given a base student model $\pi_{\theta^0}$ and an OPD fine-tuned student model $\pi_{\theta}$, the former is calculated as $\text{keep\_frac}\% = \frac{\#\text{activated tokens}}{\#\text{all tokens generated}}$ and the later is calculated as $\text{freeze} (\%) = 1 - \sum_{j=1}^N\ind\{|\theta^0_j - \theta_j| < 1e-5\}/N$, where $N$ is the dimension of $\theta$. 
The results in \Cref{tab:family4-tokenmask,tab:family8-tokenmask,tab:family9-tokenmask} show that standard OPD also updates only a small fraction of the student's parameters, consistent with the sparse-update phenomenon observed in RLVR. Furthermore, sparser token supervision generally results in sparser parameter updates. In general, we can conclude that across all nine teacher--student families, \textit{there exists at least one sparse OPD variant that supervises only $0.01\%$--$0.1\%$ of generated tokens and updates only $\sim10\%$ of the model parameters, yet matches or even outperforms plain OPD.}

\subsection{Ablation Studies and Additional Analysis}
Given the effectiveness of \textit{mintok} and \textit{maxtok}, we further sweep the amount of extremely negative and positive tokens used for supervision. Specifically, we set the token mask $m_t$ in \eqref{eq:sparse opd obj} to $m_t=\ind\{A_t < \tau\}$ or $m_t=\ind\{A_t > \tau\}$, where $\tau$ controls the fraction of the extreme tokens selected for supervision. We denote the resulting sparse OPD variants as \textit{\text{at} $<\tau$} and \textit{\text{at} $>\tau$}, respectively. Experiment results are postponed to \Cref{sec:full experiment results} (\Cref{fig:family5_threshold_passk_combined,fig:family6_threshold_passk_combined,fig:family7_threshold_passk_combined,fig:family8_threshold_passk_combined,fig:family9_threshold_passk_combined} and \Cref{tab:family1-threshold,tab:family2-threshold,tab:family3-threshold,tab:family4-threshold,tab:family5-threshold,tab:family6-threshold,tab:family7-threshold,tab:family8-threshold,tab:family9-threshold}). 

\paragraph{A Non-Monotonic Relationship Between Sparsity and Performance.}
Across our experiments, we frequently observe a non-monotonic relationship between the amount of supervised tokens and the reasoning performance. 
For example, \Cref{tab:family5-threshold,tab:family6-threshold,tab:family7-threshold} show that as the number of supervised tokens decreases from dense supervision, performance initially degrades, but then recovers and can even surpass dense OPD when the supervision becomes extremely sparse. However, when the supervision becomes too sparse, the learning signal eventually becomes insufficient to produce meaningful improvement. This suggests that reasoning performance can peak at an intermediate level of extreme sparsity, rather than varying monotonically with the amount of token-level supervision.

\paragraph{Why and How Extreme Token Supervision Works.}
One motivation for investigating tokens with extreme rewards is the mode-seeking property of reverse KL minimization \citep{jang2016variational, gu2024minillm}, particularly when the student distribution is less expressive than the teacher, as in the large-scale teacher/small-scale student setting. By penalizing probability mass assigned to regions where the teacher assigns low probability, reverse KL encourages the student to concentrate its probability mass on the teacher's high-probability modes rather than spreading it across multiple suboptimal alternatives. As illustrated in \Cref{fig:minimizing-revKL}, the optimal student distribution $q$ can assign moderately higher density than the target teacher distribution $p$ around the right mode.  
Thus upon converging, there should be few generated tokens with extremely negative or positive reward, and instead all generated tokens should receive moderate positive or negative rewards.
This suggests that during training, 
tokens with extremely negative or positive rewards provide strong directional learning signals, while tokens with moderate positive or negative rewards could largely be noise and should not contribute to the gradient update or be involved in shaping the student distribution.
There is, however, an important caveat for extremely positive rewards. When the student's representation capacity is limited, strongly reinforcing a single teacher-preferred mode may cause the student's probability mass to shift excessively toward different modes across training, potentially leading to unstable or conflicting updates. This may explain why \textit{maxtok} does not perform well in the large-scale teacher/small-scale student regime, including Qwen3-1.7B-Base (\Cref{fig:family1_passk_combined,fig:family2_passk_combined}), Qwen3-1.7B (\Cref{fig:family5_passk_combined,fig:family8_passk_combined}) and Qwen3-4B (\Cref{fig:family7_passk_combined}), whereas \textit{mintok} exhibits more stable performances.

In contrast, in the same-scale teacher--student and small-scale teacher/large-scale student settings, the student has substantially greater representation capacity and can better approximate the target distribution $p$, as illustrated on the right of \Cref{fig:minimizing-revKL}. In this regime, tokens with extreme rewards still provide particularly strong directional signals for optimization. Because the student has sufficient capacity to accommodate these updates, reinforcing such extreme signals need not force it to choose among incompatible modes. This interpretation is consistent with our results: in Families 3 (\Cref{fig:family3_passk_combined}), 4 (\Cref{fig:family4_passk_combined}), and 9 (\Cref{fig:family9_passk_combined}), \textit{maxtok} is consistently among the strongest sparse OPD variants and can even produce students that outperform their teachers.
Moreover, we observe distinct effects on the learning dynamics of actor entropy (\Cref{fig:actor-entropy}). \textit{mintok} tends to decrease actor entropy, consistent with the fact that supervision on tokens with extremely negative-reward suppresses probability mass on behaviors that the teacher strongly disfavors, inducing a \textit{spurious-mode pruning} effect. In contrast, \textit{maxtok} tends to increase actor entropy. Notably, tokens with extremely positive rewards are often associated with high student entropy. 
Such tokens correspond to the ``forking'' tokens \citep{wang2026beyond}, which are positions where the student model is uncertain about the current token choice, and the choice of this token strongly determines the subsequent generation path. Reinforcing these tokens therefore encourages the student to allocate probability to behaviors that it currently under-explores, inducing a \textit{mode recovery} effect.

\paragraph{Memory Saving and Response Length.}
A direct benefit of sparse supervision is memory saving: since only a small number of tokens contribute to the gradient calculation, we can discard the full-vocabulary logit information for the majority of masked tokens. Thus, for the sparse OPD variants the logit memory is effectively erased.
We defer a detailed discussion to \Cref{sec:Additional Results and Discussion}. 
Finally, we compare the average response length across all checkpoints. We observe no clear pattern that sparse OPD variants systematically increase or decrease response length relative to plain OPD, the response length of all variants remains within a reasonable range.

\paragraph{What Are Those Tokens?} We have demonstrated that tokens receiving extreme positive or negative rewards are particularly effective at incentivizing reasoning ability. It is therefore intriguing to ask what these tokens actually represent. Extremely sparse supervision offers a unique advantage for studying this question: with only two activated tokens per trajectory, we can directly inspect them by eye and characterize the types of decisions that receive these unusually strong learning signals. To this end, we audit the tokens activated by \textit{minmaxtok} among the first 20 steps in the sparse OPD setting with a Qwen3-1.7B student and a Qwen3-4B-Instruct-2507 teacher. We summarize the results in \Cref{tab:pos-token-buckets} and \Cref{tab:neg-token-buckets}. We find that extremely positive-reward tokens tend to have clear semantic meaning, whereas extremely negative-reward tokens are predominantly correct-but-teacher-dispreferred tokens, with only a small fraction corresponding to genuine mathematical errors.

\begin{table*}[t]
\centering
\small
\setlength{\tabcolsep}{6pt}
\renewcommand{\arraystretch}{1.35}

\begin{tabularx}{\textwidth}{
    >{\raggedright\arraybackslash}p{0.12\textwidth}
    >{\raggedright\arraybackslash}p{0.32\textwidth}
    >{\raggedright\arraybackslash}X
}
\toprule
\textbf{category} &
\textbf{description} &
\textbf{examples} \\
\midrule

\texttt{digit / value}
&
a numeral, or a single digit inside a larger number
&
\texttt{4, 9}, the \texttt{5} in \texttt{1<5>0},
the \texttt{7} in \texttt{2<7>2}, 
a variable-as-value like \texttt{a}, \texttt{x}, \texttt{k}
\\

\texttt{math-mode}
&
LaTeX delimiter or command opener --- switches into a formula
&
\texttt{\$}, \texttt{\$\$},
start of \texttt{\textbackslash frac}, \texttt{\textbackslash cdot},
\texttt{(}, \texttt{[},
\texttt{\{}, \texttt{\^{}}, \texttt{=},
 \texttt{sum}, \texttt{quad}
\\

\texttt{content word}
&
ordinary prose word (noun/verb/adjective) carrying lexical meaning
&
\texttt{determinant}, \texttt{function}, \texttt{symmetric},
\texttt{analyze}, \texttt{chooses}, \texttt{consider},
\texttt{rotates}, \texttt{circular}, \texttt{horizontal},
\texttt{valid}, \texttt{constraint}, 
\texttt{smaller}, \texttt{teams}, \texttt{axis}
\\

\texttt{discourse hedge}
&
reasoning-flow / stance marker --- steers, transitions, or backtracks
&
\texttt{But}, \texttt{Wait}, \texttt{Actually},
\texttt{So}, \texttt{Therefore}, \texttt{Thus}, \texttt{Since},
\texttt{Now}, \texttt{Let}, \texttt{Try},
\texttt{Note}, \texttt{Check}, \texttt{First}, \texttt{Alternatively}
\\

\texttt{format separator}
&
whitespace / markdown / structural punctuation --- organizes layout, no meaning
&
\texttt{space}, \texttt{\textbackslash n},
\texttt{\textbackslash n\textbackslash n}, \texttt{---},
\texttt{\#\#\#}, \texttt{\#\#\#\#}, \texttt{**},
\texttt{:}, \texttt{\textbackslash n\textbackslash n},
\texttt{- (bullet)}, \texttt{,}, \texttt{>}
\\

\texttt{CJK}
&
Chinese token (content or connective), when the trajectory reasons in Chinese
&
\zh{观察} (observe), \zh{注意} (note), \zh{回忆} (recall),
\zh{但} (but), \zh{答案} (answer),
\zh{设} (let), \zh{生成} (generate), 
\zh{我们} (we)
\\

\bottomrule
\end{tabularx}

\caption{Categories of tokens with extremely positive reward activated by \textit{minmaxtok}. We audit the trajectories generated in the first 20 steps.}
\label{tab:pos-token-buckets}
\end{table*}

\begin{table*}[t]
\centering
\small
\setlength{\tabcolsep}{6pt}
\renewcommand{\arraystretch}{1.25}

\begin{tabularx}{\textwidth}{
  >{\raggedright\arraybackslash}p{0.17\textwidth}
  >{\raggedright\arraybackslash}p{0.27\textwidth}
  >{\raggedright\arraybackslash}X
}
\toprule
\textbf{category} &
\textbf{description} &
\textbf{examples} \\
\midrule

\texttt{math / LaTeX fragment}
&
notation the student wrote (no error)
&
\texttt{\textbackslash{}frac}, \texttt{=}, \texttt{\^{}},
\texttt{\$}, \texttt{\{}, \texttt{\}}, \texttt{\textbackslash{}},
\texttt{\_}, \texttt{(}, \texttt{left},
\texttt{right}, \texttt{sqrt}
\\

\texttt{correct content word}
&
ordinary word right in context
&\texttt{function}, 
\texttt{rectangle},
\texttt{perpendicular}, \texttt{distance}
\\

\texttt{correct digit / value}
&
numeral that is arithmetically correct
&
the \texttt{7} in $\sigma(4)=7$; the \texttt{2} in
$13\times12=156$
\\

\texttt{format separator}
&
whitespace / markdown / structural punctuation
&
\texttt{space}, \texttt{\textbackslash{}n},
\texttt{\textbackslash{}n\textbackslash{}n}, \texttt{---},
\texttt{\#\#\#}, \texttt{**},
\texttt{:}, \texttt{comma}
\\

\texttt{stop <|im\_end|>}
&
premature end-of-turn after \texttt{\textbackslash boxed\{...\}}
&
\texttt{<|im\_end|>} or \texttt{<|endoftext|>}
right after the boxed answer
\\

\texttt{genuine math error}
&
wrong committed digit/value (a real mistake)
&
$\operatorname{LCM}(15,16)=272$ ($\rightarrow240$);
$\operatorname{median}\{0,120,240\}=240$ ($\rightarrow120$);
$\sin^2(\pi/2)=0$ ($\rightarrow1$)
\\

\texttt{non-English token}
&
fluent, correct Chinese the student switched to
&
\zh{生成} (generate), \zh{计算} (compute),
\zh{图} (graph)
\\

\bottomrule
\end{tabularx}

\caption{Categories of tokens with extremely negative reward activated by \textit{minmaxtok}. We audit the trajectories generated in the first 20 steps.}
\label{tab:neg-token-buckets}
\end{table*}

\section{Cross-Task, Cross-Family and Cross-Algorithm Validation}
In this section, we investigate whether the phenomenon of extremely sparse supervision for incentivizing reasoning ability generalizes to other tasks, model families and post-training algorithms. For cross-task validation, we consider code reasoning \citep{chen2021evaluating} as an additional task, which has been widely studied in the literature.
For cross-family validation, we choose the Llama 3 series \citep{grattafiori2024llama}, an established and widely used model family for LLM research. Its earlier release also provides a natural safeguard against potential data contamination in our evaluation benchmarks. For cross-algorithm validation, we choose PPO \citep{schulman2017proximal}, a widely adopted post-training algorithm for incentivizing reasoning. PPO is particularly suitable for our study because, unlike GRPO \citep{shao2024deepseekmath} and REINFORCE \citep{williams1992simple}, it provides token-level supervision through the advantage function, even when the underlying reward is sparse and outcome-based. In contrast, GRPO and REINFORCE assign the same trajectory-level reward signal to all tokens within the same response. We therefore focus on PPO to examine whether the sparse supervision phenomenon extends to a post-training algorithm with token-level learning signals beyond OPD. Our preliminary experiments with GRPO and REINFORCE found that they do not improve the model under the extremely sparse supervision regime, and we thus do not include them in this work.

\subsection{Sparse OPD on Coding Reasoning}
To evaluate whether the effectiveness of sparse OPD extends beyond mathematical reasoning, we conduct experiments on coding reasoning. We use the Eurus-RL-Code dataset \citep{cui2025process} for training, randomly sampling 12K prompts from the full dataset. For evaluation, we use LiveCodeBench v6 \citep{jain2025livecodebench} and the Eurus-RL-Code validation set. The former evaluates out-of-distribution generalization, while the latter evaluates in-distribution performance.
From each evaluation dataset, we randomly sample 50 prompts to reduce computational cost, as evaluation with $n=256$ generations per prompt is particularly time-consuming.

\begin{figure}
    \centering
    \includegraphics[width=\linewidth]{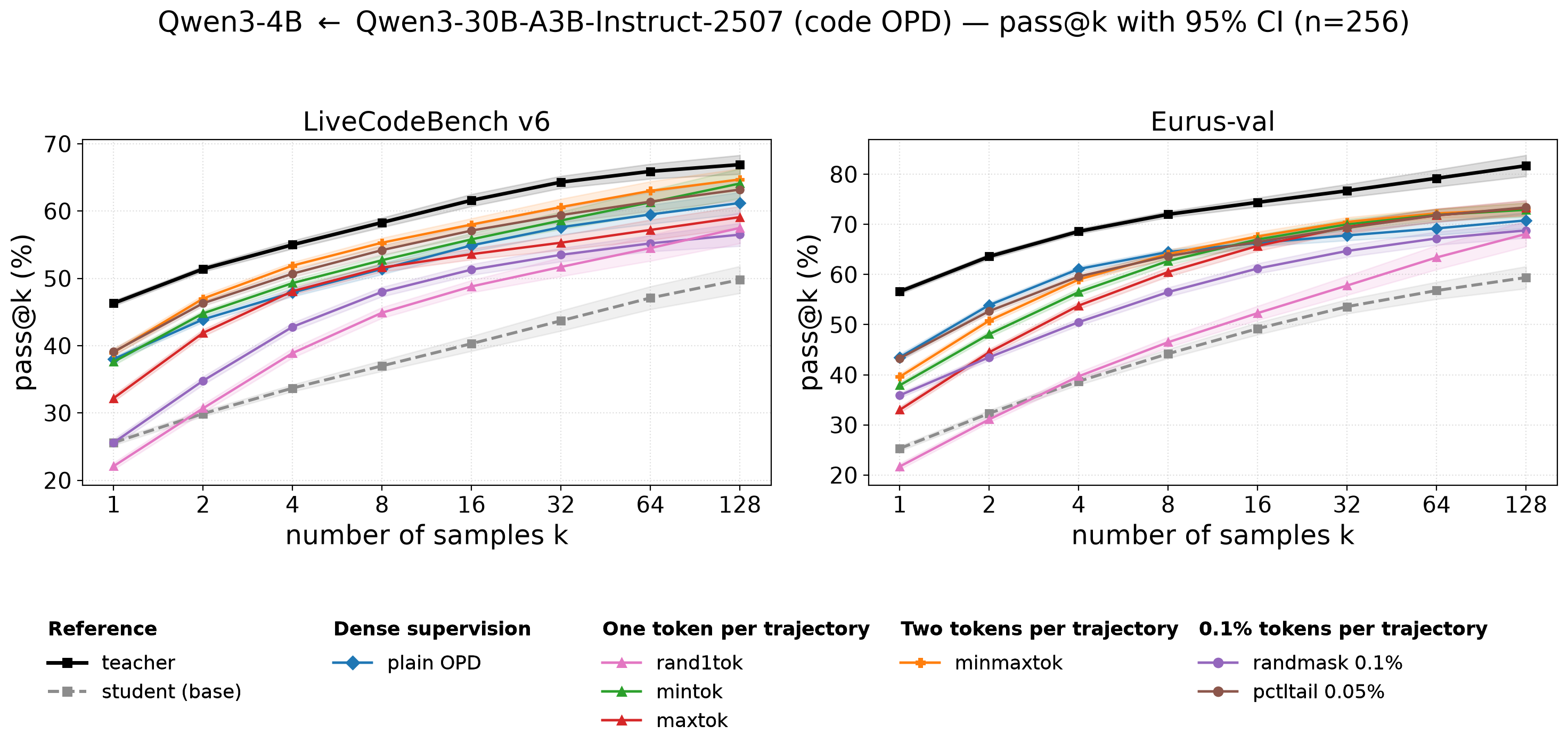}
    \caption{Cross-task validation with coding reasoning. 
    Strong-to-Weak distillation setting with Qwen3-4B as student and Qwen3-30B-A3B-Instruct-2507 as teacher. Figures are pass@k curves of base student model, teacher model, student model trained with plain OPD and student models trained with sparse OPD variants across a subset of LiveCodeBench v6 and Eurus-RL-code dataset. Shaded area is the 95\% confidence interval.}
    \label{fig:family13_passk_combined}
\end{figure}

\begin{figure}
    \centering
    \includegraphics[width=\linewidth]{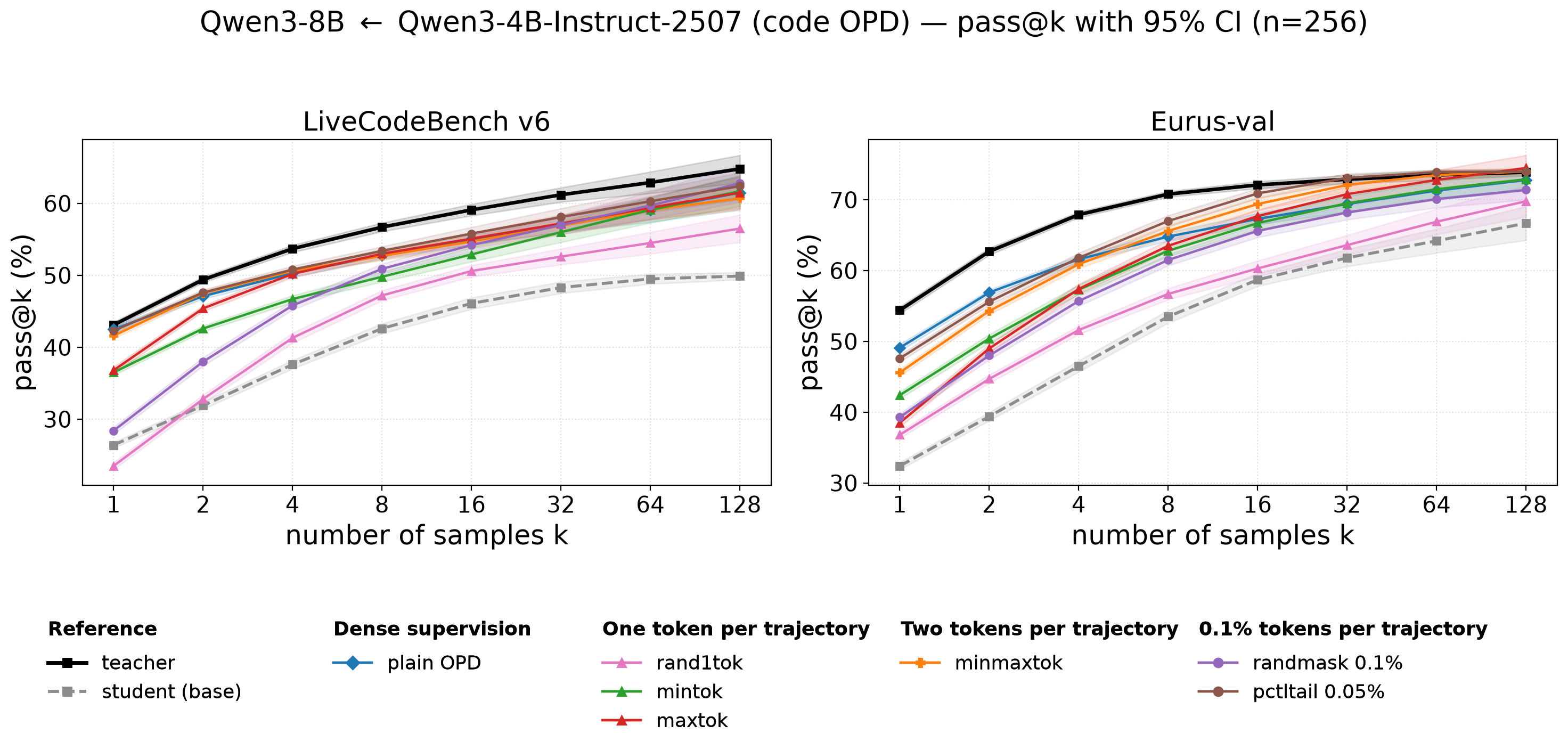}
    \caption{Cross-task validation with coding reasoning. 
    Strong-to-Weak distillation setting with Qwen3-8B as student and Qwen3-4B-Instruct-2507 as teacher. Figures are pass@k curves of base student model, teacher model, student model trained with plain OPD and student models trained with sparse OPD variants across a subset of LiveCodeBench v6 and Eurus-RL-code dataset. Shaded area is the 95\% confidence interval.}
    \label{fig:family14_passk_combined}
\end{figure}

We consider two settings of Strong-to-Weak distillation: (1) Qwen3-30B-A3B-Instruct-2507 as the teacher and Qwen3-4B as the student, representing the large-scale teacher/small-scale student setting; and (2) Qwen3-4B-Instruct-2507 as the teacher and Qwen3-8B as the student, representing the small-scale teacher/large-scale student setting. All models operate in no-think mode throughout training and evaluation, and other experimental settings follow those in \Cref{sec:Extremely Sparse Supervision Incentivizes Reasoning Ability}. Experiment results on pass@k are shown in \Cref{fig:family13_passk_combined,fig:family14_passk_combined}. 
The observations from coding reasoning distillation closely resemble those from mathematical reasoning distillation. Specifically, \textit{rand1tok} significantly improves the base student's reasoning ability. Among the sparse OPD variants, \textit{minmaxtok} and \textit{pctltail} perform best, both of them outperform plain OPD in the large $k$ regime.
In the large-scale teacher/small-scale student setting, \textit{mintok} is also quite effective and consistently outperforms \textit{maxtok} on both benchmarks, while in the small-scale teacher/large-scale student setting, \textit{maxtok} is more effective than \textit{mintok}.

\subsection{Sparse OPD with Llama Models}
\begin{figure}
    \centering
    \includegraphics[width=\linewidth]{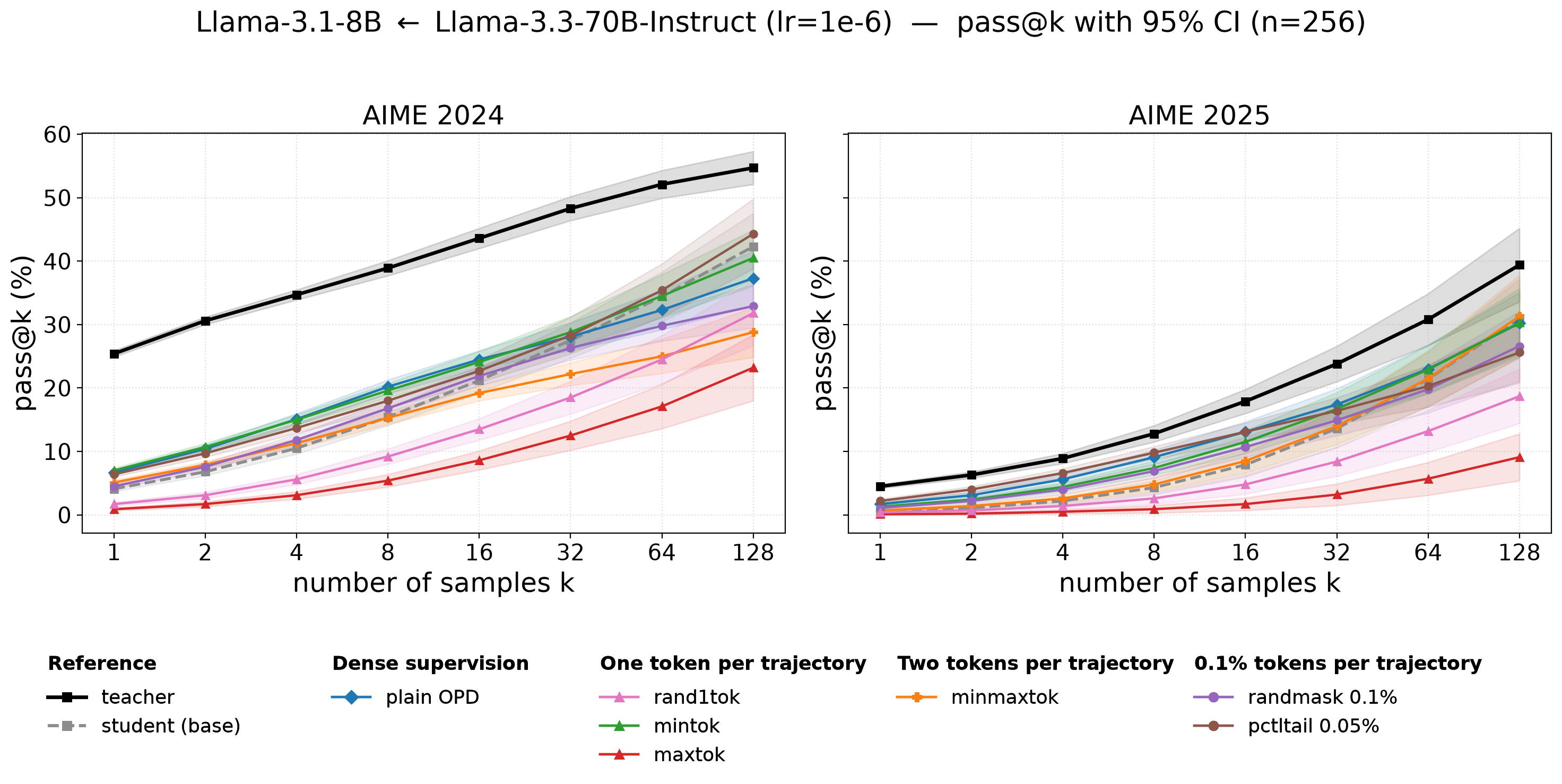}
    \caption{Cross-family validation with Llama models. 
    Strong-to-Weak distillation setting with Llama-3.1-8B-Instruct as student and Llama-3.3-70B-Instruct as teacher. Figures are pass@k curves of base student model, teacher model, student model trained with plain OPD and student models trained with sparse OPD variants across AIME 24 and AIME 25. Shaded area is the 95\% confidence interval.}
    \label{fig:family12_passk_combined}
\end{figure}

As for the cross-family validation, we choose Llama-3.1-8B-Instruct as the student and Llama-3.3-70B-Instruct as the teacher for Strong-to-Weak distillation in math reasoning tasks. All other experimental settings follow those in \Cref{sec:Extremely Sparse Supervision Incentivizes Reasoning Ability}. Experiment results are presented in \Cref{fig:family12_passk_combined}. In particular, \textit{mintok} and \textit{pctltail 0.05\%} generally match the performance of plain OPD, whereas the other sparse OPD variants fail to improve the base student. Although the reasoning improvement from plain OPD is less pronounced than that observed across the nine Qwen3 families, it is still statistically significant. More importantly, we again identify extremely sparse OPD variants that achieve performance comparable to dense OPD, suggesting that the extremely-sparse-supervision phenomenon extends beyond the Qwen3 model family. Taken together, these results lead us to conjecture that extremely sparse supervision incentivizing reasoning ability is closely associated with successful dense OPD: when the conditions for successful OPD are satisfied, the phenomenon emerges as well. In the next section, however, we show that this phenomenon is not exclusive to OPD: under PPO, sparse supervision can yield meaningful reasoning improvements even when dense supervision causes training to collapse.

\subsection{Sparse PPO for RLVR}
Sparse PPO uses the same objective function \eqref{eq:sparse opd obj} as sparse OPD, only the token-level advantage $A_t$ is replaced with generalized advantage estimation \citep{schulman2017proximal}. Thus, all sparse variants, \textit{rand1tok, mintok, maxtok, minmaxtok, rand1tok 0.1\%} and \textit{pctltail 0.05\%}, can be implemented with sparse PPO. Experiment configuration for PPO is provided in \Cref{sec:Experiment Configuration}, \Cref{tab:ppo-config}. Other experiment setup follows that in \Cref{sec:Extremely Sparse Supervision Incentivizes Reasoning Ability}.

\begin{figure}
    \centering
    \includegraphics[width=\linewidth]{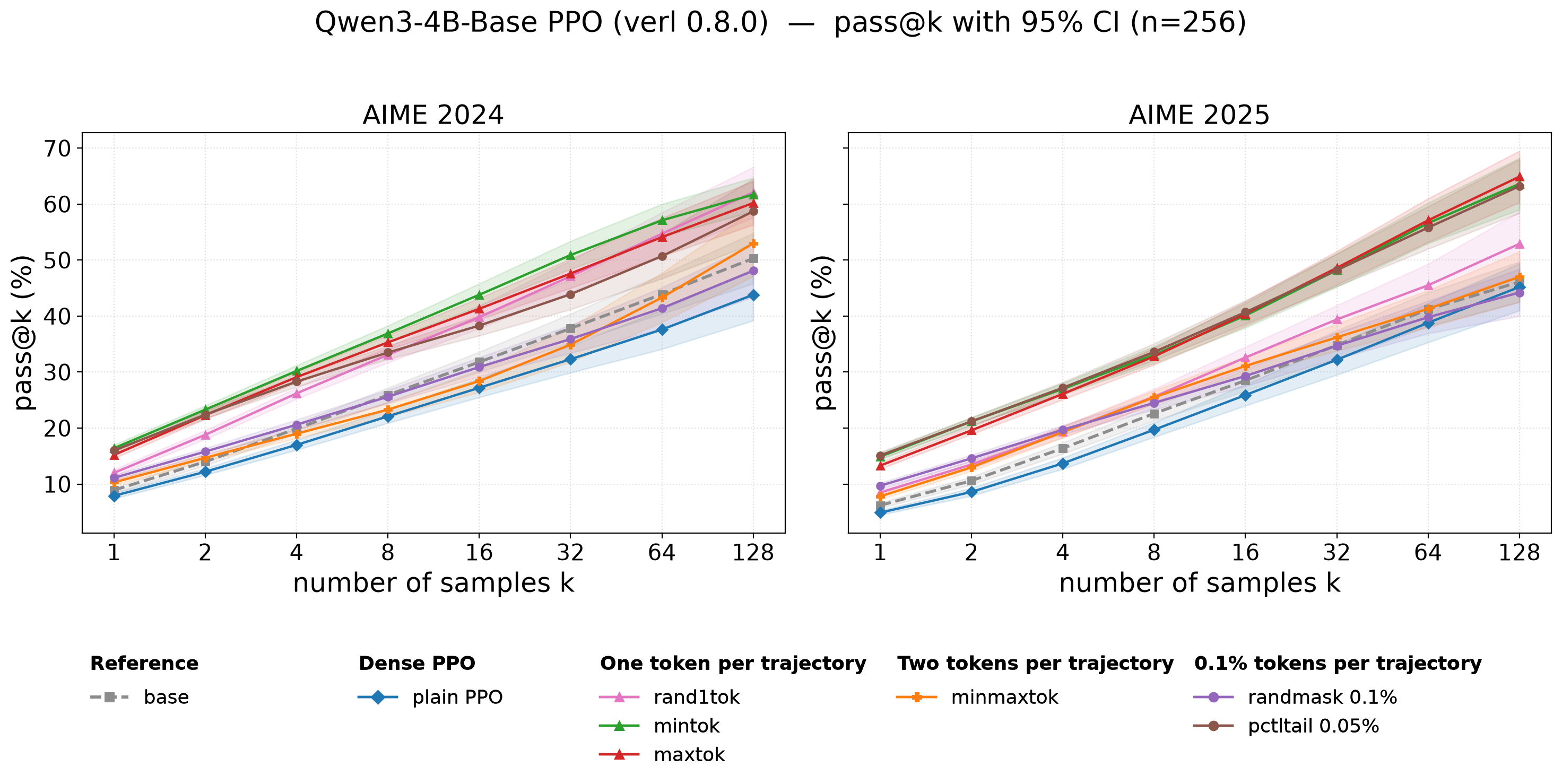}\\
    \includegraphics[width=\linewidth]{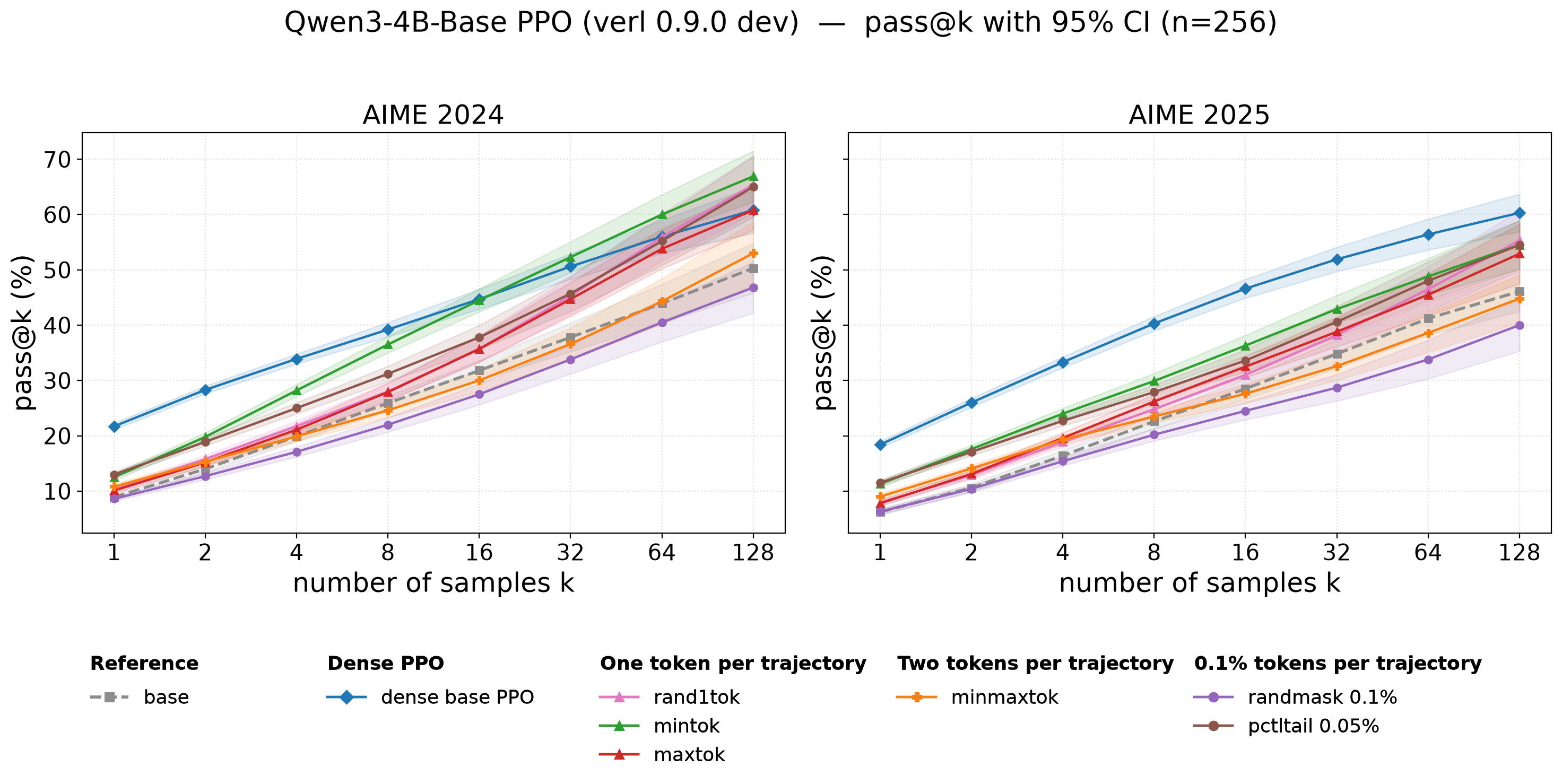}
    \caption{Pass@k curves of base model, model trained with plain PPO and models trained with sparse PPO variants across AIME 24 and AIME 25. Shaded area is the 95\% confidence interval. Top: experiment results with verl 0.8.0. Bottom: experiment results with verl 0.9.0 dev.}
    \label{fig:ppo-pass@k}
\end{figure}

Specifically, when we tune the PPO algorithm with verl \citep{sheng2024hybridflow} 0.8.0, we find PPO keeps failing, but sparse variants works. We then switch to verl 0.9.0 with the same configuration, this time PPO works well. Experiment results are provided in \Cref{fig:ppo-pass@k}, and  kind of convoluted: 0.8.0 version verl based PPO fails to improve the base model, while several 0.8.0 version verl based sparse PPO variants, including \textit{rand1tok, mintok, maxtok, pctltail 0.05\%}  significantly improves the base model. While 0.9.0 dev version verl based plain PPO successfully improves the base model. 0.9.0 dev version verl based sparse PPO variants, including \textit{rand1tok, mintok, maxtok, pctltail 0.05\%}, also significantly improves the base model, and on AIME 24 they match or outperform plain PPO when $k$ is large; on AIME 25 they fall short of the plain PPO. But a closer examination shows that 0.8.0 version verl based \textit{mintak, maxtok, pctltail 0.05\%} outperforms 0.9.0 version verl based plain PPO when $k$ is large. But overall sparse PPO variants does not achieve the same level of  sampling efficiency as 0.9.0 version verl based plain PPO. 

The weaker performances of sparse PPO compared with sparse OPD may stem from the different token-level learning signals used by the two methods. In PPO, token-level advantages are estimated from sparse outcome rewards through temporal-difference-based credit assignment, so they can be noisy and affected by the bias and variance of the advantage estimator. In contrast, OPD obtains its token-level signal directly from the teacher's token preference, without requiring token-level credit to be inferred from the outcome reward. As a result, keeping only a small number of tokens may be more effective in OPD, where the selected tokens still carry a direct teacher-derived signal. We hypothesize that this difference partly explains why sparse supervision is more effective in OPD than in PPO, and leave a more systematic investigation to future work.

\section{Discussion and Limitations}

Our empirical study is primarily conducted on the Qwen3 family and focuses on mathematical reasoning tasks, with additional validation on the Llama family. 
Qwen3 family provides an ideal testbed for studying post-training, offering a broad spectrum of model scales and capabilities that enables systematic evaluation across diverse teacher--student configurations, and has been widely studied in the literature \citep{yang2025qwen3, li2026rethinking, yang2026learning, xu2026tip}. Within this setting, plain OPD is robust and effective across all nine teacher--student families, without evident training collapse or length drift, and the sparse OPD variants built upon it exhibit similarly stable behavior. To assess whether the phenomenon extends beyond Qwen3, we additionally conduct Strong-to-Weak distillation with Llama-3.1-8B-Instruct as the student and Llama-3.3-70B-Instruct as the teacher. Although the improvement from plain OPD is less pronounced than that observed across the Qwen3 families, it remains significant, and we again identify extremely sparse OPD variants that achieve performance comparable to dense OPD. These results provide preliminary evidence that the phenomenon is not specific to the Qwen3 family. We further test the phenomenon of extremely sparse supervision on coding reasoning OPD setting, as well as with the RLVR PPO setting. Our experimental results suggest that this is a broad phenomenon that holds across different model families, reasoning tasks, and post-training algorithms.

Nevertheless, our systematic investigation remains limited in model-family and task diversity. 
When extending our experiments to other model families, including Gemma 3 \citep{team2025gemma}, Gemma 4 \citep{team2026gemma} and Mistral 3 \citep{liu2026ministral}, we found that directly training with plain sampled-token OPD does not improve the student models.
One possible explanation is that we did not first perform SFT on the student models before applying OPD. Since our sparse supervision study builds upon successful OPD training, we therefore do not include these model families in the present study.
Evaluating extremely sparse supervision across a broader range of models, post-training algorithms, and domains is a promising direction for future work.

Despite the limitation, the improvement in reasoning capability is substantial and remarkably consistent across the settings we considered in this work. We believe this phenomenon provides a useful lens for understanding the role of supervision density in post-training and motivates new algorithm development. 
An intriguing direction is whether the small number of informative tokens can be identified without access to a white-box teacher. If so, assigning targeted positive or negative supervision signal to these tokens could potentially lead to new post-training methods that require neither a white-box teacher nor an outcome-level reward, which can be difficult to obtain or define in particular tasks. Extremely sparse supervision may also provide a way to reduce the reliance on shared tokenizer between teacher and student, potentially enabling cross-family distillation even when the underlying tokenizers differ.
Beyond on-policy training, it is also interesting to study whether similar sparsification can be applied to off-policy training, where large amounts of training data are already available, and under suitable conditions sparsification may enable effective updates to focus on a small subset of informative tokens while largely excluding irrelevant or noisy supervision. Finally, another promising direction is continual learning: when a model has already acquired a capability but subsequently loses it through intensive training on other tasks, a small number of carefully selected tokens may be sufficient to efficiently re-incentivize the latent capability without requiring extensive retraining.

\bibliographystyle{ims}
\bibliography{reference}

@article{lu2025onpolicydistillation,
  author = {Kevin Lu and Thinking Machines Lab},
  title = {On-Policy Distillation},
  journal = {Thinking Machines Lab: Connectionism},
  year = {2025},
  note = {https://thinkingmachines.ai/blog/on-policy-distillation},
  doi = {10.64434/tml.20251026},
}

@inproceedings{agarwal2024policy,
  title={On-policy distillation of language models: Learning from self-generated mistakes},
  author={Agarwal, Rishabh and Vieillard, Nino and Zhou, Yongchao and Stanczyk, Piotr and Ramos Garea, Sabela and Geist, Matthieu and Bachem, Olivier},
  booktitle={International Conference on Learning Representations},
  volume={2024},
  pages={21246--21263},
  year={2024}
}

@article{team2026kimi,
  title={Kimi k3: Open frontier intelligence},
  author={Team, Kimi and Bai, Tongtong and Bai, Yifan and Bao, Yiping and Cai, Jianfeng and Cai, Xinyuan and Cao, Peizhou and Cao, Yuxuan and Chai, Ziwei and Charles, Y and others},
  journal={arXiv preprint arXiv:2607.24653},
  year={2026}
}

@article{yang2025qwen3,
  title={Qwen3 technical report},
  author={Yang, An and Li, Anfeng and Yang, Baosong and Zhang, Beichen and Hui, Binyuan and Zheng, Bo and Yu, Bowen and Gao, Chang and Huang, Chengen and Lv, Chenxu and others},
  journal={arXiv preprint arXiv:2505.09388},
  year={2025}
}

@article{guo2025deepseek,
  title={Deepseek-r1: Incentivizing reasoning capability in llms via reinforcement learning},
  author={Guo, Daya and Yang, Dejian and Zhang, Haowei and Song, Junxiao and Wang, Peiyi and Zhu, Qihao and Xu, Runxin and Zhang, Ruoyu and Ma, Shirong and Bi, Xiao and others},
  journal={arXiv preprint arXiv:2501.12948},
  year={2025}
}

@article{schulman2017proximal,
  title={Proximal policy optimization algorithms},
  author={Schulman, John and Wolski, Filip and Dhariwal, Prafulla and Radford, Alec and Klimov, Oleg},
  journal={arXiv preprint arXiv:1707.06347},
  year={2017}
}

@misc{AIME24,
    title={American invitational mathematics examination (aime)
2024},
    author={Yifan Zhang and Team Math-AI.},
    year={2024}
}

@misc{AIME25,
    title={American invitational mathematics examination (aime)
2025},
    author={Yifan Zhang and Team Math-AI.},
    year={2025}
}

@article{balunovic2026matharena,
  title={Matharena: Evaluating llms on uncontaminated math competitions},
  author={Balunovic, Mislav and Dekoninck, Jasper and Petrov, Ivo and Jovanovi{\'c}, Nikola and Vechev, Martin},
  journal={Advances in Neural Information Processing Systems},
  volume={38},
  year={2026}
}

@article{yu2026dapo,
  title={Dapo: An open-source llm reinforcement learning system at scale},
  author={Yu, Qiying and Zhang, Zheng and Zhu, Ruofei and Yuan, Yufeng and Zuo, Xiaochen and Yue, Yu and Dai, Weinan and Fan, Tiantian and Liu, Gaohong and Liu, Lingjun and others},
  journal={Advances in Neural Information Processing Systems},
  volume={38},
  pages={113222--113244},
  year={2026}
}

@article{hinton2015distilling,
  title={Distilling the knowledge in a neural network},
  author={Hinton, Geoffrey and Vinyals, Oriol and Dean, Jeff},
  journal={arXiv preprint arXiv:1503.02531},
  year={2015}
}

@article{wang2026beyond,
  title={Beyond the 80/20 rule: High-entropy minority tokens drive effective reinforcement learning for llm reasoning},
  author={Wang, Shenzhi and Yu, Le and Gao, Chang and Zheng, Chujie and Liu, Shixuan and Lu, Rui and Dang, Kai and Chen, Xiong-Hui and Yang, Jianxin and Zhang, Zhenru and others},
  journal={Advances in Neural Information Processing Systems},
  volume={38},
  pages={115452--115486},
  year={2026}
}

@article{wei2021finetuned,
  title={Finetuned language models are zero-shot learners},
  author={Wei, Jason and Bosma, Maarten and Zhao, Vincent Y and Guu, Kelvin and Yu, Adams Wei and Lester, Brian and Du, Nan and Dai, Andrew M and Le, Quoc V},
  journal={arXiv preprint arXiv:2109.01652},
  year={2021}
}

@article{zhu2026surprising,
  title={The surprising effectiveness of negative reinforcement in llm reasoning},
  author={Zhu, Xinyu and Xia, Mengzhou and Wei, Zhepei and Chen, Wei-Lin and Chen, Danqi and Meng, Yu},
  journal={Advances in Neural Information Processing Systems},
  volume={38},
  pages={126546--126573},
  year={2026}
}

@article{chen2021evaluating,
  title={Evaluating large language models trained on code},
  author={Chen, Mark and Tworek, Jerry and Jun, Heewoo and Yuan, Qiming and Pinto, Henrique Ponde De Oliveira and Kaplan, Jared and Edwards, Harri and Burda, Yuri and Joseph, Nicholas and Brockman, Greg and others},
  journal={arXiv preprint arXiv:2107.03374},
  year={2021}
}

@article{chen2026does,
  title={Does reinforcement learning really incentivize reasoning capacity in llms beyond the base model?},
  author={Chen, Zhiqi and Lu, Rui and Zhao, Andrew and Wang, Zhaokai and Yue, Yang and Song, Shiji and Huang, Gao},
  journal={Advances in Neural Information Processing Systems},
  volume={38},
  pages={57654--57689},
  year={2026}
}

@article{xu2026tip,
  title={Tip: Token importance in on-policy distillation},
  author={Xu, Yuanda and Sang, Hejian and Zhou, Zhengze and He, Ran and Wang, Zhipeng and Geramifard, Alborz},
  journal={arXiv preprint arXiv:2604.14084},
  year={2026}
}

@inproceedings{lightman2024let,
  title={Let's verify step by step},
  author={Lightman, Hunter and Kosaraju, Vineet and Burda, Yuri and Edwards, Harrison and Baker, Bowen and Lee, Teddy and Leike, Jan and Schulman, John and Sutskever, Ilya and Cobbe, Karl},
  booktitle={International Conference on Learning Representations},
  volume={2024},
  pages={39578--39601},
  year={2024}
}

@article{mukherjee2026reinforcement,
  title={Reinforcement learning finetunes small subnetworks in large language models},
  author={Mukherjee, Sagnik and Yuan, Lifan and Hakkani-Tur, Dilek and Peng, Hao},
  journal={Advances in Neural Information Processing Systems},
  volume={38},
  pages={132119--132138},
  year={2026}
}

@article{bengio2015scheduled,
  title={Scheduled sampling for sequence prediction with recurrent neural networks},
  author={Bengio, Samy and Vinyals, Oriol and Jaitly, Navdeep and Shazeer, Noam},
  journal={Advances in neural information processing systems},
  volume={28},
  year={2015}
}

@inproceedings{ross2011reduction,
  title={A reduction of imitation learning and structured prediction to no-regret online learning},
  author={Ross, St{\'e}phane and Gordon, Geoffrey and Bagnell, Drew},
  booktitle={Proceedings of the fourteenth international conference on artificial intelligence and statistics},
  pages={627--635},
  year={2011},
  organization={JMLR Workshop and Conference Proceedings}
}

@inproceedings{gu2024minillm,
  title={Minillm: Knowledge distillation of large language models},
  author={Gu, Yuxian and Dong, Li and Wei, Furu and Huang, Minlie},
  booktitle={International Conference on Learning Representations},
  volume={2024},
  pages={32694--32717},
  year={2024}
}

@article{shao2024deepseekmath,
  title={Deepseekmath: Pushing the limits of mathematical reasoning in open language models},
  author={Shao, Zhihong and Wang, Peiyi and Zhu, Qihao and Xu, Runxin and Song, Junxiao and Bi, Xiao and Zhang, Haowei and Zhang, Mingchuan and Li, YK and Wu, Yang and others},
  journal={arXiv preprint arXiv:2402.03300},
  year={2024}
}

@article{xiao2026mimo,
  title={Mimo-v2-flash technical report},
  author={Xiao, Bangjun and Xia, Bingquan and Yang, Bo and Gao, Bofei and Shen, Bowen and Zhang, Chen and He, Chenhong and Lou, Chiheng and Luo, Fuli and Wang, Gang and others},
  journal={arXiv preprint arXiv:2601.02780},
  year={2026}
}

@inproceedings{li2026rethinking,
  title={Rethinking On-Policy Distillation of Large Language Models: Phenomenology, Mechanism, and Recipe},
  author={Li, Yaxuan and Zuo, Yuxin and He, Bingxiang and Zhang, Jinqian and Xiao, Chaojun and Qian, Cheng and Yu, Tianyu and Gao, Huan-ang and Yang, Wenkai and Liu, Zhiyuan and others},
  booktitle={ICML 2026 Workshop on Foundations of Deep Generative Models: Understanding Memorization, Generalization, and Reasoning},
  year={2026}
}

@article{schulman2025lora,
  author = {John Schulman and Thinking Machines Lab},
  title = {LoRA Without Regret},
  journal = {Thinking Machines Lab: Connectionism},
  year = {2025},
  note = {https://thinkingmachines.ai/blog/lora/},
  doi = {10.64434/tml.20250929},
}

@article{qwen2025qwen25technicalreport,
  title={Qwen2.5 Technical Report},
  author={Qwen Team},
  journal={arXiv preprint arXiv:2412.15115},
  year={2025}
}

@article{kaplan2020scaling,
  title={Scaling laws for neural language models},
  author={Kaplan, Jared and McCandlish, Sam and Henighan, Tom and Brown, Tom B and Chess, Benjamin and Child, Rewon and Gray, Scott and Radford, Alec and Wu, Jeffrey and Amodei, Dario},
  journal={arXiv preprint arXiv:2001.08361},
  year={2020}
}

@article{hoffmann2022training,
  title={Training compute-optimal large language models},
  author={Hoffmann, Jordan and Borgeaud, Sebastian and Mensch, Arthur and Buchatskaya, Elena and Cai, Trevor and Rutherford, Eliza and Casas, Diego de Las and Hendricks, Lisa Anne and Welbl, Johannes and Clark, Aidan and others},
  journal={arXiv preprint arXiv:2203.15556},
  year={2022}
}

@article{team2025gemma,
  title={Gemma 3 technical report},
  author={Team, Gemma and Kamath, Aishwarya and Ferret, Johan and Pathak, Shreya and Vieillard, Nino and Merhej, Ramona and Perrin, Sarah and Matejovicova, Tatiana and Ram{\'e}, Alexandre and Rivi{\`e}re, Morgane and others},
  journal={arXiv preprint arXiv:2503.19786},
  year={2025}
}

@article{team2026gemma,
  title={Gemma 4 technical report},
  author={Team, Gemma and Abd, Sherif El and Aggarwal, Vaibhav and Algayres, Robin and Andreev, Alek and Bachem, Olivier and Ballantyne, Ian and Brick, Cormac and C{\u{a}}rbune, Victor and Casbon, Michelle and others},
  journal={arXiv preprint arXiv:2607.02770},
  year={2026}
}

@article{yang2026learning,
  title={Learning beyond teacher: Generalized on-policy distillation with reward extrapolation},
  author={Yang, Wenkai and Liu, Weijie and Xie, Ruobing and Yang, Kai and Yang, Saiyong and Lin, Yankai},
  journal={arXiv preprint arXiv:2602.12125},
  year={2026}
}

@misc{jang2016variational,
  author       = {Eric Jang},
  title        = {A Beginner's Guide to Variational Methods: Mean-Field Approximation},
  year         = {2016},
  month        = aug,
  howpublished = {\url{https://blog.evjang.com/2016/08/variational-bayes.html}},
}

@article{ouyang2022training,
  title={Training language models to follow instructions with human feedback},
  author={Ouyang, Long and Wu, Jeffrey and Jiang, Xu and Almeida, Diogo and Wainwright, Carroll and Mishkin, Pamela and Zhang, Chong and Agarwal, Sandhini and Slama, Katarina and Ray, Alex and others},
  journal={Advances in neural information processing systems},
  volume={35},
  pages={27730--27744},
  year={2022}
}

@article{bai2022training,
  title={Training a helpful and harmless assistant with reinforcement learning from human feedback},
  author={Bai, Yuntao and Jones, Andy and Ndousse, Kamal and Askell, Amanda and Chen, Anna and DasSarma, Nova and Drain, Dawn and Fort, Stanislav and Ganguli, Deep and Henighan, Tom and others},
  journal={arXiv preprint arXiv:2204.05862},
  year={2022}
}

@article{grattafiori2024llama,
  title={The llama 3 herd of models},
  author={Grattafiori, Aaron and Dubey, Abhimanyu and Jauhri, Abhinav and Pandey, Abhinav and Kadian, Abhishek and Al-Dahle, Ahmad and Letman, Aiesha and Mathur, Akhil and Schelten, Alan and Vaughan, Alex and others},
  journal={arXiv preprint arXiv:2407.21783},
  year={2024}
}

@article{williams1992simple,
  title={Simple statistical gradient-following algorithms for connectionist reinforcement learning},
  author={Williams, Ronald J.},
  journal={Machine Learning},
  volume={8},
  pages={229--256},
  year={1992},
  publisher={Springer}
}

@article{sheng2024hybridflow,
  title   = {HybridFlow: A Flexible and Efficient RLHF Framework},
  author  = {Guangming Sheng and Chi Zhang and Zilingfeng Ye and Xibin Wu and Wang Zhang and Ru Zhang and Yanghua Peng and Haibin Lin and Chuan Wu},
  year    = {2024},
  journal = {arXiv preprint arXiv: 2409.19256}
}

@article{cui2025process,
  title={Process reinforcement through implicit rewards},
  author={Cui, Ganqu and Yuan, Lifan and Wang, Zefan and Wang, Hanbin and Zhang, Yuchen and Chen, Jiacheng and Li, Wendi and He, Bingxiang and Fan, Yuchen and Yu, Tianyu and others},
  journal={arXiv preprint arXiv:2502.01456},
  year={2025}
}

@inproceedings{jain2025livecodebench,
  title={Livecodebench: Holistic and contamination free evaluation of large language models for code},
  author={Jain, Naman and Gu, Alex and Li, Wen-Ding and Yan, Fanjia and Zhang, Tianjun and Wang, Sida and Solar-Lezama, Armando and Sen, Koushik and Stoica, Ion},
  booktitle={International Conference on Learning Representations},
  volume={2025},
  pages={58791--58831},
  year={2025}
}

@article{liu2026ministral,
  title={Ministral 3},
  author={Liu, Alexander H and Khandelwal, Kartik and Subramanian, Sandeep and Jouault, Victor and Rastogi, Abhinav and Sad{\'e}, Adrien and Jeffares, Alan and Jiang, Albert and Cahill, Alexandre and Gavaudan, Alexandre and others},
  journal={arXiv preprint arXiv:2601.08584},
  year={2026}
}

\newpage
\appendix

\section{Full Experiment Results}
\label{sec:full experiment results}
\begin{figure}
    \centering
    \includegraphics[width=\linewidth]{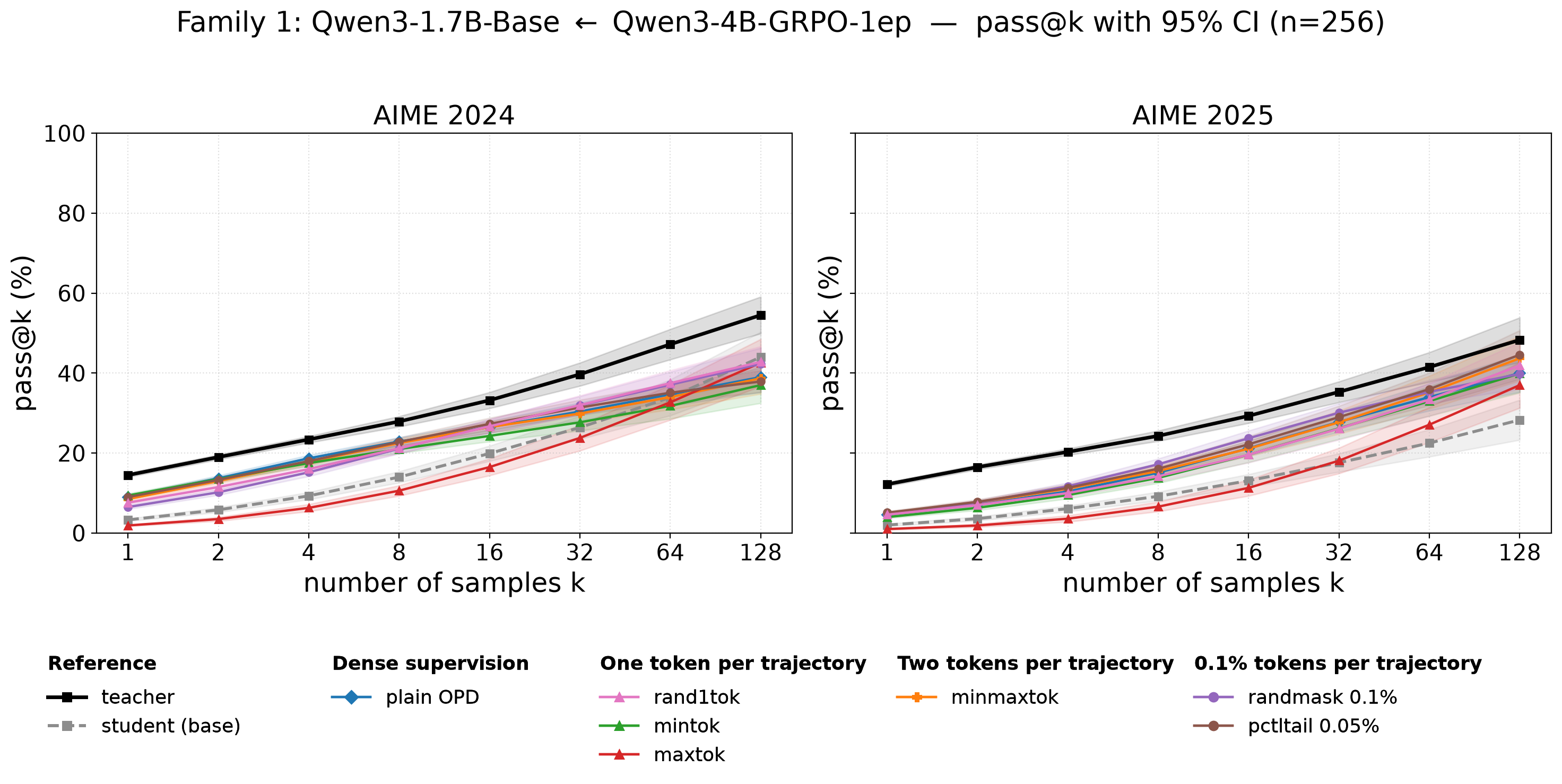}
    \caption{Family 1 pass@k curves of base student model, teacher model, student model trained with plain OPD and student models trained with sparse OPD variants across AIME 24 and AIME 25. Shaded area is the 95\% confidence interval.}
    \label{fig:family1_passk_combined}
\end{figure}

\begin{figure}
    \centering
    \includegraphics[width=\linewidth]{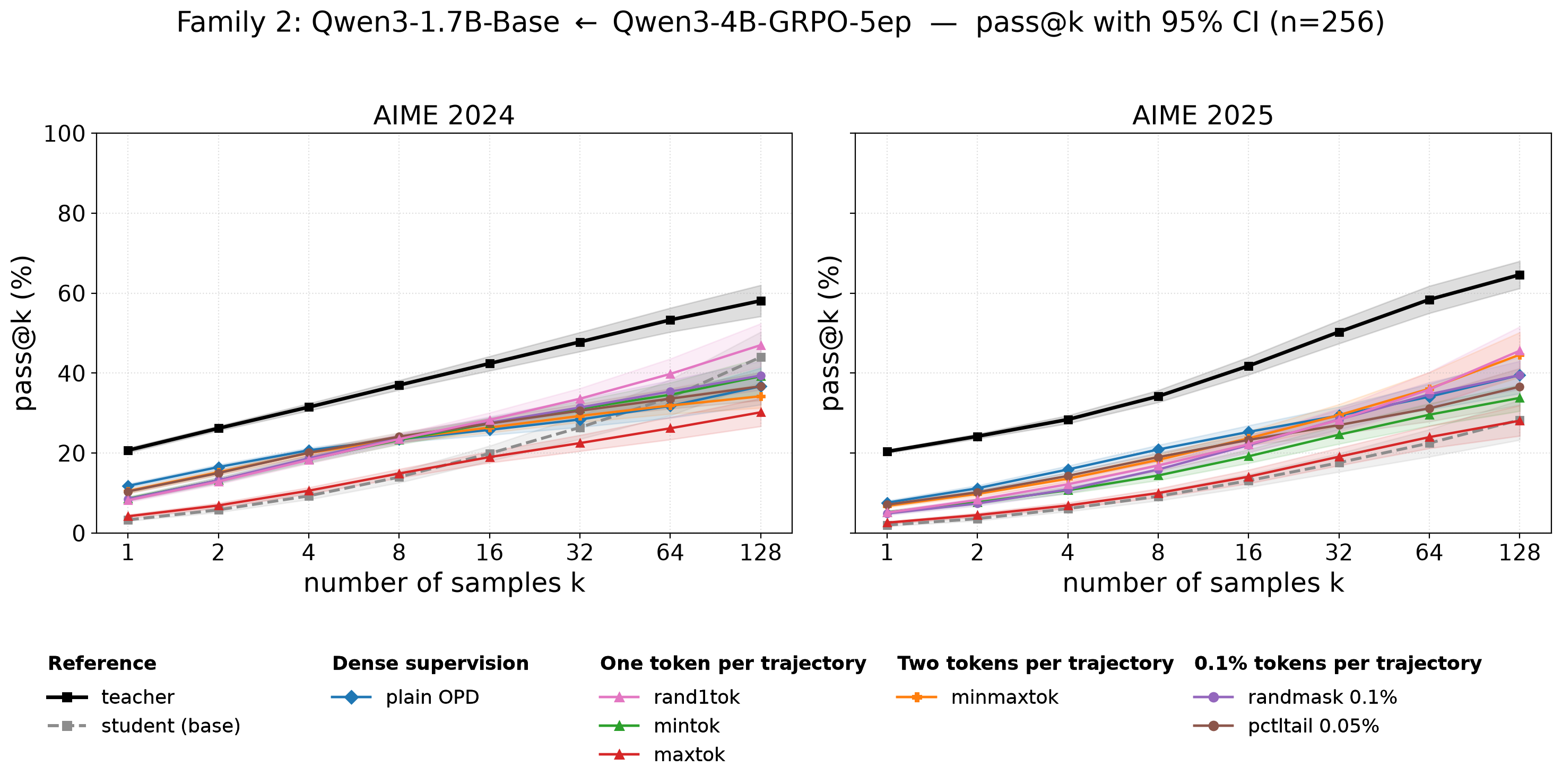}
    \caption{Family 2 pass@k curves of base student model, teacher model, student model trained with plain OPD and student models trained with sparse OPD variants across AIME 24 and AIME 25. Shaded area is the 95\% confidence interval.}
    \label{fig:family2_passk_combined}
\end{figure}

\begin{figure}
    \centering
    \includegraphics[width=\linewidth]{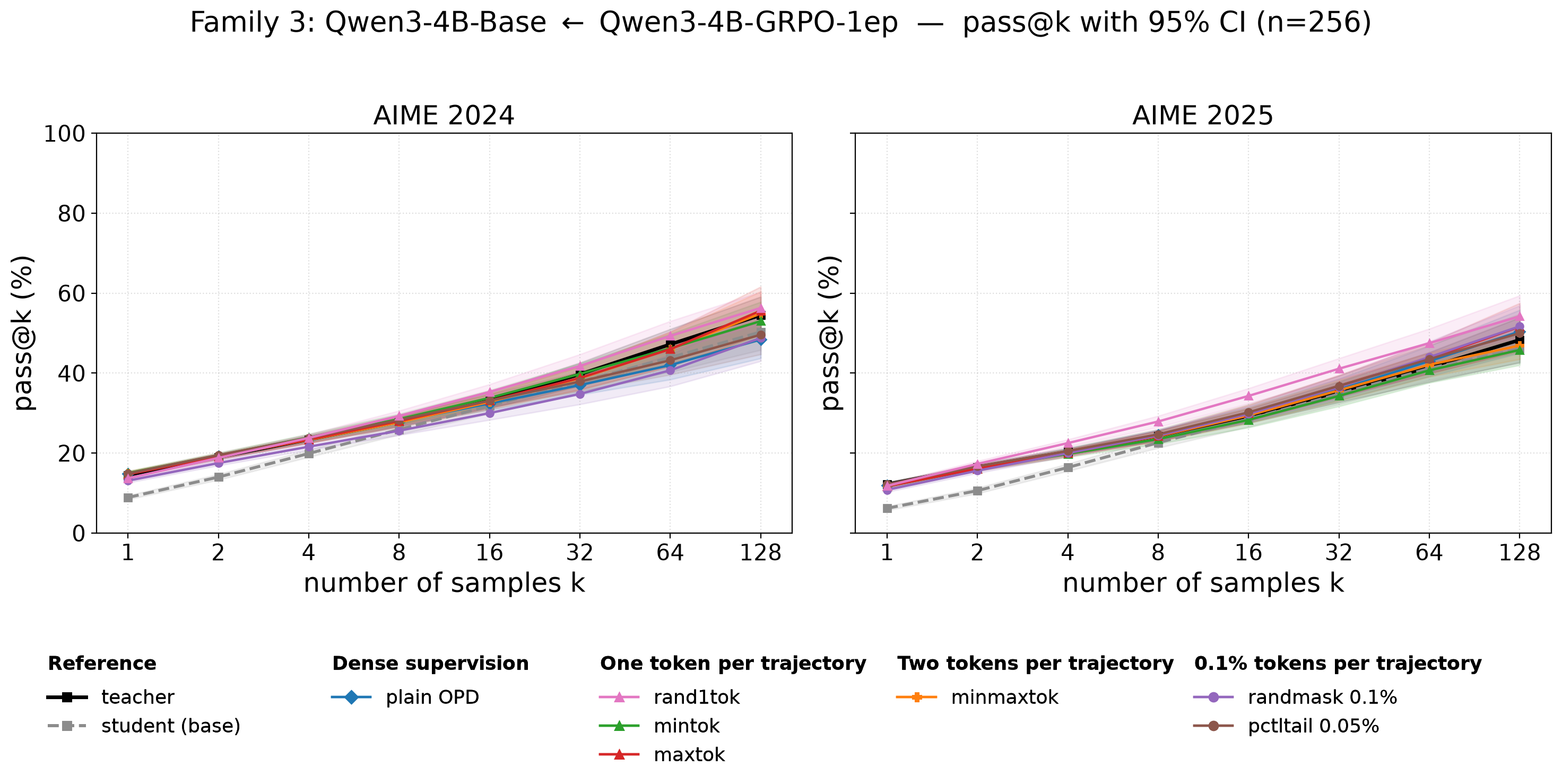}
    \caption{Family 3 pass@k curves of base student model, teacher model, student model trained with plain OPD and student models trained with sparse OPD variants across AIME 24 and AIME 25. Shaded area is the 95\% confidence interval.}
    \label{fig:family3_passk_combined}
\end{figure}

\begin{figure}
    \centering
    \includegraphics[width=\linewidth]{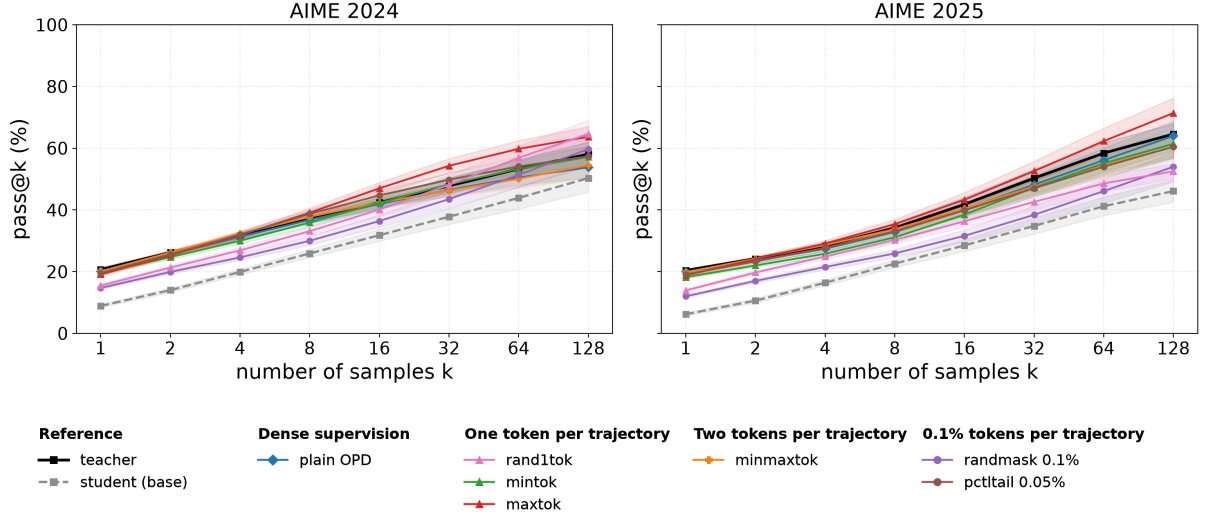}
    \caption{Family 4 pass@k curves of base student model, teacher model, student model trained with plain OPD and student models trained with sparse OPD variants across AIME 24 and AIME 25. Shaded area is the 95\% confidence interval.}
    \label{fig:family4_passk_combined-rep}
\end{figure}

\begin{figure}
    \centering
    \includegraphics[width=\linewidth]{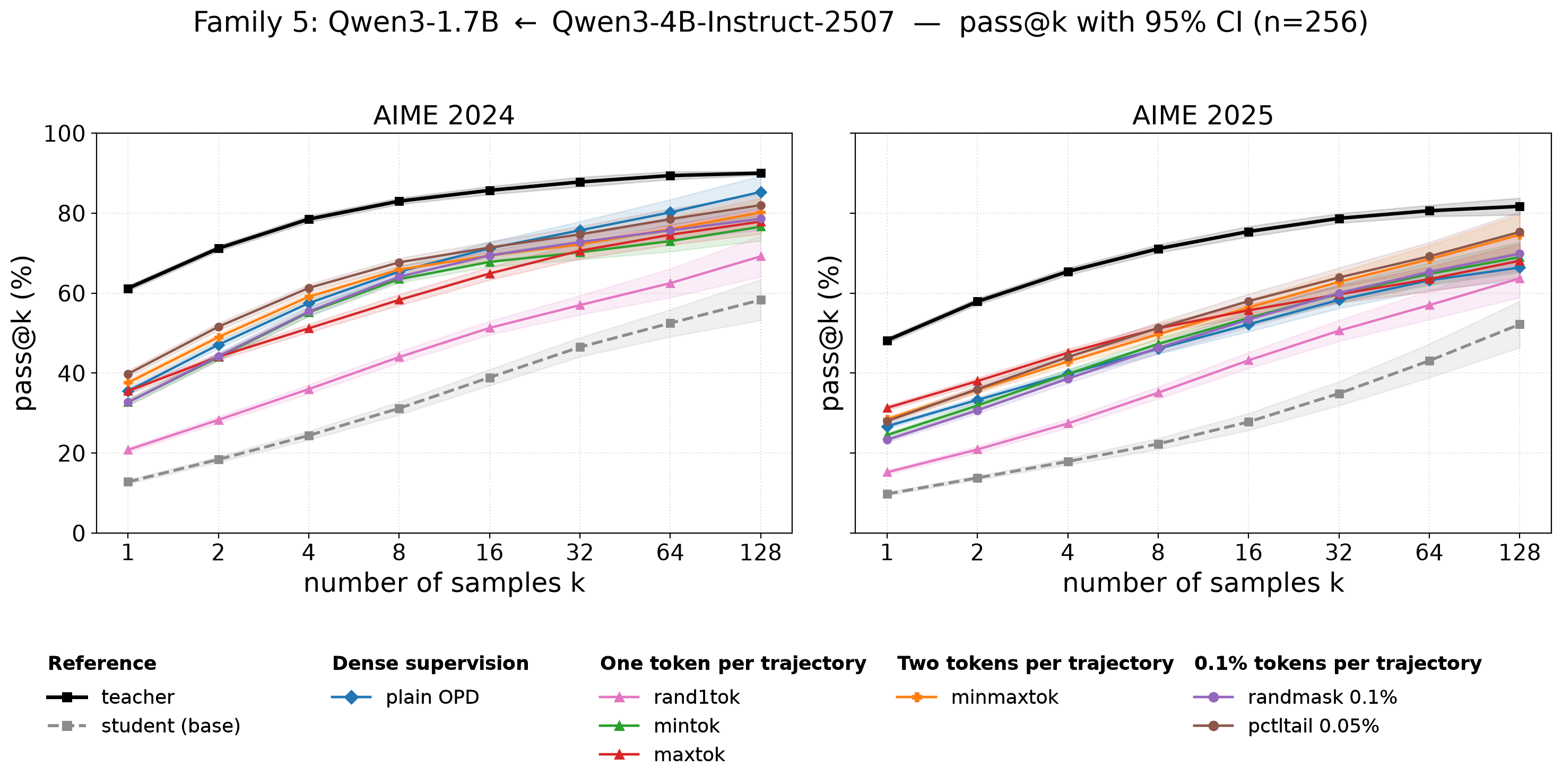}
    \caption{Family 5 pass@k curves of base student model, teacher model, student model trained with plain OPD and student models trained with sparse OPD variants across AIME 24 and AIME 25. Shaded area is the 95\% confidence interval.}
    \label{fig:family5_passk_combined}
\end{figure}

\begin{figure}
    \centering
    \includegraphics[width=\linewidth]{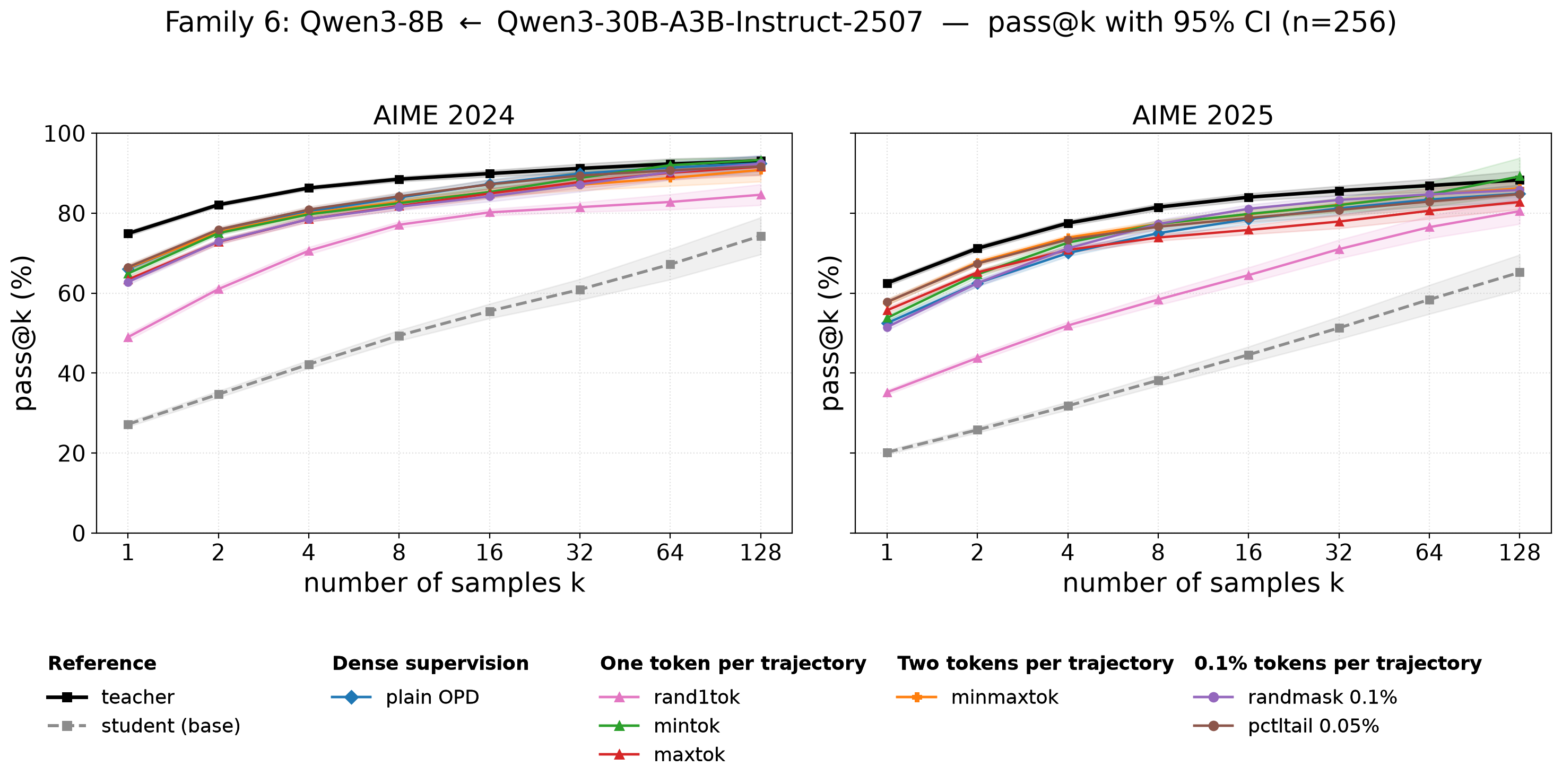}
    \caption{Family 6 pass@k curves of base student model, teacher model, student model trained with plain OPD and student models trained with sparse OPD variants across AIME 24 and AIME 25. Shaded area is the 95\% confidence interval.}
    \label{fig:family6_passk_combined}
\end{figure}

\begin{figure}
    \centering
    \includegraphics[width=\linewidth]{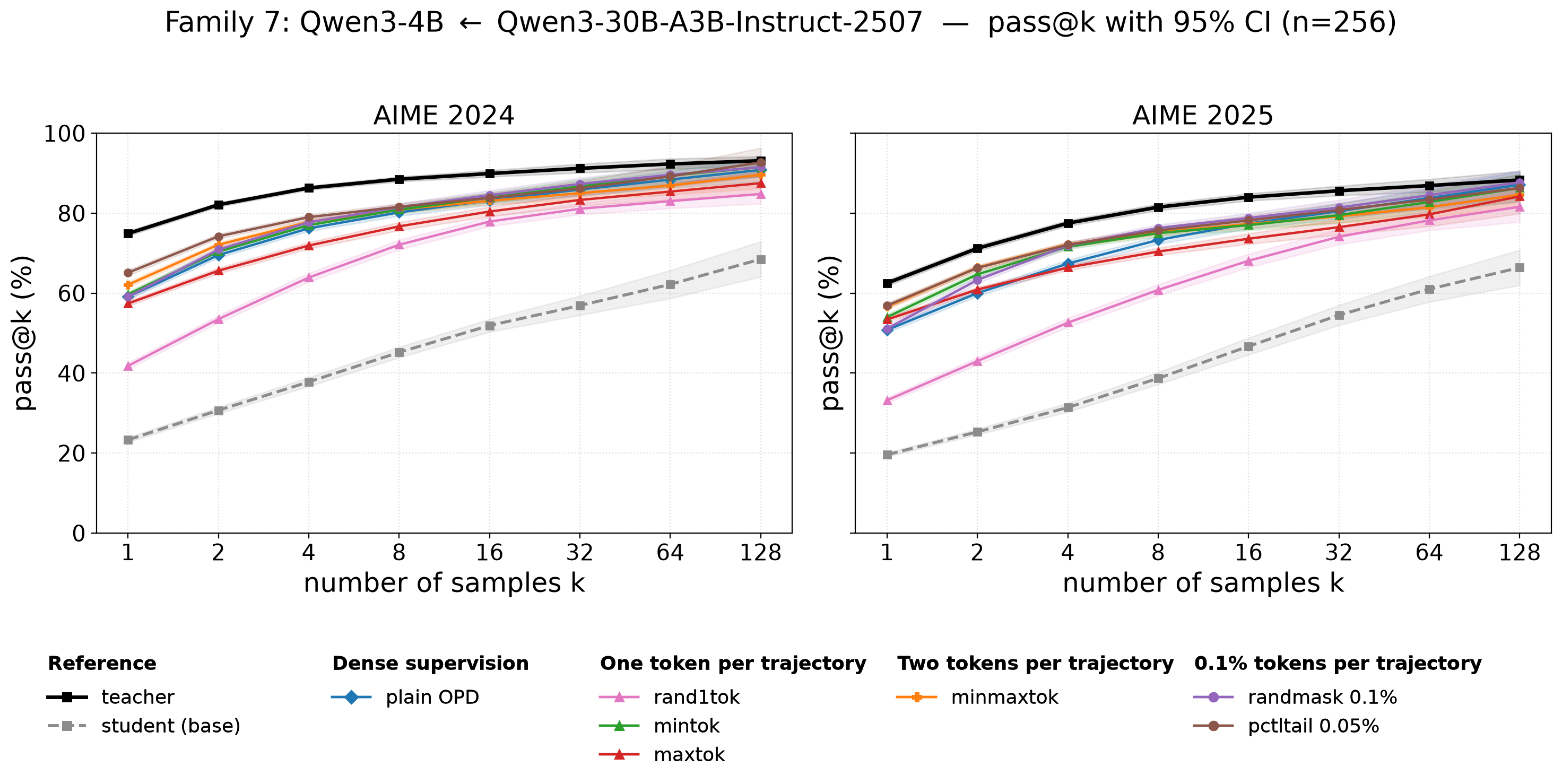}
    \caption{Family 7 pass@k curves of base student model, teacher model, student model trained with plain OPD and student models trained with sparse OPD variants across AIME 24 and AIME 25. Shaded area is the 95\% confidence interval.}
    \label{fig:family7_passk_combined}
\end{figure}

\begin{figure}
    \centering
    \includegraphics[width=\linewidth]{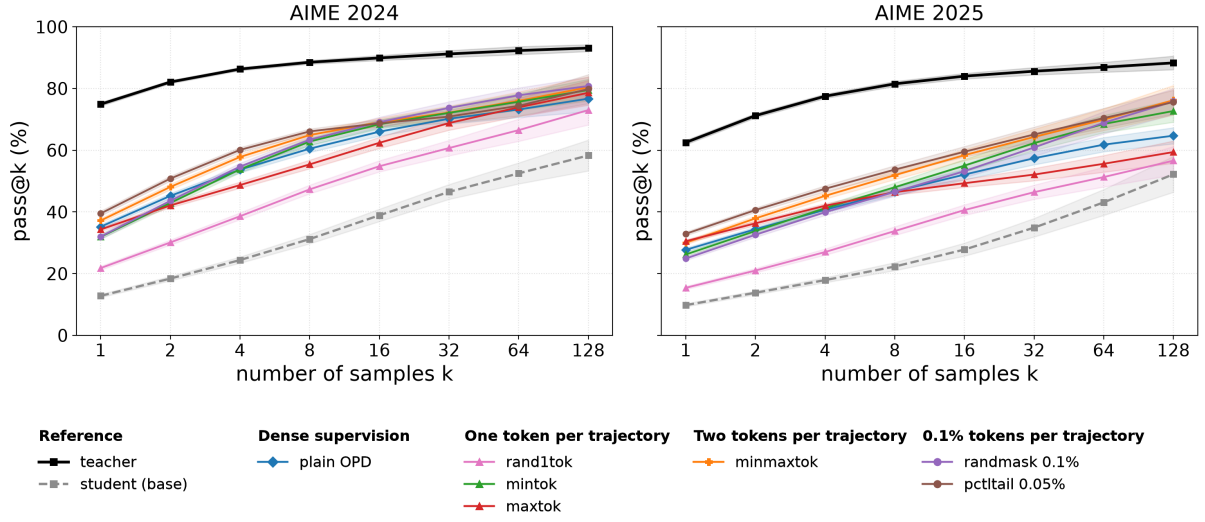}
    \caption{Family 8 pass@k curves of base student model, teacher model, student model trained with plain OPD and student models trained with sparse OPD variants across AIME 24 and AIME 25. Shaded area is the 95\% confidence interval.}
    \label{fig:family8_passk_combined-rep}
\end{figure}

\begin{figure}
    \centering
    \includegraphics[width=\linewidth]{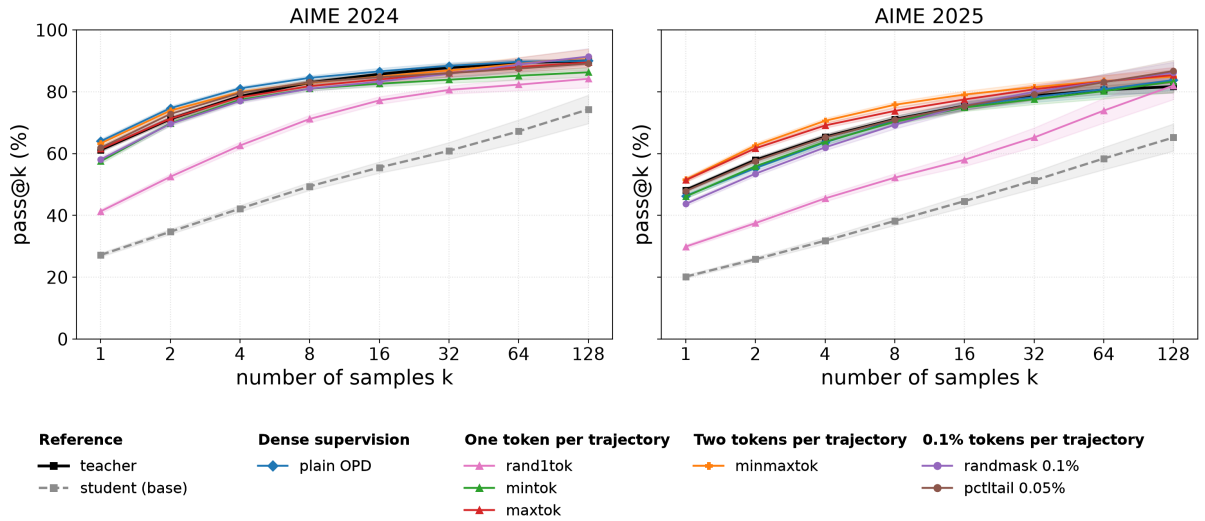}
    \caption{Family 9 pass@k curves of base student model, teacher model, student model trained with plain OPD and student models trained with sparse OPD variants across AIME 24 and AIME 25. Shaded area is the 95\% confidence interval.}
    \label{fig:family9_passk_combined-rep}
\end{figure}

\begin{figure}
    \centering
    \includegraphics[width=\linewidth]{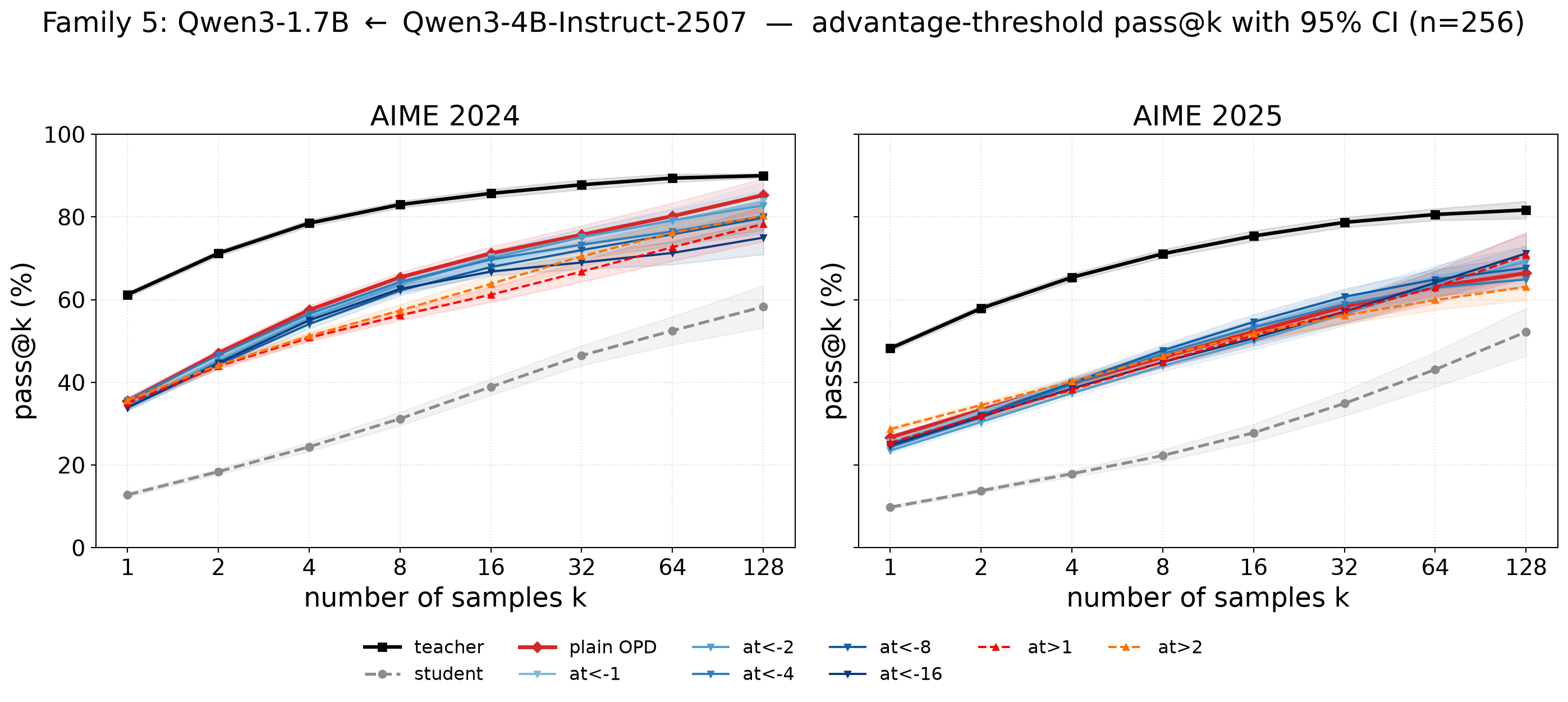}
    \caption{Family 5 pass@k curves of base student model, teacher model, student model trained with plain OPD and student models trained with sparse OPD variants across AIME 24 and AIME 25. Shaded area is the 95\% confidence interval.}
    \label{fig:family5_threshold_passk_combined}
\end{figure}

\begin{figure}
    \centering
    \includegraphics[width=\linewidth]{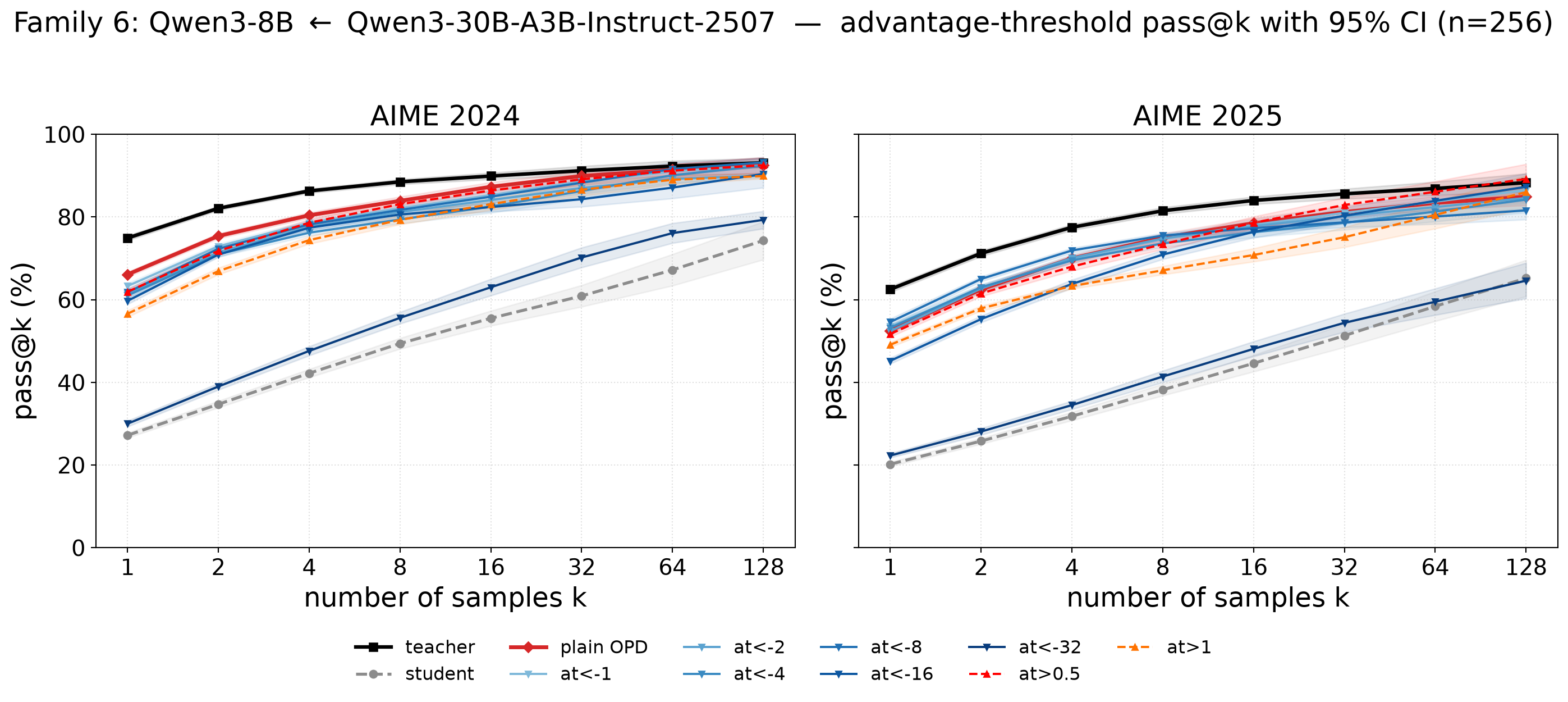}
    \caption{Family 6 pass@k curves of base student model, teacher model, student model trained with plain OPD and student models trained with sparse OPD variants across AIME 24 and AIME 25. Shaded area is the 95\% confidence interval.}
    \label{fig:family6_threshold_passk_combined}
\end{figure}

\begin{figure}
    \centering
    \includegraphics[width=\linewidth]{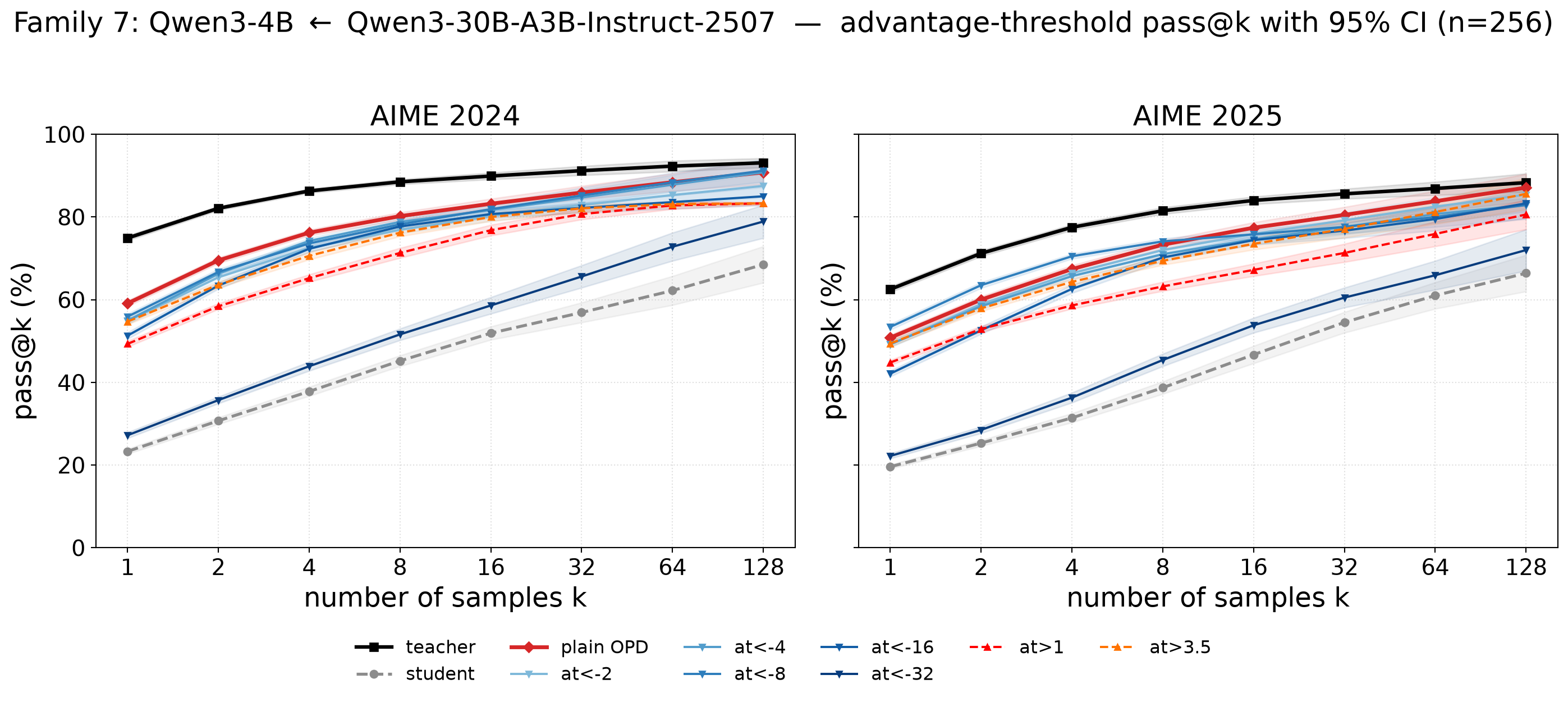}
    \caption{Family 7 pass@k curves of base student model, teacher model, student model trained with plain OPD and student models trained with sparse OPD variants across AIME 24 and AIME 25. Shaded area is the 95\% confidence interval.}
    \label{fig:family7_threshold_passk_combined}
\end{figure}

\begin{figure}
    \centering
    \includegraphics[width=\linewidth]{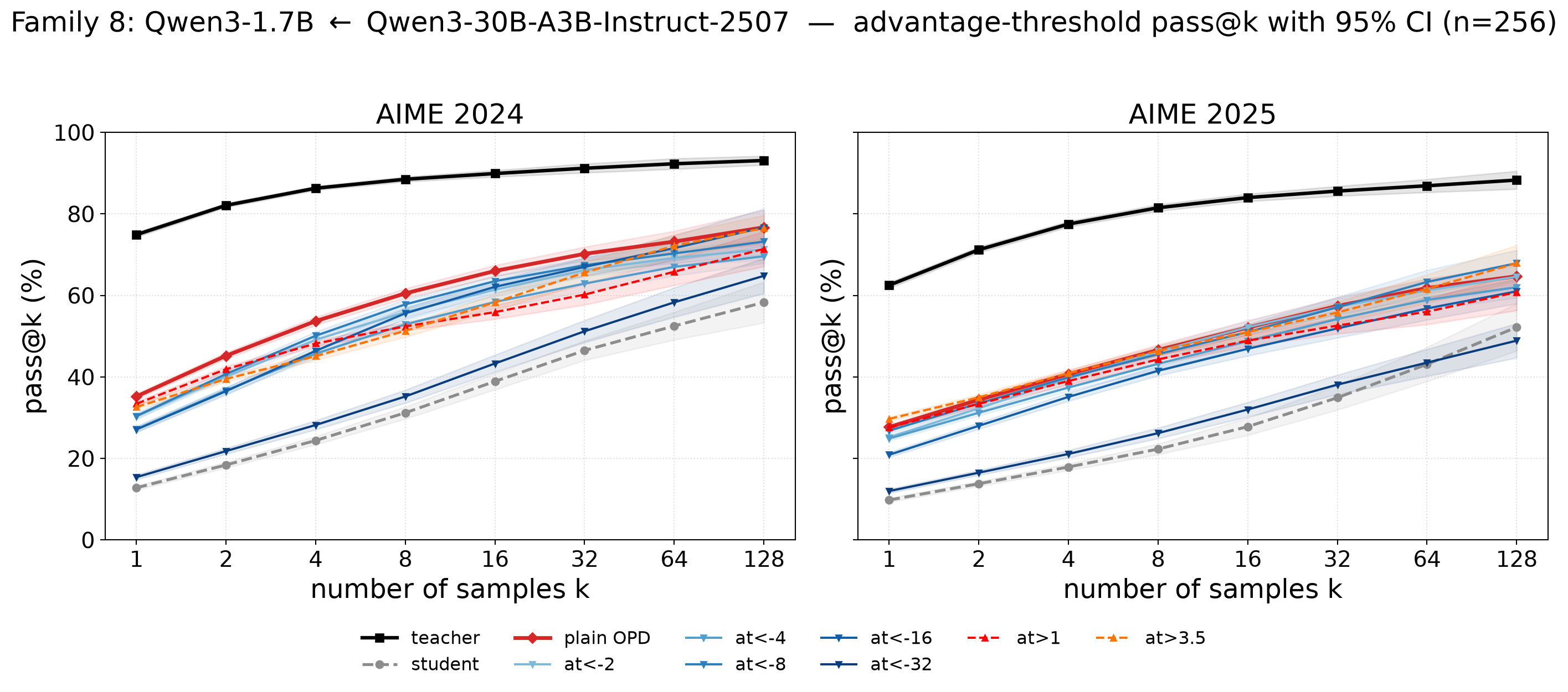}
    \caption{Family 8 pass@k curves of base student model, teacher model, student model trained with plain OPD and student models trained with sparse OPD variants across AIME 24 and AIME 25. Shaded area is the 95\% confidence interval.}
    \label{fig:family8_threshold_passk_combined}
\end{figure}

\begin{figure}
    \centering
    \includegraphics[width=\linewidth]{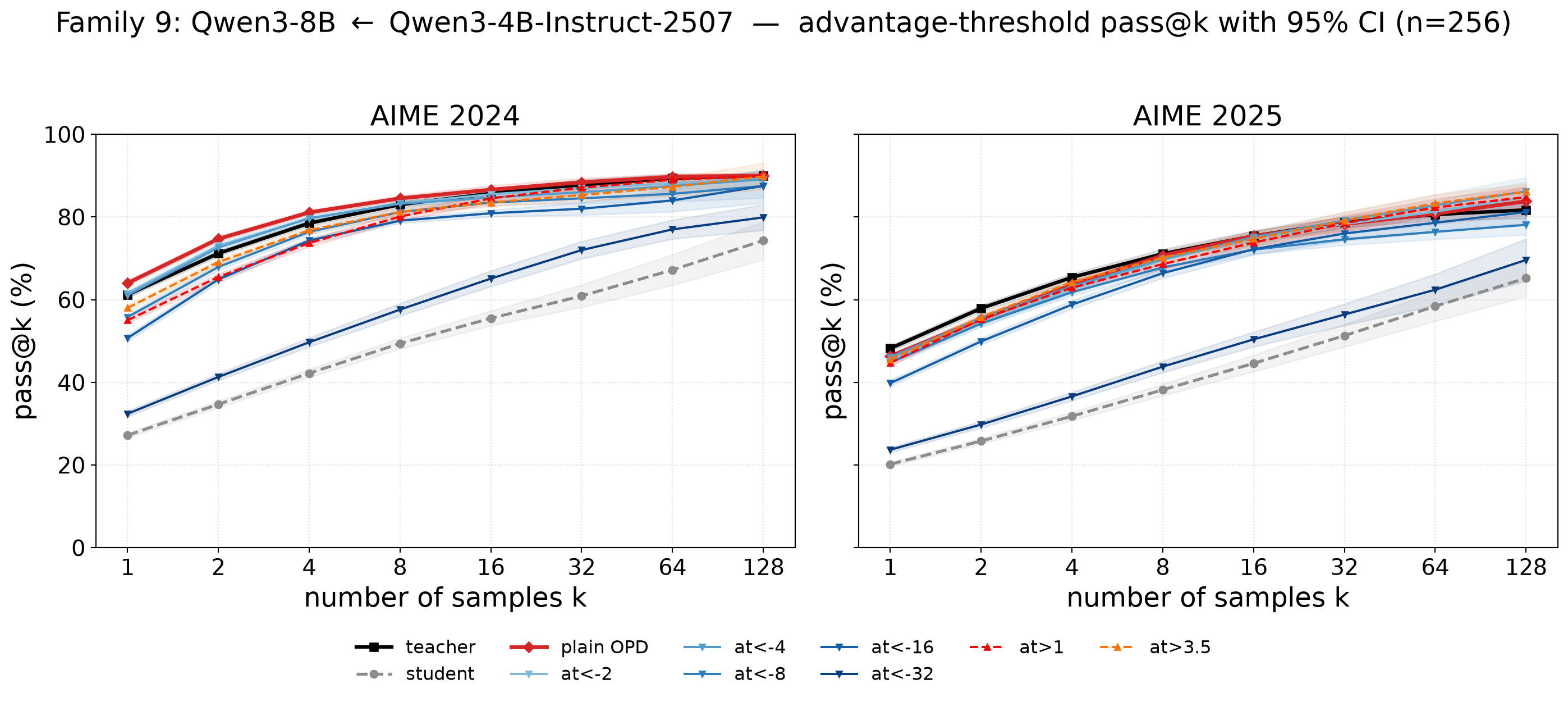}
    \caption{Family 9 pass@k curves of base student model, teacher model, student model trained with plain OPD and student models trained with sparse OPD variants across AIME 24 and AIME 25. Shaded area is the 95\% confidence interval.}
    \label{fig:family9_threshold_passk_combined}
\end{figure}

In this section, we provide complete experiment results across the nine families of Strong-to-Weak distillation. Main results on pass@k are provided in \Cref{fig:family1_passk_combined,fig:family2_passk_combined,fig:family3_passk_combined,fig:family4_passk_combined-rep,fig:family5_passk_combined,fig:family6_passk_combined,fig:family7_passk_combined,fig:family8_passk_combined-rep,fig:family9_passk_combined-rep}, main results on avg@8 are provided in \Cref{tab:family1-tokenmask,tab:family2-tokenmask,tab:family3-tokenmask,tab:family4-tokenmask-rep,tab:family5-tokenmask,tab:family6-tokenmask,tab:family7-tokenmask,tab:family8-tokenmask-rep,tab:family9-tokenmask-rep}.
Ablation results of threshold sweeping on pass@k are provided in \Cref{fig:family5_threshold_passk_combined,fig:family6_threshold_passk_combined,fig:family7_threshold_passk_combined,fig:family8_threshold_passk_combined,fig:family9_threshold_passk_combined}, ablation results of threshold sweeping on avg@8
are provided in \Cref{tab:family1-threshold,tab:family2-threshold,tab:family3-threshold,tab:family4-threshold,tab:family5-threshold,tab:family6-threshold,tab:family7-threshold,tab:family8-threshold,tab:family9-threshold}.

\begin{table}[t]
\centering
\small
\caption{Family 1: Qwen3-1.7B-Base $\leftarrow$ Qwen3-4B-GRPO-1ep. avg@8 scores of the base student model, teacher model, student model trained with plain OPD and student models trained with sparse OPD variants across AIME 2024, AIME 2025 and HMMT Feb 2025. Mean averages over AIME24/AIME25/HMMT.
{keep\_frac (\%)}: fraction of response tokens supervised among all generated tokens; 
{revKL}: reverse KL divergence between the student and teacher; {freeze (\%)}: fraction of student network unchanged parameters after training.}
\label{tab:family1-tokenmask}
\begin{tabular}{l cccc c c c}
\toprule
 & \multicolumn{4}{c}{avg@8 (\%)} & & & \\
\cmidrule(lr){2-5}
Method & AIME24 & AIME25 & HMMT & Mean & keep\_frac (\%) & revKL & freeze (\%) \\
\midrule
Teacher & 13.3 & 12.9 & 3.3 & 9.8 & --    & 0.000 & --   \\
Student & 1.7  & 0.8  & 0.4 & 1.0 & --    & 0.156 & --   \\
plain OPD & 8.3  & 5.0  & 0.4 & 4.6 & 100     & 0.030 & 82.2 \\
\cmidrule(lr){1-8}
\multicolumn{8}{c}{\emph{one token supervision per trajectory}} \\[3pt]
rand1tok & 7.1  & 5.4  & 0.8 & 4.4 & 0.0757  & 0.051 & 89.5 \\
mintok & 7.9  & 3.3  & 0.4 & 3.9 & 0.0521  & 0.033 & 86.0 \\
maxtok & 1.7  & 2.9  & 0.4 & 1.7 & 0.0427  & 0.414 & 87.6 \\
\cmidrule(lr){1-8}
\multicolumn{8}{c}{\emph{two tokens supervision per trajectory}} \\[3pt]
minmaxtok & 9.2  & 4.6  & 0.0 & 4.6 & 0.1150  & 0.054 & 86.0 \\
\cmidrule(lr){1-8}
\multicolumn{8}{c}{\emph{0.1\% tokens supervision per trajectory}} \\[3pt]
randmask 0.1\% & 8.3  & 4.6  & 0.8 & 4.6 & 0.0986  & 0.060 & 88.1 \\
pctltail 0.05\% & 10.8 & 7.1  & 0.0 & \textbf{6.0} & 0.267   & 0.044 & 85.7 \\
\bottomrule
\end{tabular}
\end{table}

\begin{table}[t]
\centering
\small
\caption{Family 2: Qwen3-1.7B-Base $\leftarrow$ Qwen3-4B-GRPO-5ep. avg@8 scores of the base student model, teacher model, student model trained with plain OPD and student models trained with sparse OPD variants across AIME 2024, AIME 2025 and HMMT Feb 2025. Mean averages over AIME24/AIME25/HMMT.
{keep\_frac (\%)}: fraction of response tokens supervised among all generated tokens; 
{revKL}: reverse KL divergence between the student and teacher; {freeze (\%)}: fraction of student network unchanged parameters after training.}
\label{tab:family2-tokenmask}
\begin{tabular}{l cccc c c c}
\toprule
 & \multicolumn{4}{c}{avg@8 (\%)} & & & \\
\cmidrule(lr){2-5}
Method & AIME24 & AIME25 & HMMT & Mean & keep\_frac (\%) & revKL & freeze (\%) \\
\midrule
Teacher & 20.8 & 17.9 & 8.8 & 15.8 & --    & 0.000 & --   \\
Student & 1.7  & 0.8  & 0.4 & 1.0  & --    & 0.361 & --   \\
plain OPD & 10.8 & 9.6  & 2.1 & \textbf{7.5} & 100     & 0.138 & 78.7 \\
\cmidrule(lr){1-8}
\multicolumn{8}{c}{\emph{one token supervision per trajectory}} \\[3pt]
rand1tok & 8.3  & 5.8  & 0.8 & 5.0 & 0.0611  & 0.156 & 89.1 \\
mintok & 7.9  & 3.3  & 0.8 & 4.0 & 0.0289  & 0.083 & 84.2 \\
maxtok & 4.6  & 3.8  & 0.4 & 2.9 & 0.0512  & 0.280 & 86.2 \\
\cmidrule(lr){1-8}
\multicolumn{8}{c}{\emph{two tokens supervision per trajectory}} \\[3pt]
minmaxtok & 10.0 & 6.7  & 0.8 & 5.8 & 0.0645  & 0.123 & 84.3 \\
\cmidrule(lr){1-8}
\multicolumn{8}{c}{\emph{0.1\% tokens supervision per trajectory}} \\[3pt]
randmask 0.1\% & 9.6  & 4.2  & 1.2 & 5.0 & 0.0995  & 0.120 & 88.2 \\
pctltail 0.05\% & 7.9  & 7.5  & 2.1 & 5.8 & 0.1840  & 0.162 & 83.3 \\
\bottomrule
\end{tabular}
\end{table}

\begin{table}[t]
\centering
\small
\caption{Family 3: Qwen3-4B-Base $\leftarrow$ Qwen3-4B-GRPO-1ep. avg@8 scores of the base student model, teacher model, student model trained with plain OPD and student models trained with sparse OPD variants across AIME 2024, AIME 2025 and HMMT Feb 2025. Mean averages over AIME24/AIME25/HMMT.
{keep\_frac (\%)}: fraction of response tokens supervised among all generated tokens; 
{revKL}: reverse KL divergence between the student and teacher; {freeze (\%)}: fraction of student network unchanged parameters after training.}
\label{tab:family3-tokenmask}
\begin{tabular}{l cccc c c c}
\toprule
 & \multicolumn{4}{c}{avg@8 (\%)} & & & \\
\cmidrule(lr){2-5}
Method & AIME24 & AIME25 & HMMT & Mean & keep\_frac (\%) & revKL & freeze (\%) \\
\midrule
Teacher & 13.3 & 12.9 & 3.3 & 9.8 & --    & 0.000 & --   \\
Student & 9.6  & 7.9  & 0.8 & 6.1 & --    & 0.082 & --   \\
plain OPD & 15.8 & 14.2 & 4.2 & 11.4 & 100     & 0.001 & 84.0 \\
\cmidrule(lr){1-8}
\multicolumn{8}{c}{\emph{one token supervision per trajectory}} \\[3pt]
rand1tok & 12.9 & 10.4 & 5.0 & 9.4  & 0.0806  & 0.004 & 93.4 \\
mintok & 14.2 & 13.3 & 2.9 & 10.1 & 0.0673  & 0.001 & 88.5 \\
maxtok & 15.4 & 15.8 & 5.8 & 12.4 & 0.0731  & 0.001 & 90.3 \\
\cmidrule(lr){1-8}
\multicolumn{8}{c}{\emph{two tokens supervision per trajectory}} \\[3pt]
minmaxtok & 16.7 & 16.2 & 5.0 & \textbf{12.6} & 0.1370  & 0.000 & 88.4 \\
\cmidrule(lr){1-8}
\multicolumn{8}{c}{\emph{0.1\% tokens supervision per trajectory}} \\[3pt]
randmask 0.1\% & 15.4 & 12.9 & 2.5 & 10.3 & 0.1020  & 0.006 & 91.0 \\
pctltail 0.05\% & 17.1 & 13.8 & 3.8 & 11.5 & 0.2770  & 0.001 & 88.1 \\
\bottomrule
\end{tabular}
\end{table}

\begin{table}[t]
\centering
\small
\caption{Family 4: Qwen3-4B-Base $\leftarrow$ Qwen3-4B-GRPO-5ep. avg@8 scores of the base student model, teacher model, student model trained with plain OPD and student models trained with sparse OPD variants across AIME 2024, AIME 2025 and HMMT Feb 2025. Mean averages over AIME24/AIME25/HMMT.
{keep\_frac (\%)}: fraction of response tokens supervised among all generated tokens; 
{revKL}: reverse KL divergence between the student and teacher; {freeze (\%)}: fraction of student network unchanged parameters after training.}
\label{tab:family4-tokenmask-rep}
\begin{tabular}{l cccc c c c}
\toprule
 & \multicolumn{4}{c}{avg@8 (\%)} & & & \\
\cmidrule(lr){2-5}
Method & AIME24 & AIME25 & HMMT & Mean & keep\_frac (\%) & revKL & freeze (\%) \\
\midrule
Teacher & 20.8 & 17.9 & 8.8 & 15.8 & --    & 0.000 & --   \\
Student & 9.6  & 7.9  & 0.8 & 6.1  & --    & 0.441 & --   \\
plain OPD & 23.3 & 17.5 & 7.5 & 16.1 & 100     & 0.005 & 75.0 \\
\cmidrule(lr){1-8}
\multicolumn{8}{c}{\emph{one token supervision per trajectory}} \\[3pt]
rand1tok & 12.1 & 13.8 & 6.7 & 10.8 & 0.0601  & 0.078 & 90.2 \\
mintok & 19.2 & 17.5 & 6.2 & 14.3 & 0.0369  & 0.018 & 81.9 \\
maxtok & 22.9 & 18.8 & 7.5 & \textbf{16.4} & 0.0364  & 0.428 & 89.5 \\
\cmidrule(lr){1-8}
\multicolumn{8}{c}{\emph{two tokens supervision per trajectory}} \\[3pt]
minmaxtok & 21.7 & 19.6 & 5.0 & 15.4 & 0.0622  & 0.015 & 81.8 \\
\cmidrule(lr){1-8}
\multicolumn{8}{c}{\emph{0.1\% tokens supervision per trajectory}} \\[3pt]
randmask 0.1\% & 16.2 & 12.5 & 5.4 & 11.4 & 0.1000  & 0.119 & 88.6 \\
pctltail 0.05\% & 19.2 & 18.3 & 5.4 & 14.3 & 0.1770  & 0.012 & 80.9 \\
\bottomrule
\end{tabular}
\end{table}

\begin{table}[t]
\centering
\small
\caption{Family 5: Qwen3-1.7B $\leftarrow$ Qwen3-4B-Instruct-2507. avg@8 scores of the base student model, teacher model, student model trained with plain OPD and student models trained with sparse OPD variants across AIME 2024, AIME 2025 and HMMT Feb 2025. Mean averages over AIME24/AIME25/HMMT.
{keep\_frac (\%)}: fraction of response tokens supervised among all generated tokens; 
{revKL}: reverse KL divergence between the student and teacher; {freeze (\%)}: fraction of student network unchanged parameters after training.}
\label{tab:family5-tokenmask}
\begin{tabular}{l cccc c c c}
\toprule
 & \multicolumn{4}{c}{avg@8 (\%)} & & & \\
\cmidrule(lr){2-5}
Method & AIME24 & AIME25 & HMMT & Mean & keep\_frac (\%) & revKL & freeze (\%) \\
\midrule
Teacher & 62.1 & 48.3 & 30.0 & 46.8 & --    & 0.000 & --   \\
Student & 12.9 & 8.3  & 5.0  & 8.7  & --    & 0.371 & --   \\
plain OPD & 36.7 & 25.8 & 15.8 & 26.1 & 100     & 0.318 & 80.1 \\
\cmidrule(lr){1-8}
\multicolumn{8}{c}{\emph{one token supervision per trajectory}} \\[3pt]
rand1tok & 19.6 & 16.7 & 9.2  & 15.1 & 0.0431  & 0.317 & 94.5 \\
mintok & 31.7 & 22.9 & 18.8 & 24.4 & 0.0240  & 0.366 & 90.8 \\
maxtok & 37.1 & 28.7 & 20.0 & \textbf{28.6} & 0.0169  & 1.227 & 89.8 \\
\cmidrule(lr){1-8}
\multicolumn{8}{c}{\emph{two tokens supervision per trajectory}} \\[3pt]
minmaxtok & 38.3 & 28.7 & 17.9 & 28.3 & 0.0414  & 0.481 & 90.8 \\
\cmidrule(lr){1-8}
\multicolumn{8}{c}{\emph{0.1\% tokens supervision per trajectory}} \\[3pt]
randmask 0.1\% & 35.4 & 21.2 & 15.0 & 23.9 & 0.0994  & 0.397 & 92.3 \\
pctltail 0.05\% & 37.9 & 31.7 & 16.2 & \textbf{28.6} & 0.138   & 0.449 & 88.8 \\
\bottomrule
\end{tabular}
\end{table}

\begin{table}[t]
\centering
\small
\caption{Family 6: Qwen3-8B $\leftarrow$ Qwen3-30B-A3B-Instruct-2507. avg@8 scores of the base student model, teacher model, student model trained with plain OPD and student models trained with sparse OPD variants across AIME 2024, AIME 2025 and HMMT Feb 2025. Mean averages over AIME24/AIME25/HMMT.
{keep\_frac (\%)}: fraction of response tokens supervised among all generated tokens; 
{revKL}: reverse KL divergence between the student and teacher; {freeze (\%)}: fraction of student network unchanged parameters after training.}
\label{tab:family6-tokenmask}
\begin{tabular}{l cccc c c c}
\toprule
 & \multicolumn{4}{c}{avg@8 (\%)} & & & \\
\cmidrule(lr){2-5}
Method & AIME24 & AIME25 & HMMT & Mean & keep\_frac (\%) & revKL & freeze (\%) \\
\midrule
Teacher & 72.5 & 62.1 & 42.5 & 59.0 & --    & 0.000 & --   \\
Student & 26.2 & 20.8 & 10.8 & 19.3 & --    & 0.268 & --   \\
plain OPD & 65.4 & 54.2 & 29.2 & 49.6 & 100     & 0.158 & 78.8 \\
\cmidrule(lr){1-8}
\multicolumn{8}{c}{\emph{one token supervision per trajectory}} \\[3pt]
rand1tok & 50.4 & 37.9 & 22.9 & 37.1 & 0.0361  & 0.214 & 96.0 \\
mintok & 63.3 & 52.9 & 32.9 & 49.7 & 0.0253  & 0.192 & 88.2 \\
maxtok & 61.7 & 55.0 & 32.1 & 49.6 & 0.0220  & 0.721 & 89.9 \\
\cmidrule(lr){1-8}
\multicolumn{8}{c}{\emph{two tokens supervision per trajectory}} \\[3pt]
minmaxtok & 63.3 & 55.4 & 30.8 & 49.9 & 0.0391  & 0.263 & 88.9 \\
\cmidrule(lr){1-8}
\multicolumn{8}{c}{\emph{0.1\% tokens supervision per trajectory}} \\[3pt]
randmask 0.1\% & 62.5 & 53.3 & 32.1 & 49.3 & 0.100   & 0.214 & 91.3 \\
pctltail 0.05\% & 67.1 & 55.8 & 34.6 & \textbf{52.5} & 0.148   & 0.236 & 86.9 \\
\bottomrule
\end{tabular}
\end{table}

\begin{table}[t]
\centering
\small
\caption{Family 7: Qwen3-4B $\leftarrow$ Qwen3-30B-A3B-Instruct-2507. avg@8 scores of the base student model, teacher model, student model trained with plain OPD and student models trained with sparse OPD variants across AIME 2024, AIME 2025 and HMMT Feb 2025. Mean averages over AIME24/AIME25/HMMT.
{keep\_frac (\%)}: fraction of response tokens supervised among all generated tokens; 
{revKL}: reverse KL divergence between the student and teacher; {freeze (\%)}: fraction of student network unchanged parameters after training.}
\label{tab:family7-tokenmask}
\begin{tabular}{l cccc c c c}
\toprule
 & \multicolumn{4}{c}{avg@8 (\%)} & & & \\
\cmidrule(lr){2-5}
Method & AIME24 & AIME25 & HMMT & Mean & keep\_frac (\%) & revKL & freeze (\%) \\
\midrule
Teacher & 72.5 & 62.1 & 42.5 & 59.0 & --    & 0.000 & --   \\
Student & 24.6 & 20.0 & 12.9 & 19.2 & --    & 0.303 & --   \\
plain OPD & 60.4 & 47.9 & 34.6 & 47.6 & 100     & 0.169 & 74.5 \\
\cmidrule(lr){1-8}
\multicolumn{8}{c}{\emph{one token supervision per trajectory}} \\[3pt]
rand1tok & 41.7 & 35.8 & 21.2 & 32.9 & 0.0362  & 0.220 & 93.2 \\
mintok & 56.7 & 55.4 & 32.5 & 48.2 & 0.0243  & 0.203 & 86.1 \\
maxtok & 56.2 & 50.8 & 27.5 & 44.9 & 0.0178  & 0.724 & 87.8 \\
\cmidrule(lr){1-8}
\multicolumn{8}{c}{\emph{two tokens supervision per trajectory}} \\[3pt]
minmaxtok & 63.3 & 57.5 & 33.3 & 51.4 & 0.0429  & 0.306 & 86.2 \\
\cmidrule(lr){1-8}
\multicolumn{8}{c}{\emph{0.1\% tokens supervision per trajectory}} \\[3pt]
randmask 0.1\% & 58.3 & 51.7 & 35.0 & 48.3 & 0.1000  & 0.223 & 91.4 \\
pctltail 0.05\% & 64.6 & 54.6 & 37.1 & \textbf{52.1} & 0.1470  & 0.271 & 84.0 \\
\bottomrule
\end{tabular}
\end{table}

\begin{table}[t]
\centering
\small
\caption{Family 8: Qwen3-1.7B $\leftarrow$ Qwen3-30B-A3B-Instruct-2507. avg@8 scores of the base student model, teacher model, student model trained with plain OPD and student models trained with sparse OPD variants across AIME 2024, AIME 2025 and HMMT Feb 2025. Mean averages over AIME24/AIME25/HMMT.
{keep\_frac (\%)}: fraction of response tokens supervised among all generated tokens; 
{revKL}: reverse KL divergence between the student and teacher; {freeze (\%)}: fraction of student network unchanged parameters after training.}
\label{tab:family8-tokenmask-rep}
\begin{tabular}{l cccc c c c}
\toprule
 & \multicolumn{4}{c}{avg@8 (\%)} & & & \\
\cmidrule(lr){2-5}
Method & AIME24 & AIME25 & HMMT & Mean & keep\_frac (\%) & revKL & freeze (\%) \\
\midrule
Teacher & 72.5 & 62.1 & 42.5 & 59.0 & --    & 0.000 & --   \\
Student & 12.9 & 8.3  & 5.0  & 8.7  & --    & 0.368 & --   \\
plain OPD & 37.1 & 27.9 & 17.5 & 27.5 & 100     & 0.190 & 80.2 \\
\cmidrule(lr){1-8}
\multicolumn{8}{c}{\emph{one token supervision per trajectory}} \\[3pt]
rand1tok & 21.7 & 15.8 & 9.2  & 15.6 & 0.0379  & 0.299 & 95.0 \\
mintok & 32.9 & 30.0 & 17.5 & 26.8 & 0.0254  & 0.260 & 91.4 \\
maxtok & 35.8 & 32.1 & 19.2 & 29.0 & 0.0176  & 0.750 & 89.5 \\
\cmidrule(lr){1-8}
\multicolumn{8}{c}{\emph{two tokens supervision per trajectory}} \\[3pt]
minmaxtok & 40.4 & 29.6 & 16.7 & 28.9 & 0.0409  & 0.374 & 91.3 \\
\cmidrule(lr){1-8}
\multicolumn{8}{c}{\emph{0.1\% tokens supervision per trajectory}} \\[3pt]
randmask 0.1\% & 37.1 & 23.3 & 16.2 & 25.6 & 0.0988  & 0.267 & 92.6 \\
pctltail 0.05\% & 38.8 & 31.2 & 20.4 & \textbf{30.1} & 0.1380  & 0.404 & 89.6 \\
\bottomrule
\end{tabular}
\end{table}

\begin{table}[t]
\centering
\small
\caption{Family 9: Qwen3-8B $\leftarrow$ Qwen3-4B-Instruct-2507. avg@8 scores of the base student model, teacher model, student model trained with plain OPD and student models trained with sparse OPD variants across AIME 2024, AIME 2025 and HMMT Feb 2025. Mean averages over AIME24/AIME25/HMMT.
{keep\_frac (\%)}: fraction of response tokens supervised among all generated tokens; 
{revKL}: reverse KL divergence between the student and teacher; {freeze (\%)}: fraction of student network unchanged parameters after training.}
\label{tab:family9-tokenmask-rep}
\begin{tabular}{l cccc c c c}
\toprule
 & \multicolumn{4}{c}{avg@8 (\%)} & & & \\
\cmidrule(lr){2-5}
Method & AIME24 & AIME25 & HMMT & Mean & keep\_frac (\%) & revKL & freeze (\%) \\
\midrule
Teacher & 62.1 & 48.3 & 30.0 & 46.8 & --      & 0.000 & --   \\
Student & 26.2 & 20.8 & 10.8 & 19.3 & --      & 0.268 & --   \\
plain OPD & 66.7 & 44.6 & 31.2 & 47.5 & 100     & 0.184 & 78.2 \\
\cmidrule(lr){1-8}
\multicolumn{8}{c}{\emph{one token supervision per trajectory}} \\[3pt]
rand1tok & 42.1 & 30.0 & 17.5 & 29.9 & 0.0392  & 0.253 & 95.5 \\
mintok & 55.4 & 47.1 & 25.8 & 42.8 & 0.0259  & 0.213 & 87.3 \\
maxtok & 63.8 & 50.8 & 29.6 & 48.1 & 0.0167  & 1.116 & 90.7 \\
\cmidrule(lr){1-8}
\multicolumn{8}{c}{\emph{two tokens supervision per trajectory}} \\[3pt]
minmaxtok & 62.9 & 55.0 & 30.0 & \textbf{49.3} & 0.0472  & 0.319 & 88.1 \\
\cmidrule(lr){1-8}
\multicolumn{8}{c}{\emph{0.1\% tokens supervision per trajectory}} \\[3pt]
randmask 0.1\% & 59.6 & 45.4 & 28.3 & 44.4 & 0.0999  & 0.279 & 94.1 \\
pctltail 0.05\% & 62.1 & 47.5 & 28.8 & 46.1 & 0.120   & 0.254 & 86.1 \\
\bottomrule
\end{tabular}
\end{table}

\begin{table}[t]
\centering
\small
\caption{Family 1: Qwen3-1.7B-Base $\leftarrow$ Qwen3-4B-GRPO-1ep. avg@8 scores of the base student model, teacher model, student model trained with plain OPD and student models trained with sparse OPD variants across AIME 2024, AIME 2025 and HMMT Feb 2025. Mean averages over AIME24/AIME25/HMMT.
{keep\_frac (\%)}: fraction of response tokens supervised among all generated tokens; 
{revKL}: reverse KL divergence between the student and teacher; {freeze (\%)}: fraction of student network unchanged parameters after training.}
\label{tab:family1-threshold}
\begin{tabular}{l cccc r r r}
\toprule
 & \multicolumn{4}{c}{avg@8 (\%)} & & & \\
\cmidrule(lr){2-5}
 & AIME24 & AIME25 & HMMT & Mean & keep\% & revKL & Spars.\% \\
\midrule
Teacher (Qwen3-4B-GRPO-1ep) & 13.3 & 12.9 & 3.3 & 9.8 & --   & 0.000 & --   \\
Student (Qwen3-1.7B-Base)   & 1.7  & 0.8  & 0.4 & 1.0 & --   & 0.156 & --   \\
\midrule
plain OPD                   & 8.3  & 5.0  & 0.4 & 4.6 & 100  & 0.030 & 82.2 \\
at$<\!-1$   & 7.9 & 5.0 & 0.4 & 4.4 & 3.7900  & 0.024 & 82.2 \\
at$<\!-2$   & 9.2 & 3.8 & 0.0 & 4.3 & 2.3700  & 0.021 & 82.2 \\
at$<\!-8$   & 7.1 & 5.8 & 1.2 & \textbf{4.7} & 0.1960  & 0.031 & 85.2 \\
at$<\!-16$  & 6.7 & 6.2 & 0.0 & 4.3 & 0.0079  & 0.033 & 87.8 \\
at$<\!-32$  & 2.5 & 2.5 & 0.0 & 1.7 & 0.0000  & 0.085 & 89.8 \\
at$>\!0.5$  & 3.8 & 0.4 & 0.8 & 1.7 & 6.4700  & 0.379 & 74.4 \\
at$>\!1$    & 0.8 & 0.0 & 0.4 & 0.4 & 1.9700  & 0.430 & 72.9 \\
at$>\!3.5$  & 2.9 & 1.7 & 0.4 & 1.7 & 0.0045  & 0.189 & 90.4 \\
\bottomrule
\end{tabular}
\end{table}

\begin{table}[t]
\centering
\small
\caption{Family 2: Qwen3-1.7B-Base $\leftarrow$ Qwen3-4B-GRPO-5ep. avg@8 scores of the base student model, teacher model, student model trained with plain OPD and student models trained with sparse OPD variants across AIME 2024, AIME 2025 and HMMT Feb 2025. Mean averages over AIME24/AIME25/HMMT.
{keep\_frac (\%)}: fraction of response tokens supervised among all generated tokens; 
{revKL}: reverse KL divergence between the student and teacher; {freeze (\%)}: fraction of student network unchanged parameters after training.}
\label{tab:family2-threshold}
\begin{tabular}{l cccc r r r}
\toprule
 & \multicolumn{4}{c}{avg@8 (\%)} & & & \\
\cmidrule(lr){2-5}
 & AIME24 & AIME25 & HMMT & Mean & keep\% & revKL & Spars.\% \\
\midrule
Teacher (Qwen3-4B-GRPO-5ep) & 20.8 & 17.9 & 8.8 & 15.8 & --   & 0.000 & --   \\
Student (Qwen3-1.7B-Base)   & 1.7  & 0.8  & 0.4 & 1.0  & --   & 0.361 & --   \\
\midrule
plain OPD                   & 10.8 & 9.6  & 2.1 & \textbf{7.5} & 100  & 0.138 & 78.7 \\
at$<\!-1$   & 9.6  & 7.1 & 2.1 & 6.2 & 4.6200  & 0.106 & 79.1 \\
at$<\!-2$   & 9.6  & 5.4 & 2.1 & 5.7 & 3.8100  & 0.053 & 79.1 \\
at$<\!-8$   & 10.8 & 7.5 & 1.2 & 6.5 & 1.4700  & 0.079 & 80.0 \\
at$<\!-16$  & 11.2 & 5.8 & 2.1 & 6.4 & 0.3540  & 0.086 & 82.1 \\
at$<\!-32$  & 8.8  & 3.8 & 0.4 & 4.3 & 0.0028  & 0.097 & 87.4 \\
at$>\!0.5$  & 6.7  & 3.3 & 1.2 & 3.8 & 5.4600  & 0.629 & 78.2 \\
at$>\!1$    & 2.5  & 0.8 & 0.4 & 1.2 & 2.1000  & 0.735 & 75.3 \\
at$>\!3.5$  & 2.1  & 0.8 & 0.0 & 1.0 & 0.0073  & 0.317 & 89.2 \\
\bottomrule
\end{tabular}
\end{table}

\begin{table}[t]
\centering
\small
\caption{Family 3: Qwen3-4B-Base $\leftarrow$ Qwen3-4B-GRPO-1ep. avg@8 scores of the base student model, teacher model, student model trained with plain OPD and student models trained with sparse OPD variants across AIME 2024, AIME 2025 and HMMT Feb 2025. Mean averages over AIME24/AIME25/HMMT.
{keep\_frac (\%)}: fraction of response tokens supervised among all generated tokens; 
{revKL}: reverse KL divergence between the student and teacher; {freeze (\%)}: fraction of student network unchanged parameters after training.}
\label{tab:family3-threshold}
\begin{tabular}{l cccc r r r}
\toprule
 & \multicolumn{4}{c}{avg@8 (\%)} & & & \\
\cmidrule(lr){2-5}
 & AIME24 & AIME25 & HMMT & Mean & keep\% & revKL & Spars.\% \\
\midrule
Teacher (Qwen3-4B-GRPO-1ep) & 13.3 & 12.9 & 3.3 & 9.8 & --   & 0.000 & --   \\
Student (Qwen3-4B-Base)     & 9.6  & 7.9  & 0.8 & 6.1 & --   & 0.082 & --   \\
\midrule
plain OPD                   & 15.8 & 14.2 & 4.2 & 11.4 & 100 & 0.001 & 84.0 \\
at$<\!-1$   & 19.6 & 16.2 & 5.0 & \textbf{13.6} & 0.1560  & 0.001 & 83.9  \\
at$<\!-2$   & 15.8 & 16.2 & 4.2 & 12.1 & 0.1400  & 0.002 & 83.8  \\
at$<\!-4$   & 18.3 & 12.9 & 3.8 & 11.7 & 0.0415  & 0.002 & 84.9  \\
at$<\!-8$   & 13.8 & 12.5 & 2.9 & 9.7  & 0.0041  & 0.008 & 88.0  \\
at$<\!-16$  & 10.8 & 7.5  & 1.2 & 6.5  & 0.0002  & 0.030 & 92.1  \\
at$<\!-32$  & 10.0 & 5.8  & 1.2 & 5.7  & 0.0000  & 0.019 & 100.0 \\
at$>\!0.5$  & 15.8 & 12.9 & 1.2 & 10.0 & 0.1660  & 0.003 & 87.9  \\
at$>\!1$    & 15.8 & 12.5 & 3.3 & 10.6 & 0.0196  & 0.024 & 89.1  \\
at$>\!3.5$  & 13.8 & 10.0 & 2.9 & 8.9  & 0.0001  & 0.030 & 92.7  \\
\bottomrule
\end{tabular}
\end{table}

\begin{table}[t]
\centering
\small
\caption{Family 4: Qwen3-4B-Base $\leftarrow$ Qwen3-4B-GRPO-5ep. avg@8 scores of the base student model, teacher model, student model trained with plain OPD and student models trained with sparse OPD variants across AIME 2024, AIME 2025 and HMMT Feb 2025. Mean averages over AIME24/AIME25/HMMT.
{keep\_frac (\%)}: fraction of response tokens supervised among all generated tokens; 
{revKL}: reverse KL divergence between the student and teacher; {freeze (\%)}: fraction of student network unchanged parameters after training.}
\label{tab:family4-threshold}
\begin{tabular}{l cccc r r r}
\toprule
 & \multicolumn{4}{c}{avg@8 (\%)} & & & \\
\cmidrule(lr){2-5}
 & AIME24 & AIME25 & HMMT & Mean & keep\% & revKL & Spars.\% \\
\midrule
Teacher (Qwen3-4B-GRPO-5ep) & 20.8 & 17.9 & 8.8 & 15.8 & --   & 0.000 & --   \\
Student (Qwen3-4B-Base)     & 9.6  & 7.9  & 0.8 & 6.1  & --   & 0.441 & --   \\
\midrule
plain OPD                   & 23.3 & 17.5 & 7.5 & 16.1 & 100  & 0.005 & 75.0 \\
at$<\!-1$   & 19.6 & 18.8 & 5.0  & 14.4 & 0.6920  & 0.007 & 75.2 \\
at$<\!-2$   & 20.4 & 18.3 & 6.7  & 15.1 & 0.6830  & 0.006 & 78.1 \\
at$<\!-4$   & 19.6 & 17.1 & 5.8  & 14.2 & 0.3780  & 0.008 & 78.7 \\
at$<\!-8$   & 17.1 & 17.9 & 5.0  & 13.3 & 0.1840  & 0.015 & 80.0 \\
at$<\!-16$  & 16.7 & 12.9 & 5.0  & 11.5 & 0.0339  & 0.024 & 82.2 \\
at$<\!-32$  & 17.1 & 13.3 & 4.6  & 11.7 & 0.0007  & 0.067 & 88.5 \\
at$>\!0.5$  & 24.2 & 21.2 & 7.9  & \textbf{17.8} & 1.1500  & 0.040 & 81.0 \\
at$>\!1$    & 19.6 & 18.3 & 10.4 & 16.1 & 0.3510  & 0.193 & 85.5 \\
at$>\!3.5$  & 18.3 & 13.8 & 5.4  & 12.5 & 0.0009  & 0.435 & 90.9 \\
\bottomrule
\end{tabular}
\end{table}

\begin{table}[t]
\centering
\small
\caption{Family 5: Qwen3-1.7B $\leftarrow$ Qwen3-4B-Instruct-2507. avg@8 scores of the base student model, teacher model, student model trained with plain OPD and student models trained with sparse OPD variants across AIME 2024, AIME 2025 and HMMT Feb 2025. Mean averages over AIME24/AIME25/HMMT.
{keep\_frac (\%)}: fraction of response tokens supervised among all generated tokens; 
{revKL}: reverse KL divergence between the student and teacher; {freeze (\%)}: fraction of student network unchanged parameters after training.}
\label{tab:family5-threshold}
\begin{tabular}{l cccc r r r}
\toprule
 & \multicolumn{4}{c}{avg@8 (\%)} & & & \\
\cmidrule(lr){2-5}
 & AIME24 & AIME25 & HMMT & Mean & keep\% & revKL & Spars.\% \\
\midrule
Teacher (Qwen3-4B-Instruct-2507) & 62.1 & 48.3 & 30.0 & 46.8 & --   & 0.000 & --   \\
Student (Qwen3-1.7B)             & 12.9 & 8.3  & 5.0  & 8.7  & --   & 0.371 & --   \\
\midrule
plain OPD                        & 36.7 & 25.8 & 15.8 & 26.1 & 100  & 0.318 & 80.1 \\
at$<\!-1$   & 32.1 & 22.9 & 17.5 & 24.2 & 9.2400  & 0.291 & 81.5 \\
at$<\!-2$   & 36.2 & 24.2 & 16.7 & 25.7 & 5.0200  & 0.300 & 81.7 \\
at$<\!-4$   & 36.2 & 22.9 & 14.2 & 24.4 & 1.8500  & 0.275 & 83.2 \\
at$<\!-8$   & 33.8 & 25.0 & 14.2 & 24.3 & 0.3700  & 0.278 & 86.3 \\
at$<\!-16$  & 32.9 & 25.4 & 15.4 & 24.6 & 0.0302  & 0.352 & 90.6 \\
at$<\!-32$  & 14.6 & 12.5 & 6.7  & 11.2 & 0.0002  & 0.379 & 93.6 \\
at$>\!0.5$  & 35.4 & 32.5 & 15.8 & \textbf{27.9} & 7.5000  & 0.686 & 79.6 \\
at$>\!1$    & 30.8 & 20.4 & 13.3 & 21.5 & 2.6600  & 1.011 & 80.2 \\
at$>\!2$    & 37.5 & 28.7 & 17.1 & 27.8 & 0.3180  & 1.357 & 83.6 \\
at$>\!3.5$  & 38.3 & 26.2 & 19.2 & \textbf{27.9} & 0.0136  & 1.231 & 89.6 \\
\bottomrule
\end{tabular}
\end{table}

\begin{table}[t]
\centering
\small
\caption{Family 6: Qwen3-8B $\leftarrow$ Qwen3-30B-A3B-Instruct-2507. avg@8 scores of the base student model, teacher model, student model trained with plain OPD and student models trained with sparse OPD variants across AIME 2024, AIME 2025 and HMMT Feb 2025. Mean averages over AIME24/AIME25/HMMT.
{keep\_frac (\%)}: fraction of response tokens supervised among all generated tokens; 
{revKL}: reverse KL divergence between the student and teacher; {freeze (\%)}: fraction of student network unchanged parameters after training.}
\label{tab:family6-threshold}
\begin{tabular}{l cccc r r r}
\toprule
 & \multicolumn{4}{c}{avg@8 (\%)} & & & \\
\cmidrule(lr){2-5}
 & AIME24 & AIME25 & HMMT & Mean & keep\% & revKL & Spars.\% \\
\midrule
Teacher (Qwen3-30B-A3B-Instruct-2507) & 72.5 & 62.1 & 42.5 & 59.0 & --   & 0.000 & --   \\
Student (Qwen3-8B)                    & 26.2 & 20.8 & 10.8 & 19.3 & --   & 0.268 & --   \\
\midrule
plain OPD                             & 65.4 & 54.2 & 29.2 & \textbf{49.6} & 100 & 0.158 & 78.8 \\
at$<\!-1$   & 64.6 & 53.8 & 30.0 & 49.4 & 6.0500  & 0.141 & 80.2 \\
at$<\!-2$   & 65.8 & 51.2 & 30.0 & 49.0 & 2.8800  & 0.130 & 80.8 \\
at$<\!-4$   & 59.2 & 50.8 & 28.3 & 46.1 & 0.8790  & 0.128 & 82.5 \\
at$<\!-8$   & 63.7 & 52.9 & 30.0 & 48.9 & 0.1490  & 0.147 & 86.2 \\
at$<\!-16$  & 57.9 & 43.8 & 29.2 & 43.6 & 0.0104  & 0.206 & 89.6 \\
at$<\!-32$  & 28.7 & 21.2 & 15.0 & 21.7 & 0.0000  & 0.218 & 96.1 \\
at$>\!0.5$  & 59.6 & 49.6 & 29.6 & 46.2 & 5.9900  & 0.415 & 80.5 \\
at$>\!1$    & 53.8 & 50.8 & 26.7 & 43.8 & 1.8900  & 0.660 & 80.3 \\
at$>\!3.5$  & 55.4 & 54.2 & 32.1 & 47.2 & 0.0104  & 0.820 & 88.0 \\
\bottomrule
\end{tabular}
\end{table}

\begin{table}[t]
\centering
\small
\caption{Family 7: Qwen3-4B $\leftarrow$ Qwen3-30B-A3B-Instruct-2507. avg@8 scores of the base student model, teacher model, student model trained with plain OPD and student models trained with sparse OPD variants across AIME 2024, AIME 2025 and HMMT Feb 2025. Mean averages over AIME24/AIME25/HMMT.
{keep\_frac (\%)}: fraction of response tokens supervised among all generated tokens; 
{revKL}: reverse KL divergence between the student and teacher; {freeze (\%)}: fraction of student network unchanged parameters after training.}
\label{tab:family7-threshold}
\begin{tabular}{l cccc r r r}
\toprule
 & \multicolumn{4}{c}{avg@8 (\%)} & & & \\
\cmidrule(lr){2-5}
 & AIME24 & AIME25 & HMMT & Mean & keep\% & revKL & Spars.\% \\
\midrule
Teacher (Qwen3-30B-A3B-Instruct-2507) & 72.5 & 62.1 & 42.5 & 59.0 & --   & 0.000 & --   \\
Student (Qwen3-4B)                    & 24.6 & 20.0 & 12.9 & 19.2 & --   & 0.303 & --   \\
\midrule
plain OPD                             & 60.4 & 47.9 & 34.6 & 47.6 & 100  & 0.169 & 74.5 \\
at$<\!-2$   & 56.7 & 47.1 & 31.7 & 45.1 & 3.1200  & 0.129 & 76.8 \\
at$<\!-4$   & 57.1 & 53.3 & 29.6 & 46.7 & 0.9820  & 0.133 & 79.0 \\
at$<\!-8$   & 58.3 & 53.3 & 32.5 & \textbf{48.1} & 0.1660  & 0.174 & 83.2 \\
at$<\!-16$  & 47.9 & 42.1 & 29.2 & 39.7 & 0.0116  & 0.221 & 87.3 \\
at$<\!-32$  & 25.8 & 22.5 & 15.8 & 21.4 & 0.0000  & 0.247 & 91.2 \\
at$>\!1$    & 51.2 & 44.6 & 22.1 & 39.3 & 1.9900  & 0.701 & 76.2 \\
at$>\!3.5$  & 54.2 & 52.1 & 30.0 & 45.4 & 0.0111  & 0.691 & 86.3 \\
\bottomrule
\end{tabular}
\end{table}

\begin{table}[t]
\centering
\small
\caption{Family 8: Qwen3-1.7B $\leftarrow$ Qwen3-30B-A3B-Instruct-2507. avg@8 scores of the base student model, teacher model, student model trained with plain OPD and student models trained with sparse OPD variants across AIME 2024, AIME 2025 and HMMT Feb 2025. Mean averages over AIME24/AIME25/HMMT.
{keep\_frac (\%)}: fraction of response tokens supervised among all generated tokens; 
{revKL}: reverse KL divergence between the student and teacher; {freeze (\%)}: fraction of student network unchanged parameters after training.}
\label{tab:family8-threshold}
\begin{tabular}{l cccc r r r}
\toprule
 & \multicolumn{4}{c}{avg@8 (\%)} & & & \\
\cmidrule(lr){2-5}
 & AIME24 & AIME25 & HMMT & Mean & keep\% & revKL & Spars.\% \\
\midrule
Teacher (Qwen3-30B-A3B-Instruct-2507) & 72.5 & 62.1 & 42.5 & 59.0 & --   & 0.000 & --   \\
Student (Qwen3-1.7B)                  & 12.9 & 8.3  & 5.0  & 8.7  & --   & 0.368 & --   \\
\midrule
plain OPD                             & 37.1 & 27.9 & 17.5 & \textbf{27.5} & 100 & 0.190 & 80.2 \\
at$<\!-2$   & 29.2 & 27.9 & 12.9 & 23.3 & 4.7700  & 0.116 & 82.1 \\
at$<\!-4$   & 26.2 & 24.2 & 11.7 & 20.7 & 1.6800  & 0.093 & 83.6 \\
at$<\!-8$   & 31.2 & 25.4 & 14.6 & 23.8 & 0.3290  & 0.157 & 87.4 \\
at$<\!-16$  & 25.4 & 21.7 & 12.1 & 19.7 & 0.0336  & 0.296 & 91.3 \\
at$<\!-32$  & 15.4 & 11.7 & 6.2  & 11.1 & 0.0002  & 0.354 & 92.3 \\
at$>\!1$    & 32.9 & 27.5 & 15.0 & 25.1 & 2.7700  & 0.685 & 79.4 \\
at$>\!3.5$  & 32.5 & 30.0 & 17.9 & 26.8 & 0.0143  & 0.789 & 89.1 \\
\bottomrule
\end{tabular}
\end{table}

\begin{table}[t]
\centering
\small
\caption{Family 9: Qwen3-8B $\leftarrow$ Qwen3-4B-Instruct-2507. avg@8 scores of the base student model, teacher model, student model trained with plain OPD and student models trained with sparse OPD variants across AIME 2024, AIME 2025 and HMMT Feb 2025. Mean averages over AIME24/AIME25/HMMT.
{keep\_frac (\%)}: fraction of response tokens supervised among all generated tokens; 
{revKL}: reverse KL divergence between the student and teacher; {freeze (\%)}: fraction of student network unchanged parameters after training.}
\label{tab:family9-threshold}
\begin{tabular}{l cccc r r r}
\toprule
 & \multicolumn{4}{c}{avg@8 (\%)} & & & \\
\cmidrule(lr){2-5}
 & AIME24 & AIME25 & HMMT & Mean & keep\% & revKL & Spars.\% \\
\midrule
Teacher (Qwen3-4B-Instruct-2507) & 62.1 & 48.3 & 30.0 & 46.8 & --   & 0.000 & --   \\
Student (Qwen3-8B)               & 26.2 & 20.8 & 10.8 & 19.3 & --   & 0.268 & --   \\
\midrule
plain OPD   & 66.7 & 44.6 & 31.2 & \textbf{47.5} & 100     & 0.184 & 78.2 \\
at$<\!-2$   & 60.4 & 49.6 & 28.3 & 46.1 & 2.7400  & 0.191 & 80.2 \\
at$<\!-4$   & 63.7 & 45.8 & 26.7 & 45.4 & 0.8540  & 0.214 & 81.8 \\
at$<\!-8$   & 55.8 & 45.0 & 28.3 & 43.0 & 0.1480  & 0.200 & 85.3 \\
at$<\!-16$  & 52.9 & 40.4 & 25.0 & 39.4 & 0.0100  & 0.170 & 89.1 \\
at$<\!-32$  & 36.2 & 24.2 & 13.3 & 24.6 & 0.0000  & 0.280 & 92.1 \\
at$>\!1$    & 51.2 & 43.8 & 23.8 & 39.6 & 1.5640  & 0.836 & 81.7 \\
at$>\!3.5$  & 63.7 & 46.7 & 25.0 & 45.1 & 0.0070  & 0.956 & 91.7 \\
\bottomrule
\end{tabular}
\end{table}

\begin{figure}
    \centering
    \includegraphics[width=\linewidth]{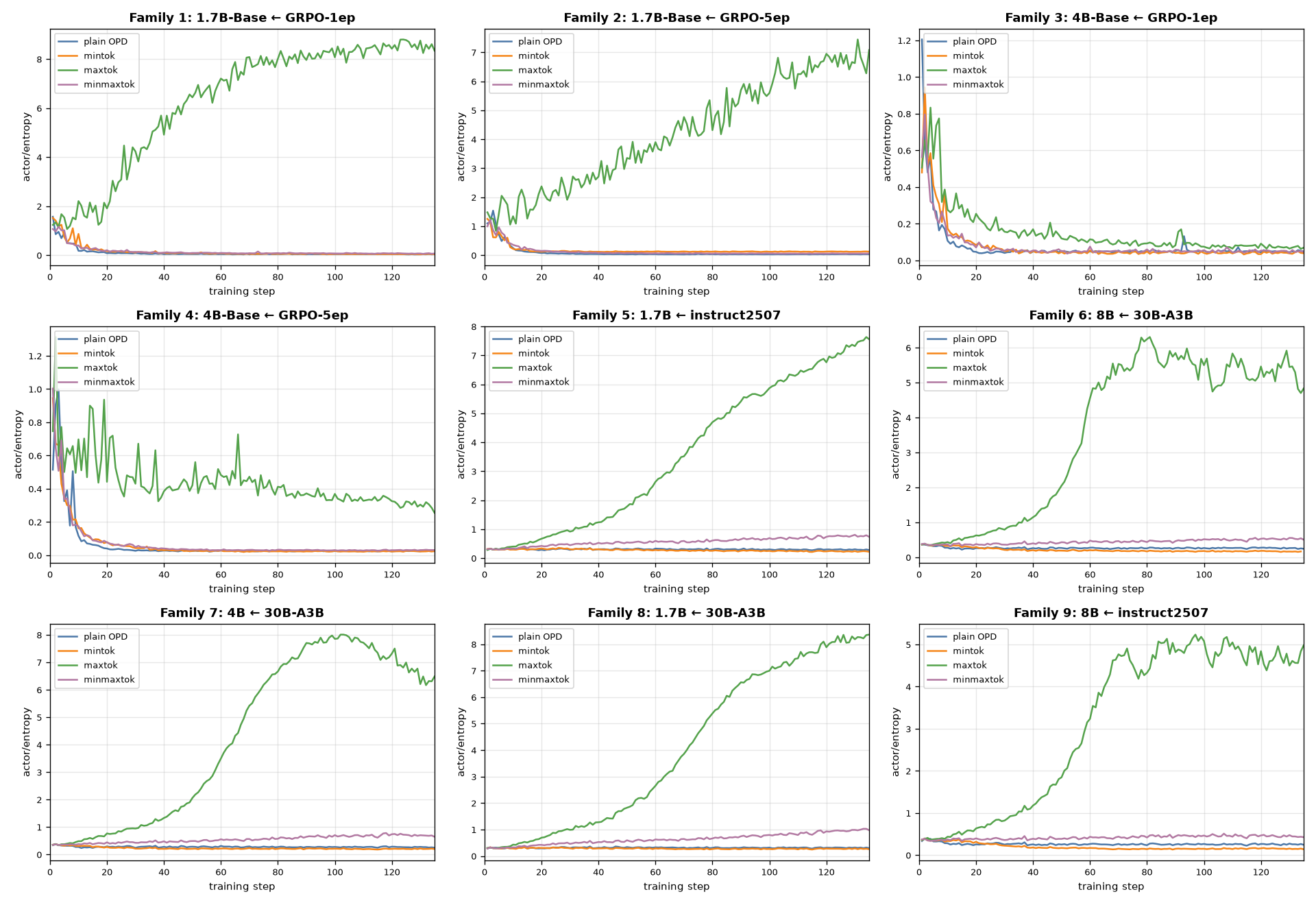}
    \caption{Learning dynamics of actor entropy across nine families.}
    \label{fig:actor-entropy}
\end{figure}

\begin{figure}
    \centering
    \includegraphics[width=0.48\linewidth]{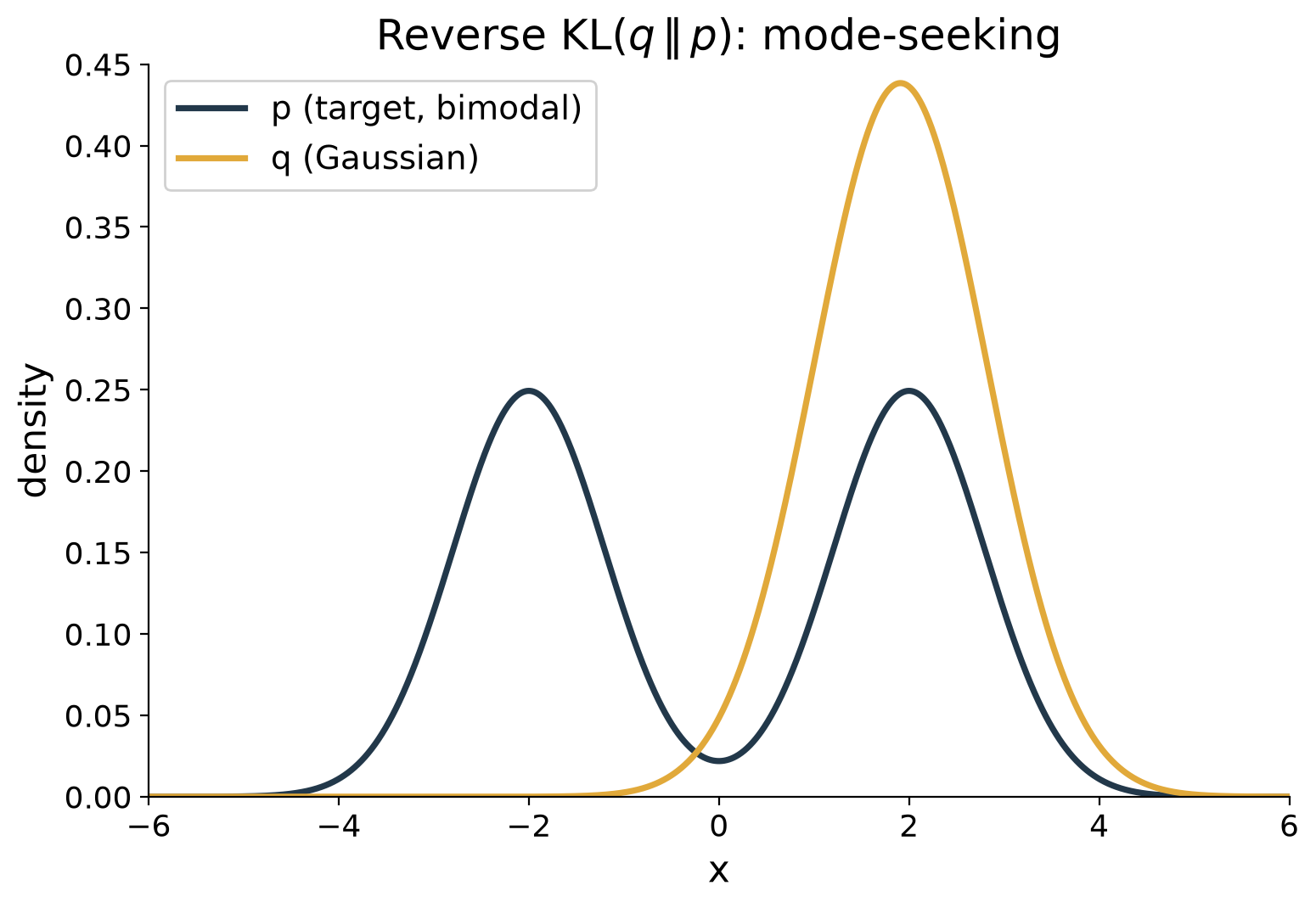}
    \includegraphics[width=0.48\linewidth]{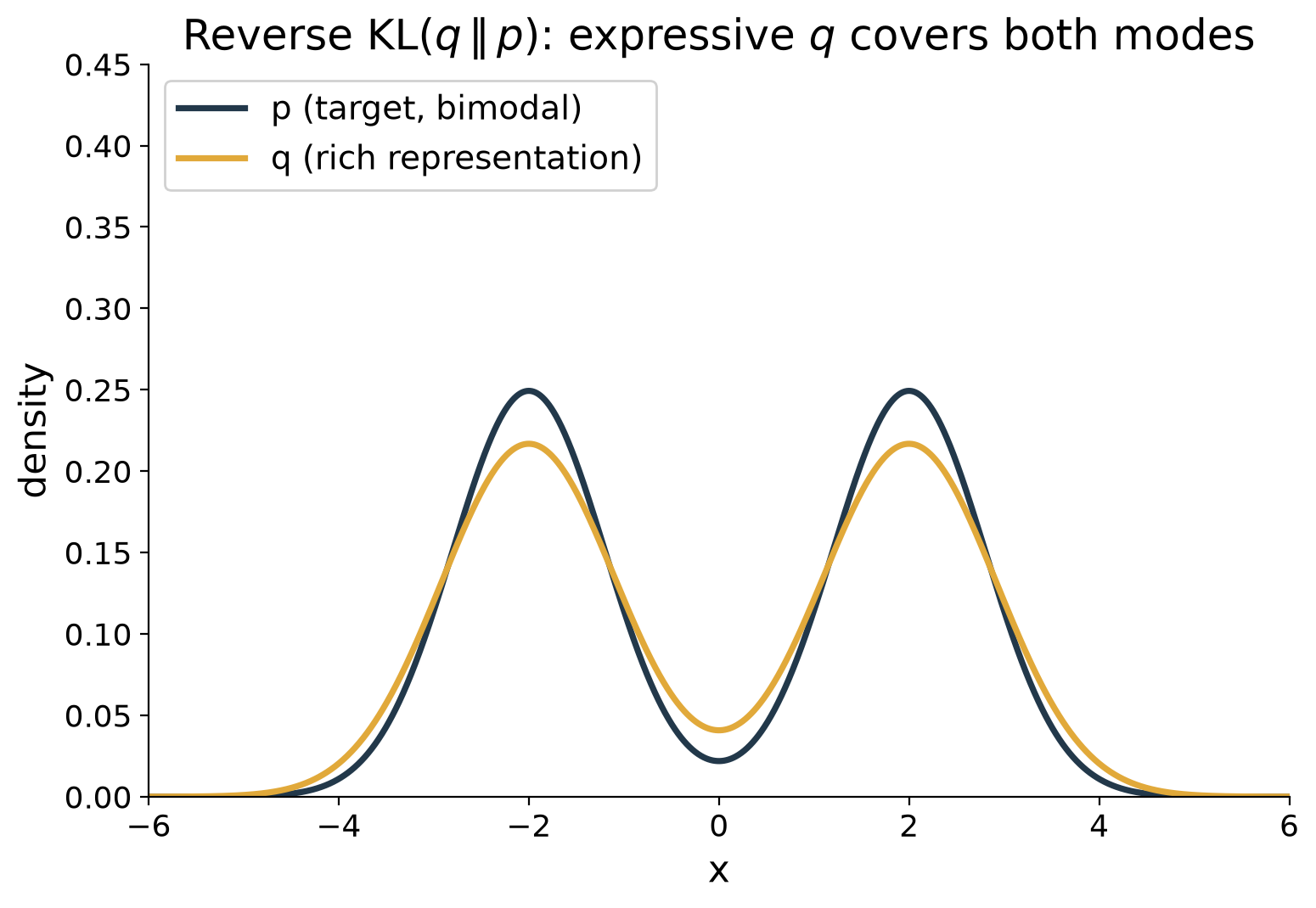}
    \caption{An toy-example illustration of minimizing reverse KL divergence between two distributions $p$ and $q$. Left: the mode-seeking behavior of minimizing reverse KL divergence between a uni-modal Gaussian distribution $q$ and a bimodal target distribution $p$. Right: if $q$ is expressive enough, then minimizing reverse KL divergence recovers the target distribution $p$.}
    \label{fig:minimizing-revKL}
\end{figure}

\section{Additional Results and Discussion}
\label{sec:Additional Results and Discussion}

\subsection{Memory Saving}
During the backpropagation at each step, the student's gradient path holds two kinds of activation: the transformer backbone activations across all layers, and the output logits $[B,T,V]$ produced by the language model (LM) head, where $B$ is the batch size, $T$ is the response length, $V$ is the vocabulary size.
Because the vocabulary size $V$ vastly exceeds the hidden width $H$, for example in Qwen3 models $V = 151{,}936$ versus $H \sim 2.5$–$4\times10^{3}$, the logit tensor, together with the softmax/KL intermediates derived from it, is the single dominant term in the footprint of the backward pass. It is precisely this term that forces small microbatches and caps the trainable context length and model size.
Sparse OPD updates the loss on only $O(B)$ positions (one or two tokens per trajectory). Under a gather-then-project realization, the LM head is applied only at those positions, so the retained logit activation shrinks from $[B,T,V]$ to $[B,V]$. Thus, for the sparse OPD variants the logit memory is effectively erased. For a detailed discussion on the implementation, which involves several engineering tricks, we refer readers to \citet{xu2026tip}.

\subsection{Response Length}
In this section, we compare the response length of all checkpoints obtained from the plain OPD and sparse OPD variants across all nine families. The response length statistics in \Cref{tab:resplen-compact} is calculated from response trajectories on AIME 2024, AIME 2025 and HMMT 2025 obtained during evaluation.

\begin{table*}[t]
\centering
\footnotesize
\setlength{\tabcolsep}{2pt}
\caption{Average response length (number of tokens) per checkpoint, each cell presents \emph{correct}/\emph{all}. \emph{correct}: mean length of correct responses; \emph{all}: mean over all responses.  F1 to F9 represent Family 1 to Family 9, respectively. `--' marks a method not included for that family.}
\label{tab:resplen-compact}
\begin{tabular}{l ccccccccc}
\toprule
Method & F1 & F2 & F3 & F4 & F5 & F6 & F7 & F8 & F9 \\
\midrule
Teacher & 1121/2789 & 2949/6293 & 1121/2789 & 2949/6293 & 4451/9028 & 5110/6606 & 5110/6606 & 5110/6606 & 4451/9028 \\
Student & 1524/2874 & 1524/2874 & 1012/1619 & 1012/1619 & 1524/2874 & 1667/3607 & 1724/3619 & 1524/2874 & 1667/3607 \\
\cmidrule(lr){1-10}
plain OPD & 1069/5740 & 3088/10193 & 1626/3988 & 2854/6486 & 5894/11138 & 7096/11760 & 7191/12395 & 7831/16643 & 5888/9059 \\
rand1tok & 950/4056 & 1623/7329 & 1644/3169 & 1722/5916 & 2457/5613 & 4684/8612 & 3891/7974 & 2744/6643 & 3146/6775 \\
minmaxtok & 1051/3691 & 2325/6864 & 1576/3923 & 2912/6189 & 7706/15692 & 9106/16248 & 10244/16328 & 8669/15479 & 7539/13320 \\
mintok & 995/5735 & 1947/9687 & 1825/4621 & 2749/6630 & 6070/13817 & 7494/14862 & 8215/15723 & 6519/13906 & 6143/12047 \\
maxtok & 1574/1749 & 3056/10694 & 1415/4028 & 3272/5182 & 12342/17283 & 15026/20484 & 15902/21346 & 15294/18449 & 12388/16326 \\
randmask 0.1\% & 1044/5090 & 2948/8367 & 1282/2947 & 2500/3676 & 4878/12749 & 7454/13130 & 7538/14130 & 6619/14809 & 5963/11074 \\
pctltail 0.05\% & 1545/4181 & 2312/6407 & 1445/4354 & 2752/6352 & 7888/15511 & 9543/16479 & 10774/18015 & 9441/18845 & 6321/10638 \\
\cmidrule(lr){1-10}
at$<\!-1$ & 1067/9122 & 2527/12958 & 1810/4092 & 3019/6920 & 5147/10665 & 7279/13425 & -- & -- & -- \\
at$<\!-2$ & 1083/9570 & 3349/12996 & 1308/4727 & 2949/7262 & 5107/10890 & 6734/14557 & 6882/15559 & 6786/23986 & 5473/8497 \\
at$<\!-4$ & -- & -- & 1613/3830 & 2443/7059 & 5349/12578 & 7789/17518 & 7837/17162 & 8471/26584 & 5441/8913 \\
at$<\!-8$ & 1119/7222 & 2039/11384 & 1255/2585 & 2042/4962 & 5674/14873 & 7908/17234 & 8435/17702 & 8698/24773 & 5368/10524 \\
at$<\!-16$ & 931/5740 & 2381/8257 & 2039/11384 & 1860/6230 & 6132/14363 & 5764/11221 & 5548/10388 & 3893/8264 & 8486/15591 \\
at$<\!-32$ & 1585/3804 & 2532/10186 & 1165/4968 & 2244/7020 & 1627/3201 & 2016/4158 & 2078/4676 & 1876/3546 & 2663/4939 \\
at$>\!0.5$ & 1362/2194 & 4437/5452 & 1321/2295 & 3117/5545 & 9993/16503 & 9682/14322 & -- & -- & -- \\
at$>\!1$ & 1429/1938 & 3800/4844 & 1205/1943 & 3261/5696 & 10727/18594 & 15320/20065 & 13730/20286 & 14712/21807 & 10432/15027 \\
at$>\!2$ & -- & -- & -- & -- & 14081/20855 & -- & -- & -- & -- \\
at$>\!3.5$ & 1260/2656 & 1560/5097 & 1393/2706 & 3047/5313 & 12358/17680 & 15159/19158 & 16598/20923 & 13893/17616 & 10319/14457 \\
\bottomrule
\end{tabular}
\end{table*}

\section{Details on Experiment Configuration}
\label{sec:Experiment Configuration}
In this section, we provide details on experiment configuration to reproduce all experiment results. 
Hyperparameters used to train the GRPO teachers, all OPD and sparse OPD variants are provided in
\Cref{tab:setup-hyperparams}. Hyperparameters used to train the PPO and sparse PPO variants are provided in \Cref{tab:ppo-config}. Prompts used for math and coding are as follows:

\paragraph{Prompt for math reasoning} 
\begin{verbatim}
MATH_INSTRUCTION = (
    "Please reason step by step, and put your final answer within \\boxed{}."
)
\end{verbatim}

\paragraph{Prompt for code reasoning} 
\begin{verbatim}
CODE_INSTRUCTION = (
    "Write Python code to solve the problem. Present the code in\n"
    "```python\n"
    "Your code\n"
    "```\n"
    "at the end.\n"
    "You need to think first then write the Python code."
)
\end{verbatim}

All OPD experiments are conducted on 8*A100 GPUs and 8*H100 GPUs, PPO experiments are conducted on 8*H100 GPUs, and evaluations are conducted on 8*H200 gpus. Each experiment takes 5-10 hours on corresponding GPUs, the total amount of GPU hours for this project is about 10,000.

\begin{table}[t]
\centering
\small
\caption{Training configuration for GRPO (teacher RL) and on-policy distillation (OPD).}
\label{tab:setup-hyperparams}
\begin{tabular}{l l l}
\toprule
 & \textbf{GRPO (teacher RL)} & \textbf{OPD (distillation)} \\
\midrule
Framework                   & verl 0.8.0                  & verl 0.8.0 (native OPD) \\
Objective                   & GRPO                        & Policy-gradient on-policy distillation (reverse KL) \\
Advantage estimator         & GRPO                        & GRPO \\
Base / student model        & Qwen3-4B-Base               & Qwen3-\{1.7B,\,4B,\,8B\}-Base \\
Teacher                     & ---                         & separate vLLM pool (Qwen3-4B / 30B-A3B) \\
Training data               & DAPO-Math-17k               & DAPO-Math-17k \\
Train batch size (prompts)  & 128                         & 128 \\
PPO mini-batch              & 64                          & 64 \\
Rollouts per prompt ($n$)   & 8                           & 1 \\
Max prompt length           & 1024                        & 1024 \\
Max response length         & 8192                        & 8192 \\
Learning rate               & $1\times10^{-6}$            & $1\times10^{-6}$ \\
LR warmup ratio             & 0.0                         & 0.0 \\
Weight decay                & 0.0                         & 0.0 \\
Gradient clip               & 1.0                         & 1.0 \\
PPO clip ratio ($\epsilon$) & 0.2                         & 0.2 (low = high) \\
Entropy coefficient         & 0.0                         & 0.0 \\
KL loss                     & on, coef 0.001, low\_var\_kl & off \\
KL in reward                & off                         & off \\
Rollout temperature / top-p & 1.0 / 1.0                   & 1.0 / 1.0 \\
Epochs                      & 5 (and 1-epoch variant)     & 1 (4B/8B), 2 (1.7B) \\
Optimizer                   & AdamW (verl default)        & AdamW (verl default) \\
Precision                   & bf16                        & bf16 \\
Gradient checkpointing      & yes                         & yes \\
\bottomrule
\end{tabular}
\end{table}

\begin{table}[t]
\centering
\small
\caption{PPO training configuration.}
\label{tab:ppo-config}
\begin{tabular}{ll}
\toprule
\textbf{Component} & \textbf{Configuration} \\
\midrule
\multicolumn{2}{l}{\textbf{Algorithm}} \\
Advantage estimator & GAE ($\gamma=1.0$, $\lambda=1.0$) \\
KL regularization & Disabled \\
KL reward / loss coefficient & $0$ \\
\midrule
\multicolumn{2}{l}{\textbf{Actor}} \\
Learning rate & $1\times10^{-5}$ \\
LR schedule / warmup & Constant / None \\
Weight decay & $0.1$ \\
PPO epochs / mini-batch & $1$ / $64$ \\
Clip range & $0.2$ / $0.2$ \\
Dual clip & $10.0$ \\
Entropy coefficient & $0$ \\
Gradient clipping & $1.0$ \\
Loss aggregation & Token-mean \\
Max tokens per GPU & $24{,}576$ \\
Precision & BF16 \\
FSDP size & $8$ \\
Optimizer offload & Enabled \\
Parameter offload & Disabled \\
\midrule
\multicolumn{2}{l}{\textbf{Critic}} \\
Learning rate & $1\times10^{-5}$ \\
LR schedule / warmup & Constant / None \\
Weight decay & $0.1$ \\
Value clip range & $0.5$ \\
Gradient clipping & $1.0$ \\
Critic warmup & $50$ steps \\
Max tokens per GPU & $24{,}576$ \\
Precision & BF16 \\
FSDP size & $8$ \\
Optimizer / parameter offload & Enabled / Enabled \\
\midrule
\multicolumn{2}{l}{\textbf{Rollout (vLLM)}} \\
Trajectories per prompt & $8$ \\
Temperature / top-$p$ / top-$k$ & $1.0$ / $1.0$ / $-1$ \\
Maximum model length & $10{,}240$ \\
Tensor parallelism & $1$ \\
GPU memory utilization & $0.5$ \\
Chunked prefill & Enabled \\
Max batched tokens / sequences & $49{,}152$ / $128$ \\
\midrule
\multicolumn{2}{l}{\textbf{Data and Training}} \\
Dataset & DAPO-math ($17{,}398$ prompts) \\
Training batch size & $128$ \\
Steps per epoch & $135$ \\
Training epochs / total steps & $1$ / $135$ \\
Maximum prompt / response length & $2{,}048$ / $8{,}192$ \\
Reward function & \texttt{math\_verify} (naive manager) \\
Hardware & $8$ GPUs, 1 node \\
\midrule
\multicolumn{2}{l}{\textbf{RL Framework}} \\
verl version & $0.8.0$ and $0.9.0$-dev \\
\bottomrule
\end{tabular}
\end{table}

\end{document}